\documentclass{ieeetj}
\usepackage{cite}
\usepackage{amsmath,amssymb,amsfonts}
\usepackage{algorithmic}
\usepackage{graphicx,color}
\usepackage{textcomp}
\usepackage{xcolor}
\usepackage{hyperref}
\usepackage{hanging}
\hypersetup{hidelinks=true}
\usepackage{algorithm}
\usepackage{array}
\usepackage{multirow}

\def\BibTeX{{\rm B\kern-.05em{\sc i\kern-.025em b}\kern-.08em
    T\kern-.1667em\lower.7ex\hbox{E}\kern-.125emX}}
\AtBeginDocument{\definecolor{tmlcncolor}{cmyk}{0.93,0.59,0.15,0.02}\definecolor{NavyBlue}{RGB}{0,86,125}}

\def\OJlogo{}
\def\seclogo{}

\def\authorrefmark#1{\ensuremath{^{\textbf{#1}}}}

\begin{document}
\receiveddate{}
\reviseddate{}
\accepteddate{}
\publisheddate{}
\currentdate{}
\doiinfo{}

\markboth{LARGE-SCALE TUNNEL AIR--GROUND COLLABORATION WITH FLISP}{GUO {ET AL.}}

\title{Large-Scale Tunnel Air--Ground Collaboration With FLISP: Fast LiDAR-IMU Synchronized Path Planner}

\author{FENGHE GUO\authorrefmark{1,2}, \IEEEmembership{Student Member, IEEE}, RUNJIE SHEN\authorrefmark{2,3}, CHENYANG SUN\authorrefmark{3}, JUNRUI ZHANG\authorrefmark{3}, QUANXI ZHAN\authorrefmark{3}, YONGCHUN WANG\authorrefmark{3}, AND JUNJIE ZHANG\authorrefmark{3}}
\affil{Shanghai Research Institute for Intelligent Autonomous Systems, Tongji University, Shanghai 201210, China}
\affil{State Key Laboratory of Autonomous Intelligent Unmanned Systems, Tongji University, Shanghai 201210, China}
\affil{College of Electronic and Information Engineering, Tongji University, Shanghai 201804, China}
\corresp{CORRESPONDING AUTHOR: RUNJIE SHEN (shenrunjie@tongji.edu.cn)}
\authornote{This work was supported by State Key Laboratory of Autonomous Intelligent Unmanned Systems, China.}

\begin{abstract}
Hydropower tunnel inspection is critical for infrastructure integrity yet remains 
inefficient and hazardous using manual methods. We propose FLISP (Fast LiDAR-IMU 
Synchronized Path Planner), a mapless planning framework for cooperative UGV-UAV 
inspection.  Unlike traditional map-based paradigms, FLISP features three core 
contributions: (1) a unified architecture where a single UGV-mounted LiDAR-IMU 
suite drives synchronized path generation for both platforms; (2) platform-specific 
solvers utilizing an enhanced Firefly Algorithm for UGV obstacle avoidance and a 
dynamic iterative optimizer for UAV flight; and (3) a hierarchical refinement 
strategy ensuring kinematic feasibility without state estimation drift. Benchmarks 
in a 1.2 km operational tunnel demonstrate that FLISP circumvents structural 
bottlenecks of map-based methods, eliminating map rasterization overhead (Fast-LIO2 + A*) 
and sampling instability (LIO-SAM + RRT*). FLISP achieves a 100\% success rate with 
$\sim$7 ms latency, representing a 7-fold speedup over grid-based and three-order-of-magnitude 
improvement over sampling-based baselines.  Validated in operational hydropower tunnels, 
this approach offers a scalable solution for robotic inspection in feature-degraded 
linear infrastructure.  A demonstration video is available at 
\url{https://youtu.be/Y_ezs1PfLJ4}, and the code at 
\url{https://github.com/ArchibaldGuo/FLISP.git}.
\end{abstract}



\maketitle

\section{INTRODUCTION}
\IEEEPARstart{L}{arge-scale} civil infrastructure, particularly hydropower water conveyance tunnels, constitutes a critical backbone for sustainable energy management. Ensuring the structural integrity of these kilometers-long facilities is paramount. However, current maintenance relies predominantly on manual inspection, a labor-intensive, hazardous process prone to subjective errors, thereby compromising long-term health assessments \cite{bib1, bib3, bib4, bib5, bib6}. The adverse conditions within these confined spaces, characterized by poor lighting, homogeneous textures, and uneven terrain, necessitate automated solutions to enhance safety and data quality.

\begin{figure*}
	\centerline{\includegraphics[width=6.2in]{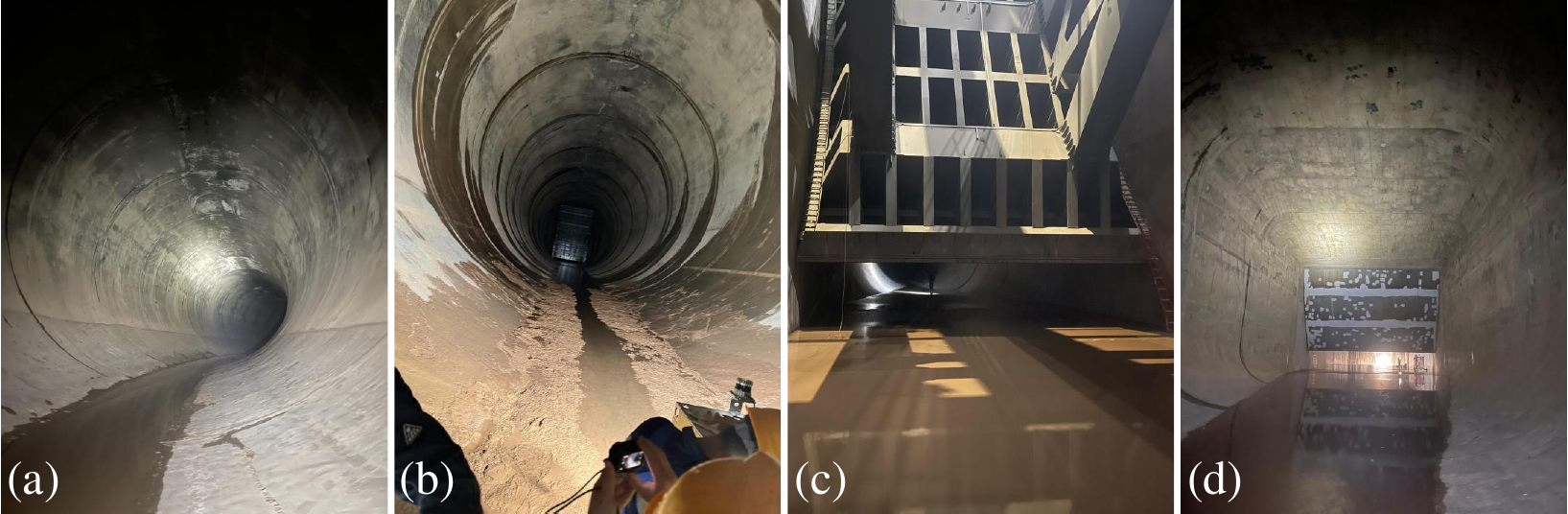}}
	\caption{The challenging operational environment of a large-scale hydropower tunnel. Insets show a typical textureless section (a), uneven silt accumulation (b), and structural obstacles like gates (c, d), underscoring the necessity for a collaborative UGV-UAV system.\label{fig0}}
\end{figure*}

Recent advancements in automation and robotics offer a transformative approach to this long-standing industry problem \cite{bib7, bib8, bib9}. While single-robot systems have shown promise, the scale and complexity of kilometer-long, high-arched tunnels render them ineffective for comprehensive inspection. An unmanned ground vehicle (UGV) cannot access the tunnel crown, while an aerial vehicle (UAV) is limited by endurance. This necessitates a collaborative, heterogeneous robotic system. However, deploying such systems in these GPS-denied, perceptually-degraded environments (Fig.~\ref{fig0}) faces significant automation challenges: robust localization over long distances with repetitive textures, stable navigation on uneven or debris-covered surfaces, and real-time, collision-free path planning for multiple agents.

Existing multi-robot planners, often derived from manufacturing or logistics, typically rely on structured environments or computationally expensive map-based frameworks (e.g., SLAM methods in the DARPA SubT Challenge \cite{bib49}). These approaches are ill-suited for the linear, repetitive geometry of tunnels, where maintaining a globally consistent map is both prone to drift and computationally prohibitive for routine inspection. Conversely, substituting global maps with purely reactive local planners avoids mapping overhead but renders the system highly susceptible to topological deadlocks and kinematic inconsistencies in blind curves (detailed in Appendix A).

To bridge this gap, we introduce the Fast LiDAR-IMU Synchronized Path Planner (FLISP), a novel framework for robust UGV-UAV tunnel navigation. Diverging from map-centric paradigms, FLISP employs a lightweight, ``mapless'' strategy. It leverages local geometry from a single UGV-mounted LiDAR-IMU suite to generate synchronized, collision-free paths for the entire team in real time. This approach significantly reduces computational load and eliminates SLAM-inherent drift accumulation. We validate FLISP through extensive simulations and a full-scale deployment in a 1.2 km hydropower tunnel. Comparative analysis reveals the limitations of traditional map-based methods and highlights our approach's orders-of-magnitude efficiency gain.

Distinct from loosely coupled multi-robot systems, FLISP establishes a vertically integrated, asymmetric collaboration model. Instead of isolated planning, we centrally process environmental geometry on the UGV. This creates a hierarchical dependency where the ground-based perception layer directly drives lightweight aerial path generation. Consequently, the system functions as a unified pipeline, extending the robustness of the ``mapless'' philosophy from state estimation down to multi-agent control.

Our primary contributions are:
\begin{enumerate}
	\item A hierarchical navigation framework using a UGV-centric, asymmetric collaboration model, enabling real-time planning for a UGV-UAV team using a single sensor suite in degenerate environments.
	\item Platform-specific, multi-objective obstacle avoidance strategies tailored to the unique challenges of constructed environments, such as a UGV navigating on uneven, curved floors and a UAV maneuvering around fixed infrastructure like sluice gates.
	\item A multi-level optimization mechanism that ensures path smoothness and computational efficiency, demonstrating a robust pipeline from raw sensor data to executable trajectories.
\end{enumerate}

To facilitate benchmarking, we release the source code and a curated representative LiDAR-IMU dataset capturing critical geometric degeneracy in tunnels. Resources are available at: \url{https://github.com/ArchibaldGuo/FLISP.git}.

\section{RELATED WORK}

\subsection{CLASSIC PATH PLANNING ALGORITHMS}

Path planning algorithms are broadly categorized into global paradigms, reliant on prior environmental models (GIS or static maps), and local paradigms driven by real-time perception \cite{bib24, bib25}. Global planning optimizes reference trajectories against topological, energy, and safety constraints \cite{bib26}. Conversely, local planning focuses on online trajectory generation, utilizing sensor fusion for dynamic obstacle avoidance and kinematic feasibility \cite{bib27}.

Classical global algorithms include A* \cite{bib28}, Dijkstra \cite{bib29}, and RRT \cite{bib30}, while local approaches encompass Artificial Potential Fields (APF) \cite{bib31}, Dynamic Window Approach (DWA) \cite{bib32}, and deep learning techniques \cite{bib33}. In the specific context of penstock and tunnel inspection, Özaslan et al. \cite{bib_ozaslan} pioneered a seminal framework for autonomous UAV navigation. Their work established a critical methodology for standalone Micro Aerial Vehicles (MAV), demonstrating how onboard 3D mapping can effectively support localization and planning in these confined environments. While this map-centric approach is essential for single-agent operations, FLISP introduces a collaborative UGV-UAV framework to address endurance limits and adopt a ``mapless'' paradigm to mitigate the catastrophic mapping failures inherent to feature-degraded tunnels (as demonstrated in Section VI-C).

Contemporary research increasingly integrates heuristic strategies to enhance robustness. For instance, HA-RRT \cite{bib34} incorporates adaptive mechanisms to resolve RRT's slow convergence in maritime contexts. Similarly, hybrid algorithms combining honey badger optimization with cubic splines \cite{bib35}, and improved Q-learning with dynamic exploration \cite{bib36}, have demonstrated superior efficiency and convergence over traditional baselines.

\subsection{PATH PLANNING FOR MULTI-ROBOT SYSTEMS}

In recent years, research on path planning for multi-robot systems has increasingly focused on learning-based approaches to handle complex coordination. For instance, reinforcement learning frameworks, such as the Soft Actor-Critic (SAC) algorithm for UAVs \cite{bib37} and Multi-Agent Proximal Policy Optimization (MAPPO) with experience replay \cite{bib39}, have been utilized to optimize obstacle avoidance and goal achievement. Similarly, imitation learning has been employed to enable agents to rapidly acquire complex behaviors from expert demonstrations \cite{bib38}. 

In parallel with learning-based paradigms, traditional heuristic scheduling approaches, such as hybrid-state A* combined with conflict-based search \cite{bib40}, have been developed for multiple UGVs in warehouse logistics. However, these approaches predominantly target homogeneous agents and are strictly trained or evaluated in simulated, well-structured grid environments. Consequently, they struggle to generalize to the non-convex, unmapped geometries and distinct kinematic constraints inherent to subterranean tunnels.

\subsection{PATH PLANNING FOR HETEROGENEOUS MULTI-ROBOT SYSTEMS}

For aerial-ground heterogeneous multi-robot systems, combining UAV and UGV capabilities significantly enhances joint mapping and planning. UAVs leverage their elevated vantage point for dynamic local mapping, which serves as the foundation for subsequent planning tasks \cite{bib41}. Typically, UAV vision systems capture sequential imagery that undergoes rectification, stitching, and obstacle recognition to construct a global ground map. Concurrently, UGVs employ 2D LiDAR to generate local feature sets. Based on these integrated maps, global and local planning are executed using A* and DWA algorithms, respectively. Similarly, Niu et al. \cite{bib42} employ a UAV-mounted wide-angle camera to localize multiple UGVs and obstacles, proposing a concave-convex programming algorithm to solve non-convex energy minimization problems for optimal UGV path generation.

To address synchronization challenges, various architectures introduce intermediary agents or centralized nodes. Hu et al. \cite{bib43} project 3D LiDAR data onto 2D occupancy grids constructed via UGV SLAM. They utilize an enhanced Long Short-Term Memory (LSTM) model for path planning, employing a mediator agent to relay data and resolve conflicts between platforms. Li et al. \cite{bib44} implement a centralized data processing center acting as a relay hub to facilitate seamless aerial-ground coordination. 

Sharing a perception strategy similar to our work, Martinez-Rozas et al. \cite{bib45} utilize LiDAR-generated local maps and an optimal RRT* method to simultaneously plan trajectories for a UAV, a UGV, and their connecting tether. However, the computational overhead of solving tether optimization equations significantly restricts performance. In field experiments, this resulted in stepwise movement patterns, highlighting the difficulty of achieving real-time agility in complex environments due to algorithmic complexity.

\subsection{METHODS IN TUNNEL ENVIRONMENTS}

Standard planning algorithms struggle in large-scale tunnels due to uneven terrain, GNSS denial, and geometric uniformity. In our previous work, we explored alternative methods with mixed success. Zhang et al. \cite{bib46} utilized optical flow to mitigate UAV drift; however, this approach proves unreliable in low-light or curved sections and fails to address UGV attitude estimation. Similarly, Zhan et al. \cite{bib47} implemented a parametric cylindrical fitting methodology via Levenberg-Marquardt optimization for incomplete point clouds. While geometrically accurate, its iterative nature incurs prohibitive computational costs for simultaneous multi-vehicle planning, and it neglects the critical roll-constraints of ground vehicles.

Research on synchronous UGV-UAV tunnel planning remains scarce. Existing collaborative frameworks \cite{bib48} often rely on the UGV maintaining a global SLAM map to georeference the UAV. This dependency creates a single point of failure: if SLAM drifts due to featureless geometry, the entire system collapses.

Furthermore, state-of-the-art subterranean systems from the DARPA SubT Challenge \cite{bib49, bib_costar, bib_step, bib_csiro_frontier} primarily target the exploration of complex networks via computationally intensive SLAM. Even frontier-based approaches that successfully bypass dense volumetric mapping \cite{bib_csiro_frontier} assume flat traversable surfaces, lacking the kinematic constraints essential for heavy UGVs navigating curved tunnel floors. Similarly, recent single-platform studies, such as those optimizing UGV navigation in mines \cite{bib51, bib50} or extracting UAV centerlines via occupancy grids \cite{bib52} and vision \cite{bib6}, fail to address heterogeneous team synchronization.

To summarize, existing methods typically depend on drift-prone SLAM, assume flat ground traversability, or isolate platforms. FLISP bridges this gap by enabling simultaneous, mapless UAV-UGV path planning with platform-specific kinematic constraints tailored for geometrically degraded linear infrastructure.

\section{SYSTEM ARCHITECTURE}

\subsection{FRAMEWORK AND CONTROL FLOW}

\begin{figure*}
	\centerline{\includegraphics[width=4.7in]{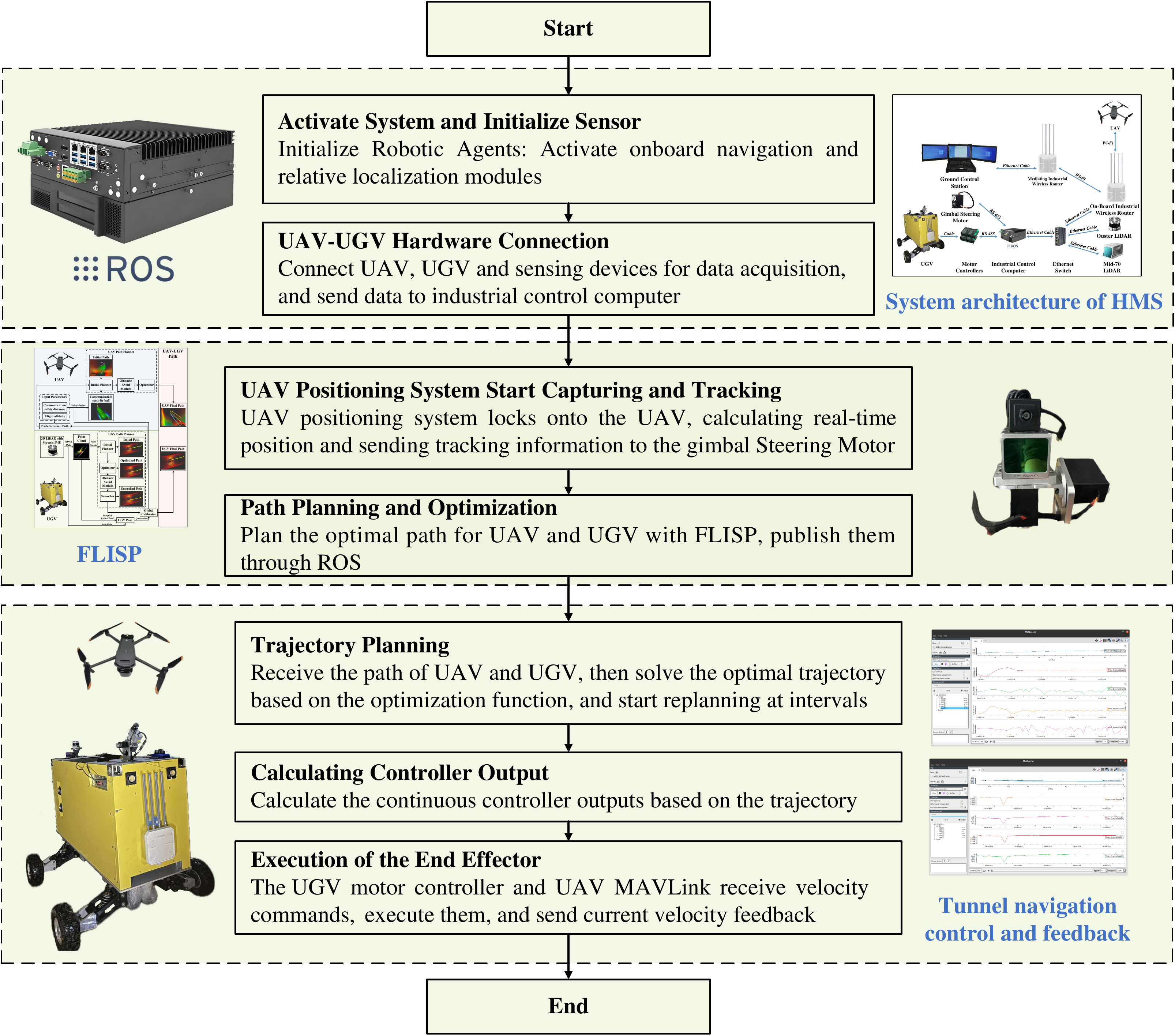}}
	\caption{Operational workflow of the UAV-UGV collaborative system. The architecture comprises three principal modules: (1) Hardware initialization, (2) Cooperative path planning, and (3) Motion control  with feedback mechanisms. \label{fig1}}
\end{figure*}

Fig.~\ref{fig1} illustrates the overall system architecture. Following initialization, a gimbal-mounted tracking suite (LiDAR and RGB camera) monitors the UAV's relative position. Simultaneously, a navigation LiDAR-IMU on the UGV provides environmental point clouds and attitude data. The UGV's path is generated via dynamic binning and multimodal fitting, switching to a Firefly Algorithm (FA) for local avoidance upon obstacle detection. The UAV path is then derived hierarchically from the UGV path, adhering to altitude and communication safety constraints.

A cascaded control architecture executes these trajectories: outer-loop controllers transmit desired states to inner-loop actuators, while continuous feedback from UGV local odometry (i.e., wheel encoders and IMU) and UAV positioning closes the loop. It should be explicitly stated that while FLISP does not require a globally consistent map, it fundamentally relies on accurate local odometry from the UGV platform for short-term trajectory tracking.

\subsection{UAV AND UGV HARDWARE}

\begin{figure}
	\centerline{\includegraphics[width=2.5in]{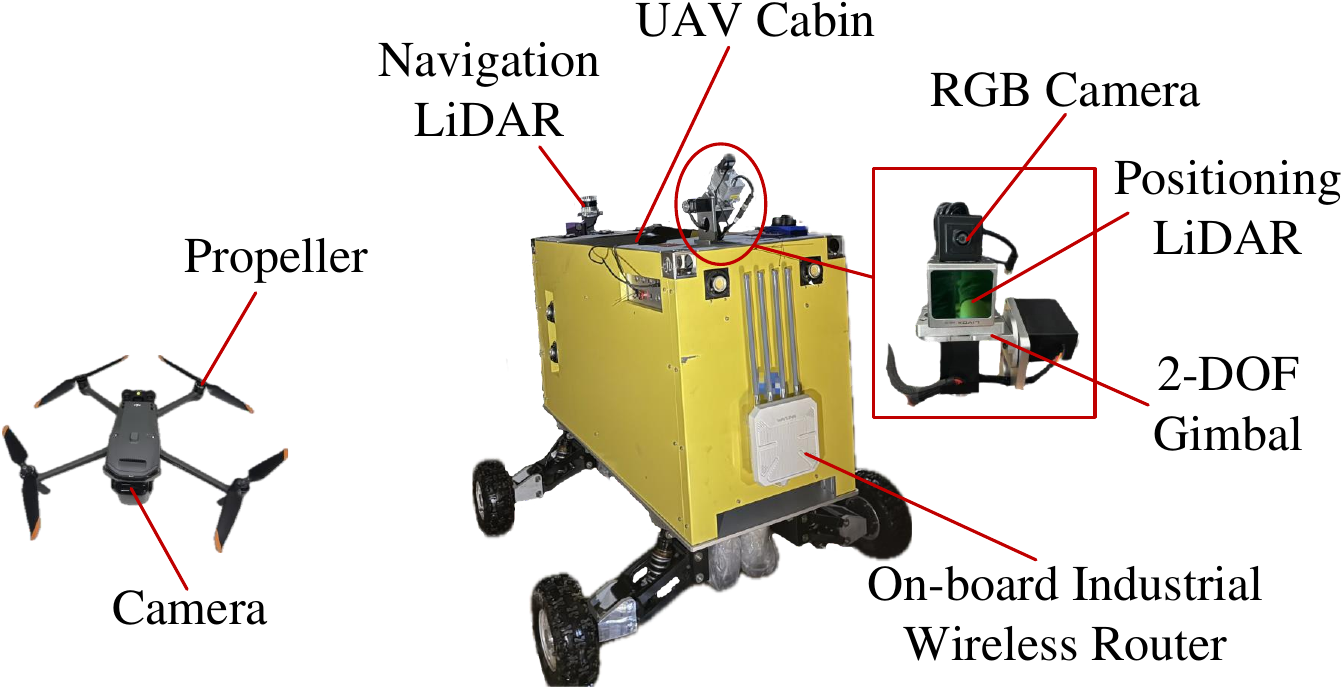}}
	\caption{The hardware configuration comprising a UGV and an inspection UAV. The front gimbal is deactivated for this study.\label{fig2}}
\end{figure}

Fig.~\ref{fig2} depicts the hardware configuration. The aerial platform is a DJI Mavic 3T, utilizing COTS controllers for low-level stabilization to offload trajectory generation. The custom UGV features a GPU-accelerated industrial computer processing data from a 64-line navigation LiDAR-IMU, a UAV tracking module (comprising a separate, short-range LiDAR and a high-resolution monocular RGB camera on a two-degree-of-freedom gimbal), and wireless infrastructure. Crucially, to accommodate the narrow clearance during vertical hoisting into the tunnel, the UGV employs a heavy-duty chassis with independently retractable wheel arms. This mechanical design, while robust, results in a low steering response bandwidth that precludes rapid, large-angle rotations.

The navigation LiDAR, which contains an integrated IMU, is mounted on the front-top of the UGV to maximize lookahead for curvature detection. While this creates a near-field blind spot, it poses negligible impact on FLISP. The planner extracts the global tunnel centerline via geometric fitting rather than relying on immediate terrain analysis. Furthermore, obstacle detection volumes (Section IV-A) are calibrated to effectively ignore this unmonitored region. The intrinsic integration of the IMU within the LiDAR assembly ensures synchronized state estimation. Operationally, the UGV serves as the primary surveyor and mobile base, while the UAV inspects the tunnel crown and structural anomalies inaccessible from the ground.

\subsection{COMMUNICATION TOPOLOGY}

\begin{figure}
	\centerline{\includegraphics[width=3.3in]{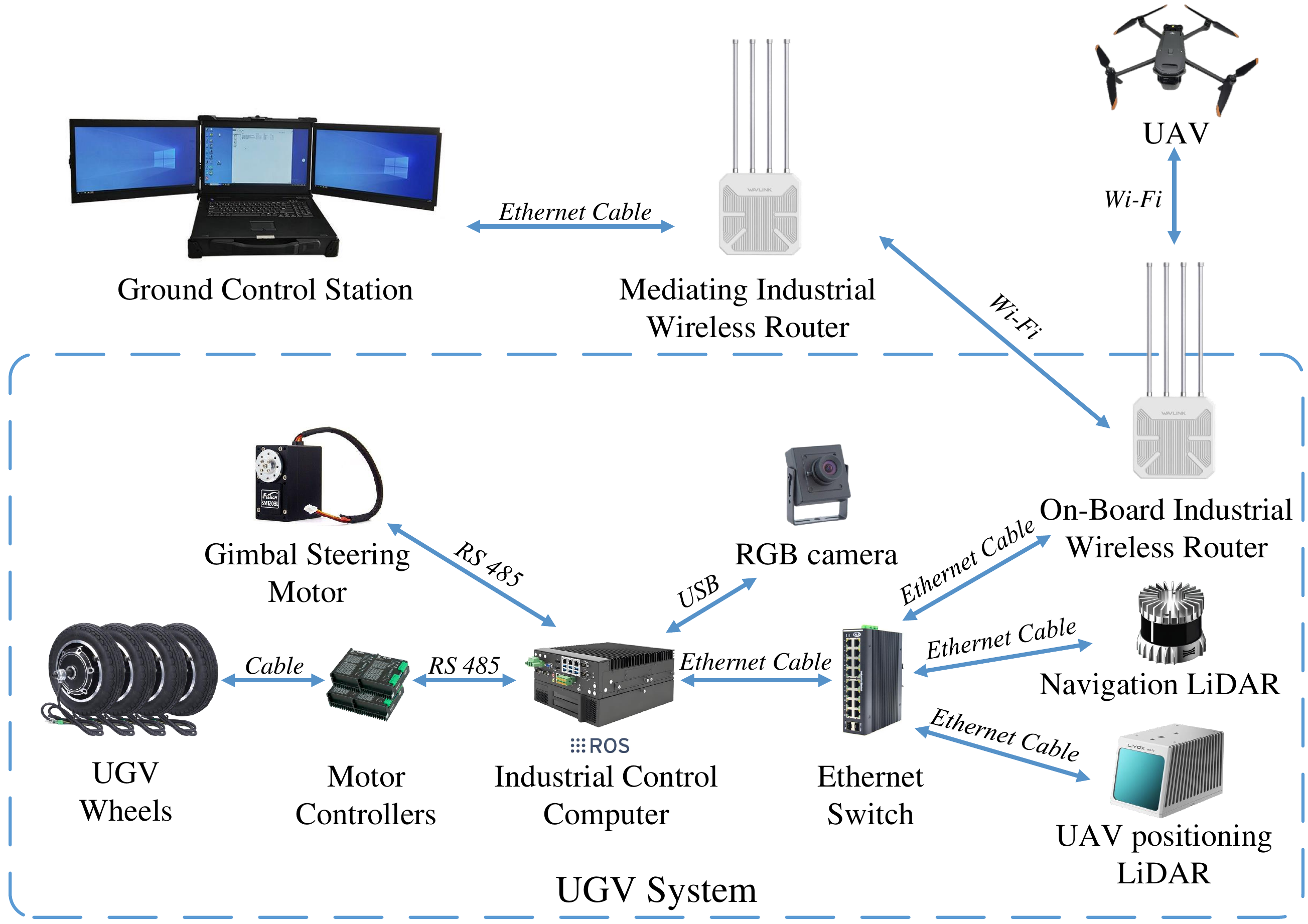}}
	\caption{Schematic of the communication structure. The architecture establishes a UGV-centric mobile network, linking the Ground Control Station, UGV, and UAV via a wireless bridge and local Wi-Fi. \label{fig3}}
\end{figure}

Fig.~\ref{fig3} illustrates the communication architecture, designed as a UGV-centric mobile network utilizing a wireless bridge for long-range connectivity. Specifically, the UGV's onboard industrial router establishes a local Wi-Fi zone for the UAV while maintaining a wireless link to a static mediating router at the tunnel entrance, which connects to the Ground Control Station via Ethernet. This dual-router configuration ensures robust, real-time data transmission over extended operational distances. Internally, the UGV computer functions as the integration hub: high-bandwidth sensors (LiDARs) and the router connect via Ethernet, while motion controllers and vision peripherals interface via RS-485 and USB, respectively.

\subsection{OPERATIONAL PHILOSOPHY}
FLISP is designed based on an ``Online Collection + Offline Reconstruction'' strategy. We adopt this approach because real-time SLAM algorithms frequently suffer from drift in featureless tunnels, which significantly compromises the accuracy of online defect localization. Instead, we utilize a gimbal-mounted camera that captures images at a fixed frequency while rotating. By integrating this fixed visual sampling rate with the UGV's velocity, we perform 3D reconstruction offline. Crucially, this reconstruction logic necessitates a strictly constant velocity and a jitter-free trajectory to prevent alignment artifacts. Therefore, our algorithmic focus is explicitly placed on optimizing the spatial path geometry to ensure the smooth motion required for high-fidelity data acquisition, as evidenced by the complete tunnel model shown in Fig.~\ref{fig17}.

\section{FAST LIDAR-IMU SYNCHRONIZED PATH PLANNER}

\begin{figure*}
	\centerline{\includegraphics[width=6.0in]{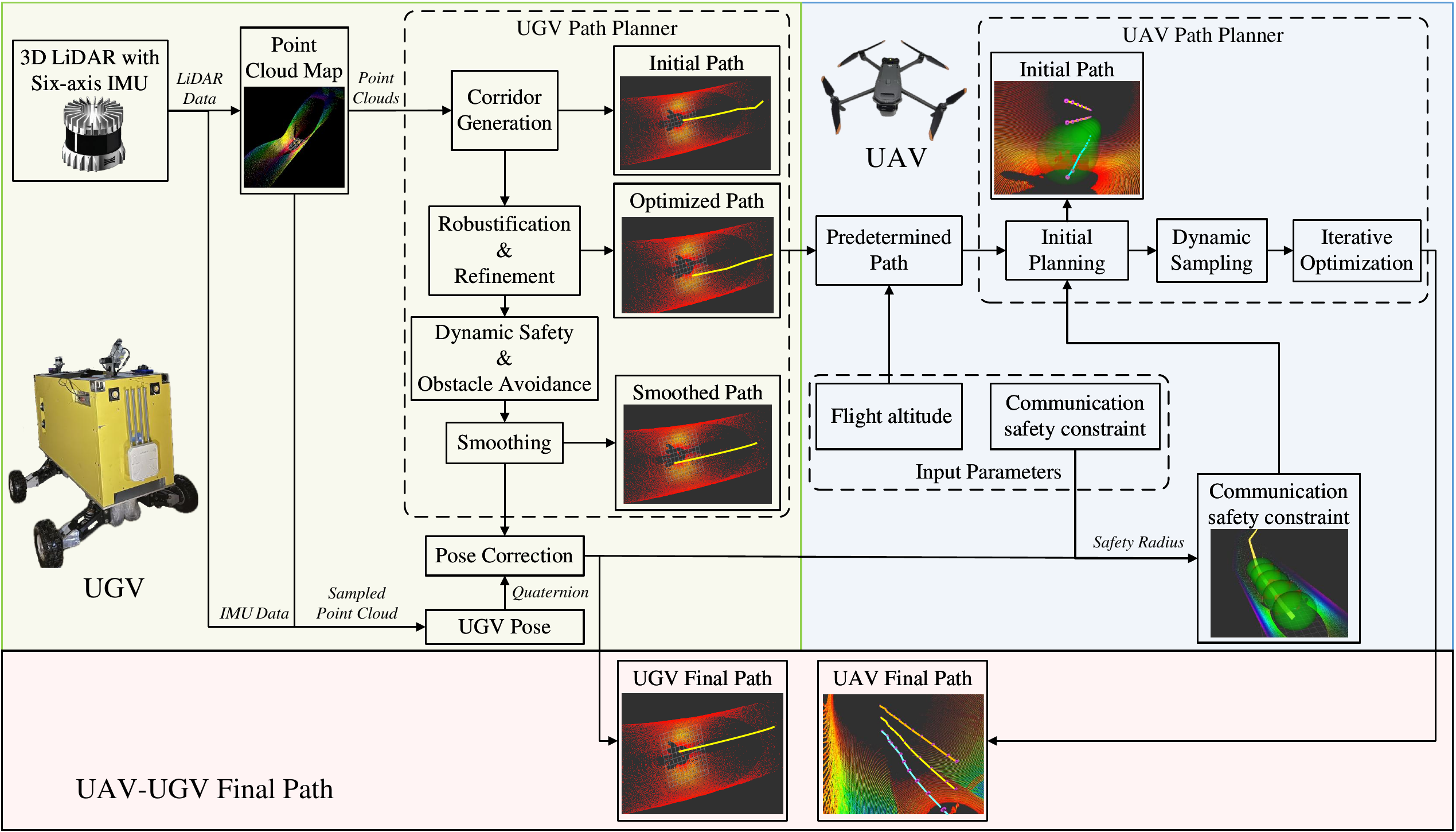}}
	\caption{Overview of FLISP. Starting from the LiDAR with the IMU, the initial path of the UGV is planned and optimized, and then the UAV path is planned and optimized based on the communication safety constraint established by the UGV path.\label{fig4}}
\end{figure*}

This section details FLISP, designed for GNSS-denied tunnel environments. FLISP integrates three synchronized subsystems: (1) a UGV-based state estimation and perception layer that serves as the computational core; (2) a hierarchical planner that generates dynamically feasible trajectories for the UAV; and (3) a robust control interface that handles non-linear flight dynamics. As illustrated in Fig.~\ref{fig4}, this architecture ensures real-time responsiveness by decoupling heavy perception tasks from the lightweight execution layer of the UAV. The subsystems are linked via a synchronized state machine, ensuring that the UAV's planning is continuously updated by the UGV's high-fidelity perception without latency-inducing global map consistency checks.

For this work, we used only a 3D LiDAR and IMU. This design choice is driven by three factors: First, LiDAR provides reliable geometric data in the low-light conditions typical of tunnels, where visual sensors often underperform. Second, direct geometric measurement is robust to the textureless tunnel surfaces that frequently cause failure in visual SLAM algorithms. Third, avoiding dense global mapping mitigates errors caused by structural homogeneity. Adding further sensors would increase system complexity and computational overhead, which aligns with the efficiency requirements of the proposed planner.

\subsection{Hierarchical UGV Path Planner}
The UGV path planner comprises three stages: corridor generation, refinement, and dynamic constraints/safety constraints (smoothing).

\subsubsection{Foundational Corridor Generation}
The first stage aims to rapidly extract the tunnel's fundamental geometric structure from raw sensor data, establishing a coarse but directionally correct corridor for navigation. This involves two key steps: estimating the vehicle's orientation within the tunnel and deriving an initial centerline path.

Since six-axis IMUs lack absolute yaw information in magnetically complex environments, we first estimate the relative yaw angle to the tunnel walls, which is integrated with IMU data to derive the complete vehicle pose. Following \cite{bib46}, we utilize a point cloud sampling technique (Fig.~\ref{fig5}) to derive a robust tunnel wall normal vector $\overrightarrow v$. The yaw angle deviation $\theta$ is then calculated:
\begin{figure}
	\centerline{%
		\begin{tabular}{cc}
			\includegraphics[width=1.7in]{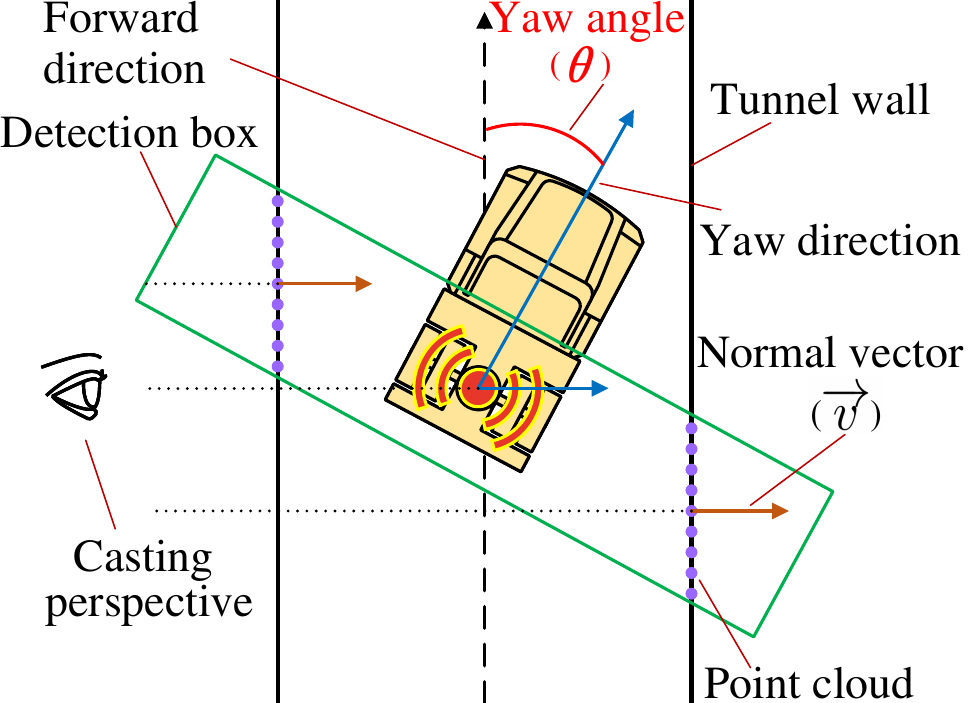} & \includegraphics[width=1.7in]{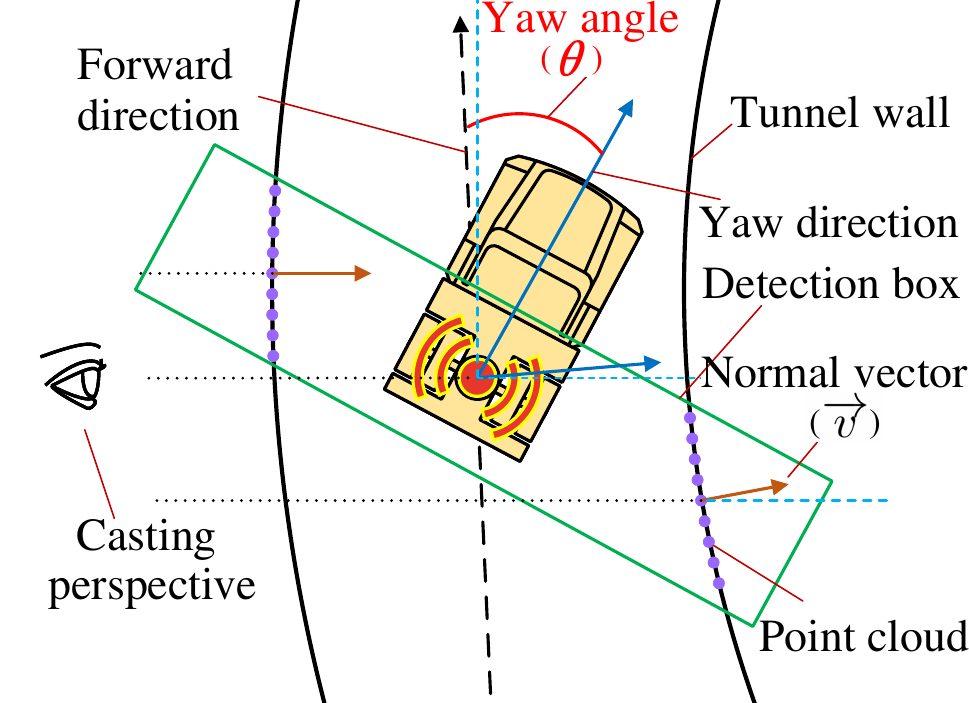} \\
			(a) & (b)
		\end{tabular}
	}
	\caption{Yaw angle estimation in tunnel environments. (a) Straight section with surface normal detection. (b) Curved section with wall normal extraction.\label{fig5}}
\end{figure}
\begin{equation} \label{eq:1} \theta = \pi - \arccos \left( {\left[ {\begin{array}{*{20}{c}} 1\ 0\ 0 \end{array}} \right]^T \cdot \overrightarrow v } \right) \end{equation}
This angle is then fused into the vehicle's attitude quaternion to achieve a comprehensive state estimation.
\begin{equation} \label{eq:2} {q_d} = {q_b} \otimes \left[ {\begin{array}{*{20}{c}} {\cos (\theta /2)}\ 0\ 0\ {\sin (\theta /2)} \end{array}} \right]^T \end{equation}

\begin{figure}
	\centerline{%
		\begin{tabular}{cc}
			\includegraphics[width=1.6in]{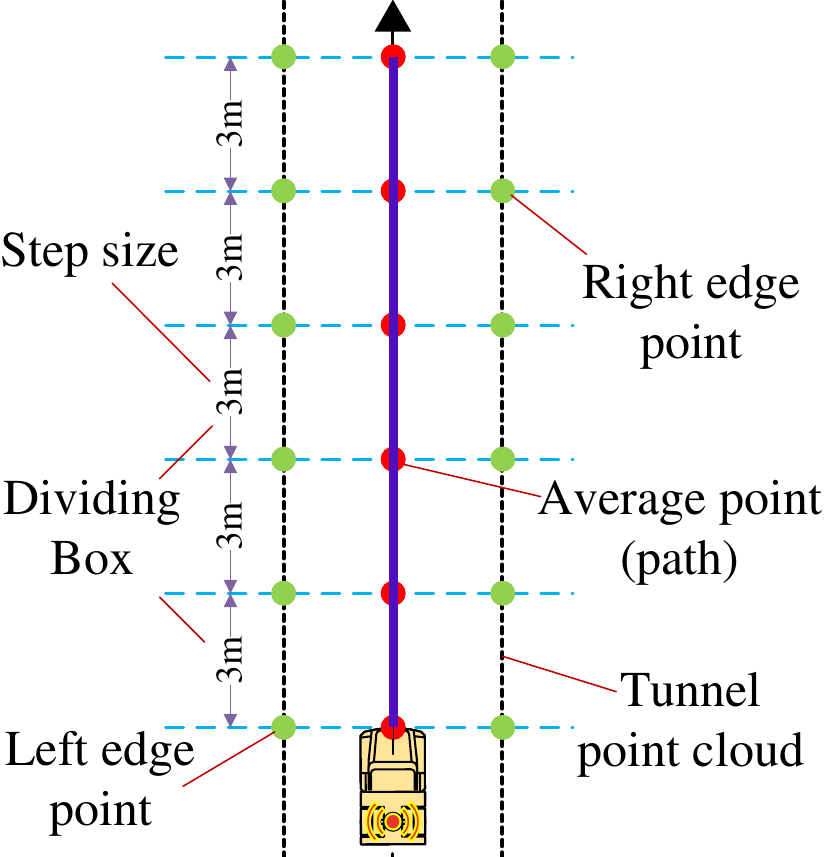} & \includegraphics[width=1.6in]{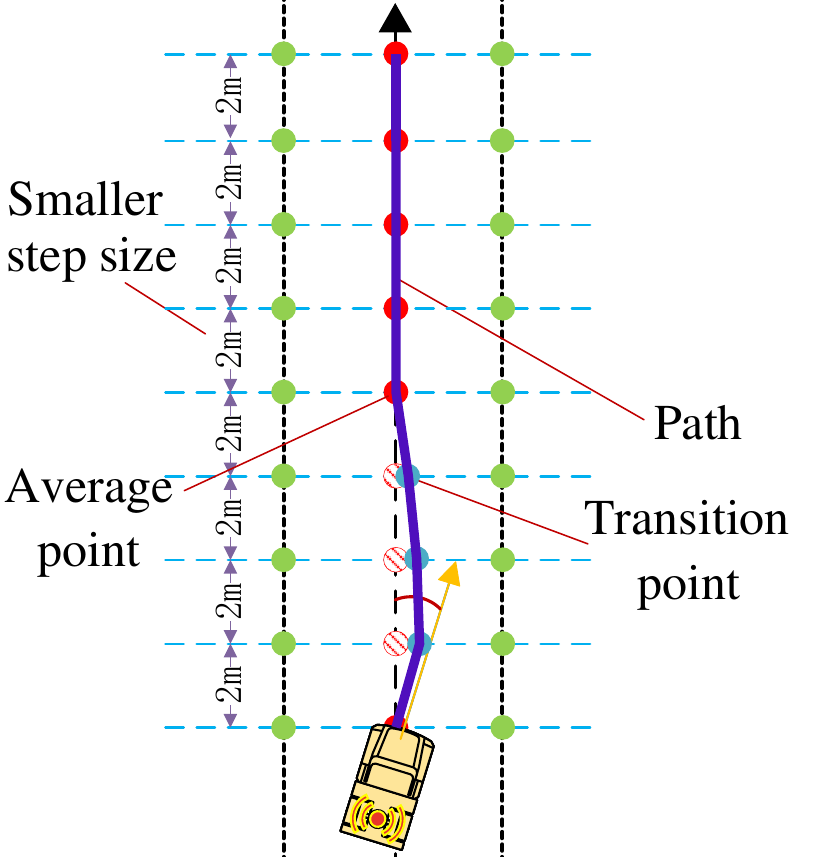} \\
			(a) & (b)
		\end{tabular}
	}
	\caption{Initial path planning visualization. (a) Straight tunnel with uniform step size. (b) Yaw angle presence resulting in finer path resolution with additional transition points.\label{fig6}}
\end{figure}

With vehicle orientation established, the planner generates a centerline path using a binning strategy with a dynamically adjusted step size $\Delta s$, governed by yaw angle $\theta$ (Eq.~\eqref{eq:3}). This allows finer resolution during turns and faster planning in straight sections (Fig.~\ref{fig6}).
\begin{equation} \label{eq:3}
	\Delta s =
	\begin{cases}
		c_1 \cdot \left(1 - \dfrac{|\theta|}{c_2}\right)^{\kappa} + \lambda |\theta|^3, & |\theta| \leqslant c_2 \\
		c_3 + \mu (|\theta| - c_2),                & |\theta| > c_2
	\end{cases}
\end{equation}

The point cloud is first partitioned into bins based on step size:
\begin{equation} \label{eq:4}
	B_j = \{(x_i, y_i, z_i) \in \mathcal{P} | (j-1)\Delta s \leq x_i < j\Delta s\}
\end{equation}
Within each bin, the leftmost and rightmost boundary points are extracted.

\begin{equation} \label{eq:5}
	\begin{aligned}
		p_j^L &= \arg\min_{(x, y, z) \in B_j} y \\
		p_j^R &= \arg\max_{(x, y, z) \in B_j} y
	\end{aligned}
\end{equation}

\begin{figure}
	\begin{tabular}{cc}
		\includegraphics[width=1.7in]{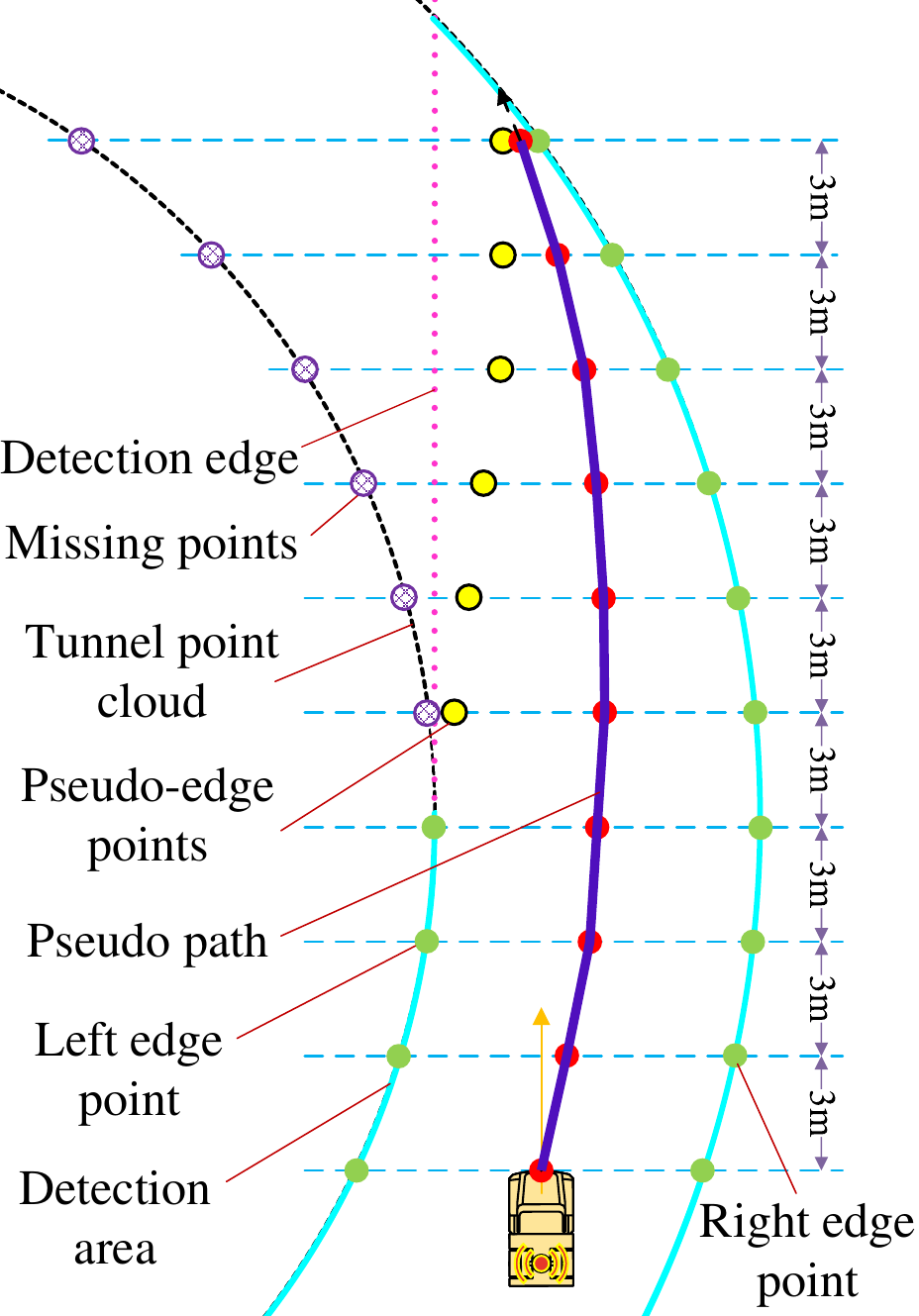} & \includegraphics[width=1.7in]{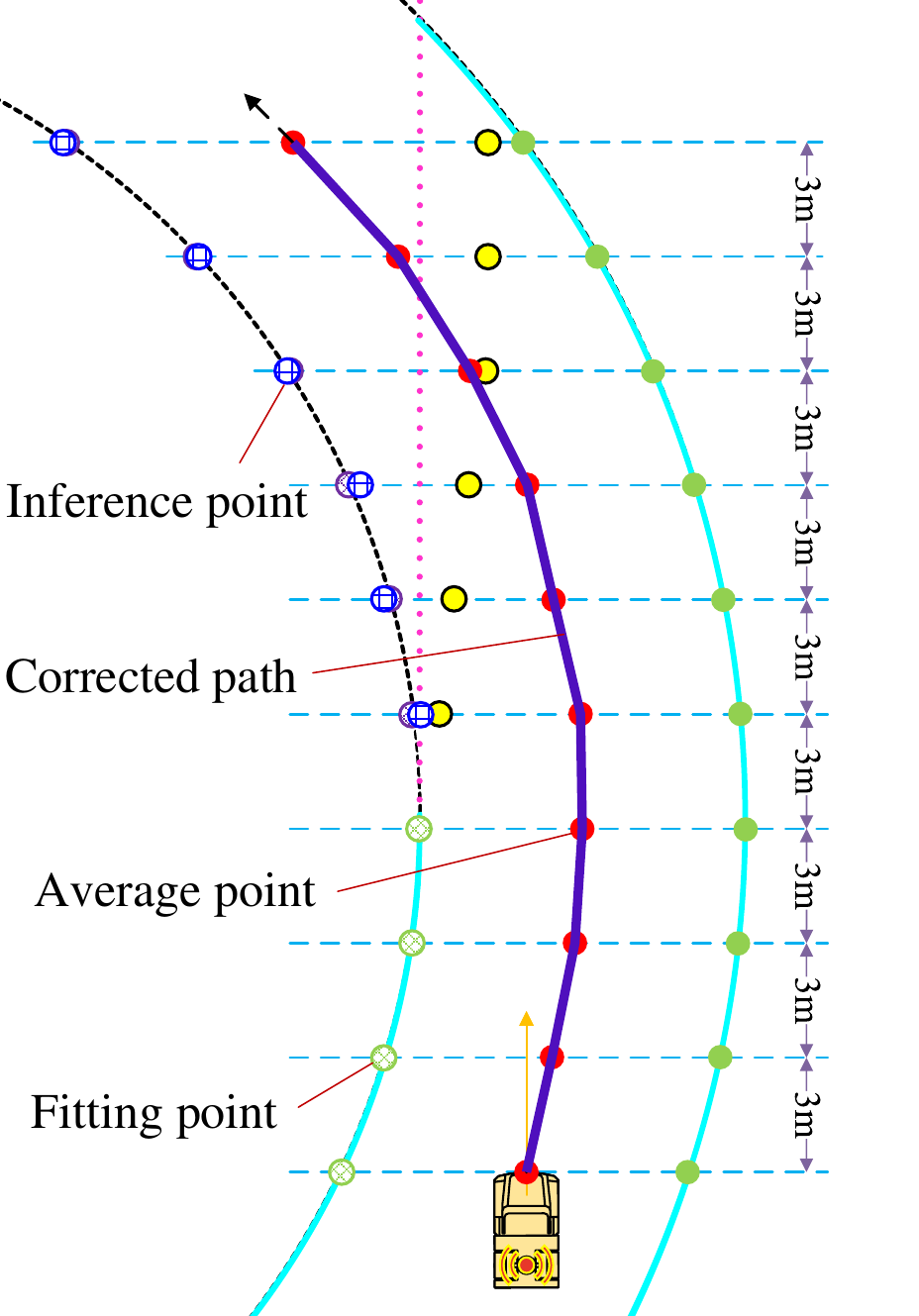} \\
		(a) & (b)
	\end{tabular}
	\caption{Curved tunnel path planning. (a) Occluded point clouds generating inaccurate pseudo-paths. (b) Corrected path through model-based fitting and inference.\label{fig7}}
\end{figure}

To handle LiDAR occlusion in curved sections, we analyze the local slope sequence $K$ to classify the segment as straight or curved, using its variance $\sigma^2_K$.
\begin{equation} \label{eq:6}
	K = \{k_i | k_i = \frac{y_{i+\delta} - y_i}{x_{i+\delta} - x_i}, i = 1, 2, ..., n-\delta\}
\end{equation}
\begin{equation} \label{eq:7}
	\sigma^2_K = \frac{1}{|K|}\sum_{k_i \in K}(k_i - \bar{k})^2
\end{equation}
A multi-level fitting model is then applied to infer the complete tunnel boundaries from sparse data (Fig.~\ref{fig7}).
\begin{equation} \label{eq:8}
	y =
	\begin{cases}
		\beta_0 + \beta_1 x, & \sigma^2_K \leqslant \tau_{\text{curve}} \\
		\sum_{j=0}^{d} \alpha_j x^j, & \sigma^2_K > \tau_{\text{curve}.}
	\end{cases}
\end{equation}

\noindent Crucially, this model-based inference inherently enhances robustness against environmental anomalies. Specifically, it bridges potential point cloud cavities caused by water reflection on the tunnel floor, ensuring path continuity even when raw LiDAR data is locally fragmented.

The initial UGV path, $P_{\mathrm{initial}}$, is finally calculated as the midpoint between these fitted boundaries, providing a foundational but potentially noisy centerline.
\begin{equation}\label{eq:9}
	P_{\mathrm{initial}} = \left\{
	\frac{p_i^L + p_i^R}{2} \;\middle|\; p_i^L \in \overline{\mathcal{E}}^{\,L}, \,
	p_i^R \in \overline{\mathcal{E}}^{\,R}, \,
	i = 1, 2, \dots
	\right\}.
\end{equation}

Extracting this long-horizon centerline is essential to accommodate the UGV's aforementioned mechanical constraints. A myopic, short-horizon planner would command sudden steering inputs that the low-bandwidth wheel arms cannot execute in time, risking collisions. Instead, FLISP's long-horizon geometry provides an anticipatory yaw gradient, enabling the slow-response actuators to smoothly execute gradual micro-adjustments.

\subsubsection{Path Robustification and Refinement}
The initial path $P_{\mathrm{initial}}$ serves as a coarse blueprint but is susceptible to sensor noise. This second stage refines this path, enhancing its smoothness and reliability using a Bayesian inference approach.

First, for each point $p_i$, we construct a local prediction $\hat{y}_i$ of its position based on its neighbors, using both forward and backward linear interpolation.
\begin{subequations} \label{eq:10}
	\begin{align}
		\hat{y}_{i,\text{prev}} &= y_{i-1} + \frac{y_{i-1} - y_{i-2}}{x_{i-1} - x_{i-2}} \cdot (x_i - x_{i-1}) \label{eq:10a} \\
		\hat{y}_{i,\text{next}} &= y_{i+1} + \frac{y_{i+1} - y_{i+2}}{x_{i+1} - x_{i+2}} \cdot (x_i - x_{i+1}) \label{eq:10b}
	\end{align}
\end{subequations}
The final prediction is the average of these two values:
\begin{equation} \label{eq:11}
	\hat{y}_i = \frac{\hat{y}_{i,\text{prev}} + \hat{y}_{i,\text{next}}}{2}
\end{equation}
Next, we define the measurement error $\varepsilon_i$ as the difference between the actual and predicted values.
\begin{equation} \label{eq:12}
	\varepsilon_i = y_i - \hat{y}_i
\end{equation}
We model the error as following a Gaussian distribution, with a smaller standard deviation $\sigma_n$ for normal points and a larger one $\sigma_o$ for outliers.
\begin{subequations}\label{eq:13}
	\begin{align}
		p(\varepsilon_i|S_n) &= \frac{1}{\sigma_n\sqrt{2\pi}}e^{-\frac{\varepsilon_i^2}{2\sigma_n^2}} \label{eq:13a} \\
		p(\varepsilon_i|S_o) &= \frac{1}{\sigma_o\sqrt{2\pi}}e^{-\frac{\varepsilon_i^2}{2\sigma_o^2}} \label{eq:13b}
	\end{align}
\end{subequations}
Finally, according to Bayes' theorem, we compute the posterior probability that a point is an outlier given the observed error. Points exceeding a probability threshold are then corrected based on the prediction, as detailed in Algorithm~\ref{alg:bayesian_outlier}.
\begin{equation} \label{eq:14}
	P(S_o|\varepsilon_i) = \frac{p(\varepsilon_i|S_o)P(S_o)}{p(\varepsilon_i|S_n)P(S_n) + p(\varepsilon_i|S_o)P(S_o)}
\end{equation}

\begin{algorithm}[H]
	\caption{Bayesian Outlier Detection and Correction}
	\label{alg:bayesian_outlier}
	\begin{algorithmic}[1]
		\STATE \textbf{Input:} Path $P_{initial}$; prior $P(S_o)$; standard deviations $\sigma_n, \sigma_o$; threshold $\theta$
		\STATE \textbf{Output:} Corrected path $P_{initial}^{\prime}$
		\FOR{each point $p_i$ where $i \in [3, n-2]$}
		\STATE Compute forward prediction $\hat{y}_{i,\text{prev}}$ using \eqref{eq:10a}
		\STATE Compute backward prediction $\hat{y}_{i,\text{next}}$ using \eqref{eq:10b}
		\STATE Compute combined prediction $\hat{y}_i$ using \eqref{eq:11}
		\STATE Compute measurement error $\varepsilon_i$ using \eqref{eq:12}
		\STATE Compute likelihoods $p(\varepsilon_i|S_n)$ and $p(\varepsilon_i|S_o)$ using \eqref{eq:13a}, \eqref{eq:13b}
		\STATE Compute posterior probability $P(S_o|\varepsilon_i)$ using \eqref{eq:14}
		\IF{$P(S_o|\varepsilon_i) > \theta$}
		\STATE Correct the point: $p_i \gets (x_i, \hat{y}_i)$
		\ENDIF
		\ENDFOR
		\RETURN Corrected path $P_{initial}^{\prime}$
	\end{algorithmic}
\end{algorithm}

\begin{figure}
	\centerline{\includegraphics[width=2.5in]{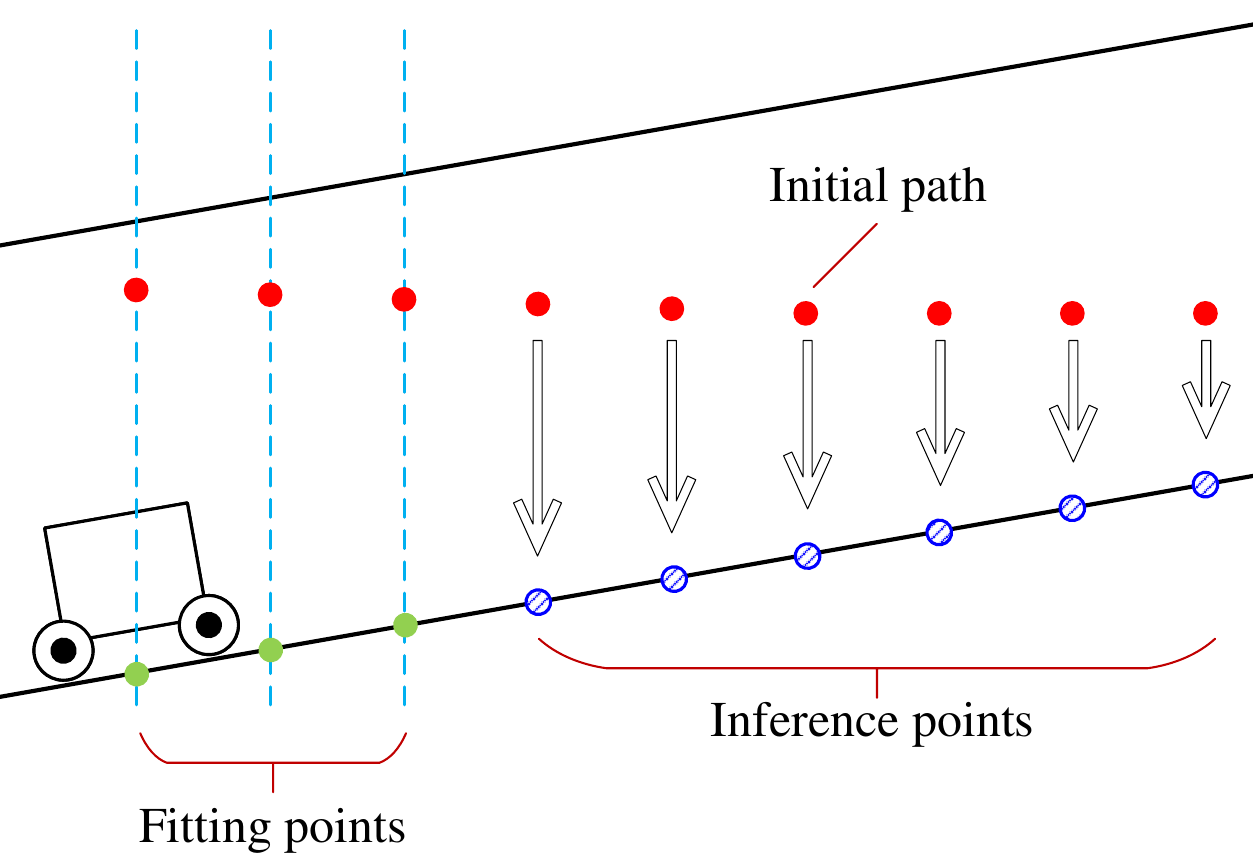}}
	\caption{Illustration of projecting path points onto the bottom of the tunnel.\label{fig8}}
\end{figure}

The corrected 2D path $P_{initial}^{\prime}$ is then projected onto the tunnel floor by fitting the elevation profile from the 3D point cloud, resulting in a 3D path, $P_{fit}$, that accurately follows the ground topology (Fig.~\ref{fig8}).

\subsubsection{Dynamic Safety and Obstacle Avoidance}
With a robust path $P_{fit}$, the third stage ensures safe passage by handling dynamic obstacles. FLISP adopts a path-centric detection strategy, constructing a safety corridor along the path (Fig.~\ref{fig8.5}) and checking for obstacles only within this volume. The detection box $B_i$ for each path segment is defined as a rotated volume in the global frame:

\begin{figure}
	\centerline{\includegraphics[width=3.3in]{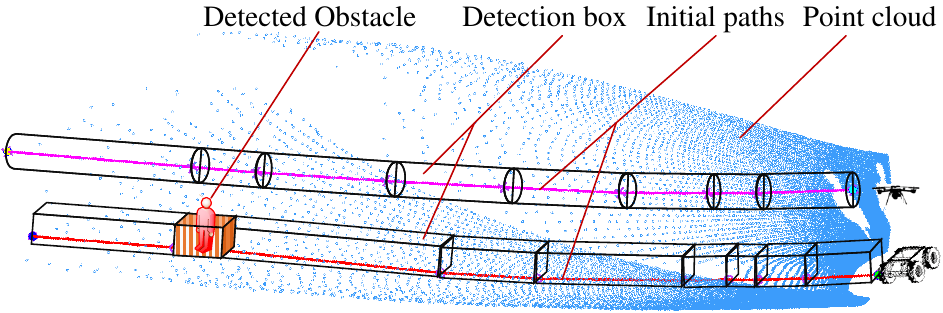}}
	\caption{FLISP's path-centric obstacle detection strategy. A safety corridor is constructed around the planned path. The corridor's shape is tailored to each platform: rectangular volumes for the UGV and cylindrical volumes for the UAV.\label{fig8.5}}
\end{figure}

\begin{equation} \label{eq:15}
	B_i = \left\{ \mathbf{p} \;\middle|\; \mathbf{R}_i^T (\mathbf{p} - \mathbf{c}_i) \in \mathcal{D}_i \right\}
\end{equation}

In this formulation, $\mathcal{D}_i$ represents the local, axis-aligned bounding box before rotation, whose dimensions are determined by the segment length $L_i$, a lateral safety radius $r$, and vertical limits $[z_{\min}, z_{\max}]$. It is defined as:

\begin{equation}\label{eq:16}
	\mathcal{D}_i = \left[ -\frac{L_i}{2}, \frac{L_i}{2} \right] \times [-r, r] \times [z_{\min}, z_{\max}]
\end{equation}

While Eqs. \eqref{eq:15} and \eqref{eq:16} establish the rigorous geometric criteria for safety, evaluating them via brute-force iteration over high-density LiDAR frames ($O(N \cdot M)$) introduces latency that threatens control stability. To guarantee robust operation within the 10 ms control cycle required for high-speed tracking, we implement a spatial partitioning strategy as a broad-phase filter.

The detection domain is discretized into a path-aligned 1D grid. LiDAR points are mapped to these cells via fast integer indexing ($O(1)$). Only points in path-occupied cells undergo the precise geometric containment check in Eq. \eqref{eq:15} (narrow-phase). Crucially, this optimization preserves geometric accuracy while reducing computational load for viable onboard processing.

Upon detecting an obstacle, an enhanced FA optimizes a local avoidance path (Algorithm~\ref{alg:firefly_tunnel}). To further accelerate convergence, we simplify the optimization problem from full 3D space to a 1D lateral search within the tunnel cross-section. The search is governed by two critical constraints. First, the path must remain within the physical boundaries of the tunnel, defined by:

\begin{subequations}\label{eq:17}
	\begin{align}
		y_{min} &= -R_{\text{pipe}} + \frac{W_{\text{v}}}{2} + M_s \label{eq:17a} \\
		y_{max} &= R_{\text{pipe}} - \frac{W_{\text{v}}}{2} - M_s \label{eq:17b}
	\end{align}
\end{subequations}

where $R_{\text{pipe}}$ denotes the tunnel radius, $W_{\text{v}}$ is the vehicle width, and $M_s$ represents a mandatory safety buffer. Second, to prevent the UGV from rolling over on the curved floor, a maximum inclination angle constraint is imposed:

\begin{equation}\label{eq:18}
	|\frac{y}{R_{\text{pipe}}}| \leq \theta_{max}
\end{equation}

Within these constraints, the quality of a potential path point is evaluated by its brightness $I(y)$. This multi-objective cost function rewards obstacle clearance while penalizing center deviation and excessive tilt:

\begin{equation}\label{eq:19}
	I(y) = d_{obstacle} - w_{\text{tilt}} \cdot \left(\frac{|y|}{R_{\text{pipe}}}\right)^2 - w_{\text{center}} \cdot |y|
\end{equation}

Unlike standard FA, our approach introduces a dynamically-weighted cost function where the penalty for tilt ($w_{\text{tilt}}$) increases non-linearly as the UGV approaches the wall. This adaptation creates a ``soft barrier'' against rollovers, a unique challenge in circular geometries.

Based on this brightness, fireflies update their positions using an attraction-exploration rule:
\begin{align}\label{eq:20}
	y_i^{(t+1)} =\; & y_i^{(t)} + \beta_0 e^{-\gamma r_{ij}^2} \left( y_j^{(t)} - y_i^{(t)} \right) \notag \\
	& + \alpha^{(t)} \cdot \mathrm{rand}(-0.5, 0.5) \cdot (y_{max} - y_{min})
\end{align}
Here, $\beta_0$ and $\gamma$ control the attractiveness, while the randomization parameter $\alpha^{(t)}$ decays geometrically over iterations (e.g., $\alpha^{(t+1)} = 0.95\alpha^{(t)}$). This decay mechanism allows broad exploration in early stages while ensuring stable convergence to the global optimum in the final steps.

\begin{algorithm}[H]
	\caption{Firefly Algorithm for Obstacle Avoidance}
	\label{alg:firefly_tunnel}
	\begin{algorithmic}[1] 
		\STATE \textbf{Input:} FA parameters ($N, T, \dots$); Tunnel parameters ($y_{\min}, y_{\max}, \dots$)
		\STATE \textbf{Output:} Optimized collision-free path $P_{\text{cf}}^*$
		\STATE \textbf{Initialize} firefly positions $y_i$ randomly within $[y_{\min}, y_{\max}]$
		\FOR{$t = 1$ \TO $T$}
		\STATE Calculate brightness $I(y_i)$ for all fireflies using Eq.~\eqref{eq:19}
		\FOR{$i = 1$ \TO $N$}
		\FOR{$j = 1$ \TO $N$}
		\IF{$I(y_j) > I(y_i)$}
		\STATE Update firefly position $y_i$ using Eq.~\eqref{eq:20}
		\STATE Enforce boundary constraint on $y_i$ using Eq.~\eqref{eq:17}
		\STATE Enforce inclination constraint on $y_i$ using Eq.~\eqref{eq:18}
		\ENDIF
		\ENDFOR
		\ENDFOR
		\STATE Decay random factor: $\alpha^{(t+1)} \gets \alpha^{(t)} \cdot \rho$
		\ENDFOR
		\STATE $y_{\text{best}} \gets \arg\max_{y_i} I(y_i)$
		\STATE Apply smoothers and prune redundant points from the path.
		\STATE \textbf{return:} Optimized path $P_{\text{cf}}^*$
	\end{algorithmic}
\end{algorithm}

\subsubsection{Final Path Smoothing and Pose Correction}
The final stage transforms the path $P_{\text{cf}}^*$ into a smooth, executable path. To mitigate rollover risk, FLISP applies a pose correction mechanism, first attenuating measured angles using a shrinkage factor (Eq.~\eqref{eq:21}). 
\begin{equation}\label{eq:21}
	\theta_{\text{adjusted}} = \theta_{\text{original}} \cdot f_{\text{shrink}}
\end{equation}
An adaptive number of transition points, $n_{\text{transition}}$, is calculated based on the yaw angle's magnitude (Eq.~\eqref{eq:22}).
\begin{equation}\label{eq:22}
	n_{\text{transition}} = \min\left(n_{\text{total}}, \frac{n_{\text{total}}}{1 + e^{-|\theta_{yaw}|}}\right)
\end{equation}
Vehicle orientation is then linearly interpolated over these points (Eq.~\eqref{eq:23}), before the final transformation (Eq.~\eqref{eq:24}) uses quaternions to yield the executable UGV path $\phi$.
\begin{equation}\label{eq:23}
	\theta_{yaw, roll}(i) = \theta_{yaw, roll} \left(1 - \frac{i}{n_{\text{transition}}}\right)
\end{equation}
\begin{equation} \label{eq:24}
	\vec{p}_{global}(i) =
	\begin{cases}
		q_{\theta}(i) \otimes \vec{p}_{local}(i) & \text{if } i < n_{\text{transition}}  \\
		q_0 \otimes \vec{p}_{local}(i) & \text{otherwise}
	\end{cases}
\end{equation}

\subsection{UAV PATH PLANNER}
The UAV path planner takes the final, refined UGV path $\phi$ as a foundational reference, generating a safe and efficient aerial path that respects communication, altitude, and obstacle constraints unique to the aerial platform. This hierarchical dependency ensures synchronized movement while capitalizing on the UAV's distinct maneuverability.

\subsubsection{Initial Path Planning}
\begin{figure}
	\centerline{\includegraphics[width=3.2in]{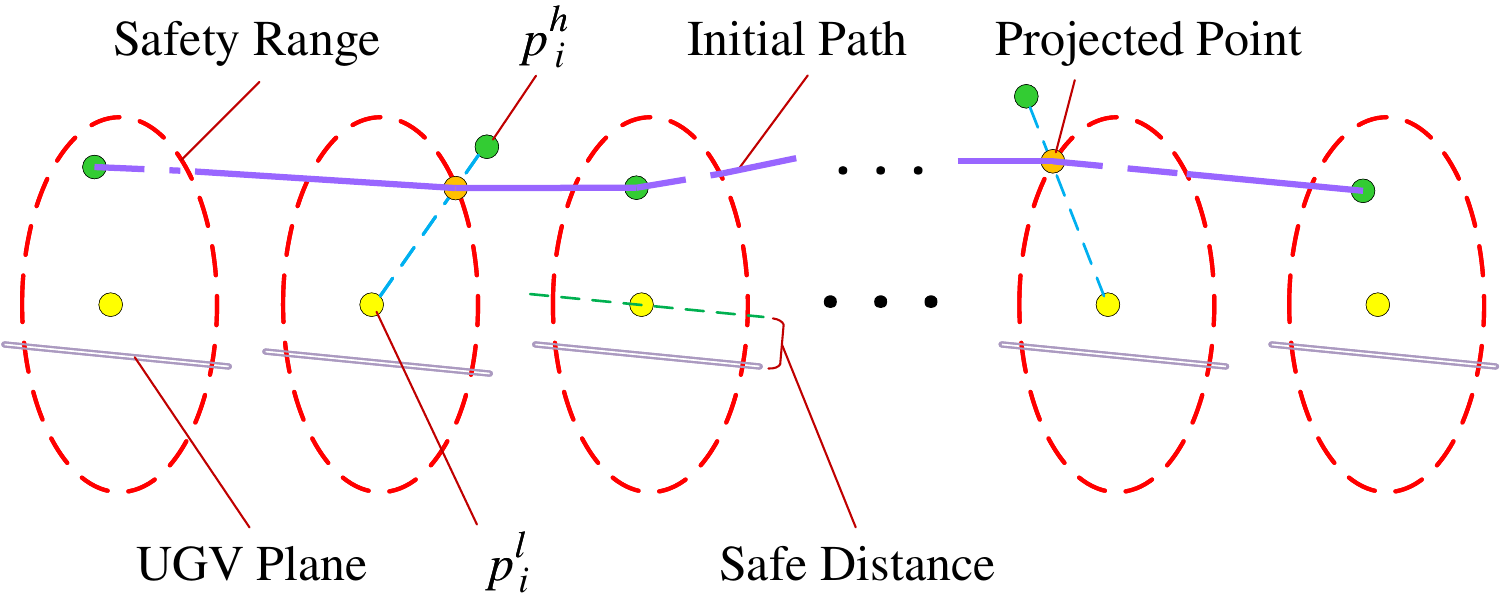}}
	\caption{Illustration of UAV initial path planning, ensuring points remain within the communication safety constraint relative to the UGV path.\label{fig9}}
\end{figure}

Instead of suboptimally projecting the UGV path vertically, the planner first establishes a safe flight constraint. A minimum safe altitude path, $P_{lowest}$, is generated from $\phi$. A desired altitude path, $P_h$, is projected from the tunnel centerline. As shown in Fig.~\ref{fig9}, a communication safety constraint is constructed around each point on $P_{lowest}$. The initial UAV path, $\Gamma_{\mathrm{initial}}$, is formed by ensuring each point from $P_h$ lies within or projects onto the corresponding safety constraints.
\begin{equation} \label{eq:25}
	\gamma_i =
	\begin{cases}
		p_i^h, & \text{if } \| p_i^h - p_i^l \| \leqslant r \\
		p_i^l + r \cdot \dfrac{p_i^h - p_i^l}{\|p_i^h - p_i^l\|}, & \text{otherwise}.
	\end{cases}
\end{equation}

\subsubsection{Dynamic Sampling Iterative Path Optimization}
\begin{figure}
	\centerline{\includegraphics[width=2.5in]{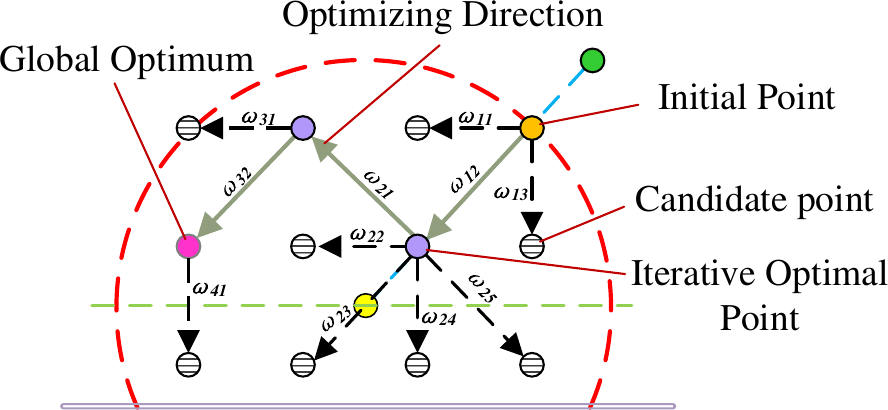}}
	\caption{Overview of the iterative optimization process for the UAV path.\label{fig10}}
\end{figure}

The initial path requires optimization for smoothness and obstacle avoidance. FLISP uses a dynamic sampling iterative approach  (Fig.~\ref{fig10}). For each waypoint, candidate points $\boldsymbol{\zeta}_i$ are generated in its vicinity using a polar-coordinate strategy.
\begin{equation} \label{eq:26}
	\boldsymbol{\zeta}_i = \boldsymbol{\zeta}_0 + r_{sp} \begin{bmatrix} \cos\theta_i \\ \sin\theta_i \\ 0 \end{bmatrix}
\end{equation}

Each candidate is evaluated using a multi-objective cost function, formulated as a weighted sum of four key metrics:
\begin{equation} \label{eq:27}
	J(\boldsymbol{\zeta}) = \alpha {J_1}(\boldsymbol{\zeta}) + \beta {J_2}(\boldsymbol{\zeta}) - \eta {J_3}(\boldsymbol{\zeta}) + \delta {J_4}(\boldsymbol{\zeta})
\end{equation}
\noindent\textbf{1) Safety Cost ($J_1$):} Penalizes proximity to obstacles within a cylindrical safety corridor. If an obstacle point $e_j$ is detected (Eq.~\eqref{eq:28}), a quadratic penalty is applied (Eq.~\eqref{eq:30}), based on its distance to the path segment (Eq.~\eqref{eq:29}).
\begin{equation} \label{eq:28}
	\exists {e_j} \in E,d({e_j},\overline {{\gamma_i}{\gamma_{i + 1}}} ) < {r_\text{safe}}
\end{equation}
\begin{equation} \label{eq:29}
	d({e_j},\overline {{\gamma_i}{\gamma_{i + 1}}} ) = \frac{{\left| {({\gamma_{i + 1}} - {\gamma_i}) \times ({\gamma_i} - {e_j})} \right|}}{{\left| {{\gamma_{i + 1}} - {\gamma_i}} \right|}}
\end{equation}
\begin{equation} \label{eq:30}
	{J_1}(\boldsymbol{\zeta}) = 
	\begin{cases}
		{{(r - d({e_j},\overline {{\gamma_i}{\gamma_{i + 1}}} ))}^2}, & d({e_j},\overline {{\gamma_i}{\gamma_{i + 1}}} ) < r \\
		0,                & \text{otherwise}.
	\end{cases}
\end{equation}

\noindent\textbf{2) Smoothness Cost ($J_2$):} Penalizes deviation from a straight line to encourage smooth paths.
\begin{equation} \label{eq:31}
	{J_2}(\boldsymbol{\zeta}) = d(\boldsymbol{\zeta},\overline {{\boldsymbol{\zeta}_{start}}{\boldsymbol{\zeta}_{end}}} )
\end{equation}

\noindent\textbf{3) Progress Reward ($J_3$):} Rewards efficient progress towards the goal.
\begin{equation} \label{eq:32}
	{J_3}(\boldsymbol{\zeta}) = \frac{{\left| {{\boldsymbol{\zeta}_{start}} - {\boldsymbol{\zeta}_{end}}} \right| - \left| {\boldsymbol{\zeta} - {\boldsymbol{\zeta}_{end}}} \right|}}{{\left| {{\boldsymbol{\zeta}_{start}} - {\boldsymbol{\zeta}_{end}}} \right|}}
\end{equation}

\noindent\textbf{4) Height Consistency Cost ($J_4$):} Penalizes unnecessary altitude variations to save energy.
\begin{equation} \label{eq:33}
	{J_4}(\boldsymbol{\zeta}) = {({\boldsymbol{\zeta}_z} - {\boldsymbol{\zeta}_{start,z}})^2}
\end{equation}
The optimization problem is to find the candidate point $\boldsymbol{\zeta}^*$ that minimizes the total cost.
\begin{equation} \label{eq:34}
	{\boldsymbol{\zeta}^*} = \arg \mathop {\min }\limits_{_{\boldsymbol{\zeta} \in {\mathcal{Z}}}} J(\boldsymbol{\zeta})
\end{equation}
The final optimized UAV path $\epsilon$ is the sequence of these optimal points.
\begin{equation} \label{eq:35}
	\epsilon = \{\boldsymbol{\zeta}_0^*, \boldsymbol{\zeta}_1^*, \ldots, \boldsymbol{\zeta}_{n}^*\}
\end{equation}

\subsection{TRAJECTORY PLANNING AND CONTROLLER}
\begin{figure}
	\centerline{\includegraphics[width=3.5in]{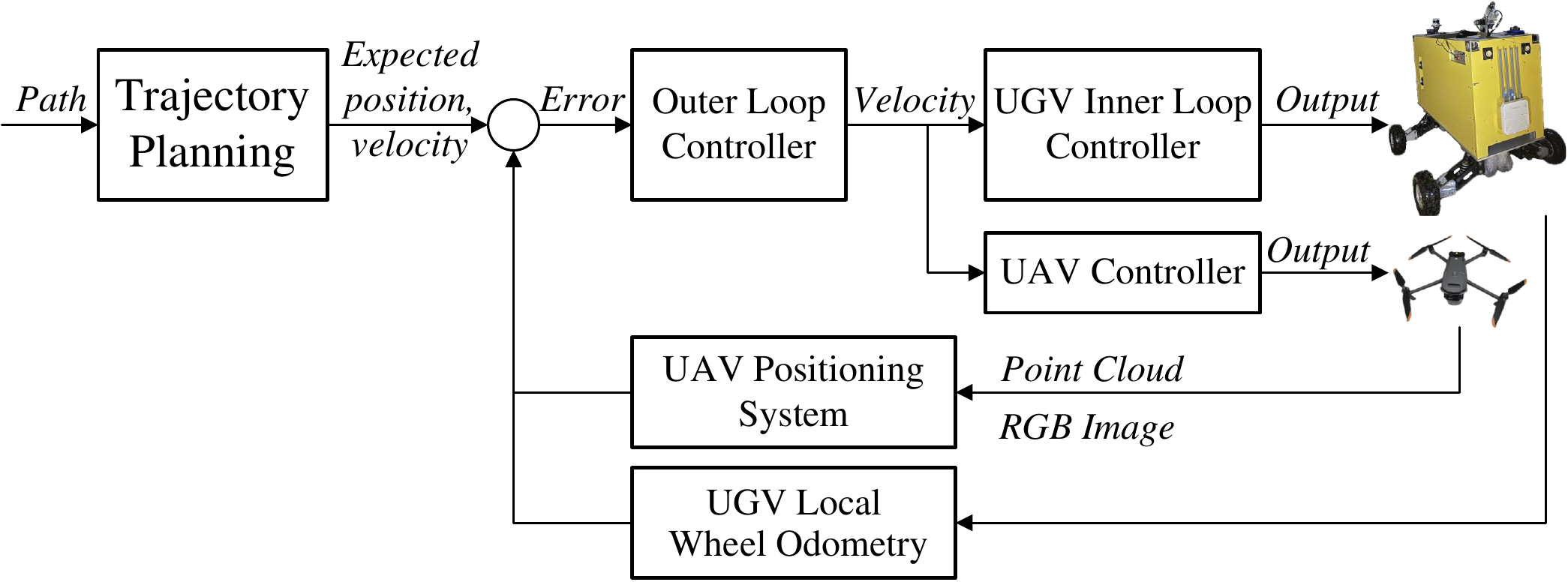}}
	\caption{Control framework following FLISP path generation.\label{fig11}}
\end{figure}

A key advantage of the FLISP framework is that it generates high-quality, kinematically aware paths directly from sensor data. This obviates the need for complex, computationally expensive trajectory optimization layers that are often required to smooth out paths from conventional planners. Consequently, we can employ standard, lightweight trajectory generation methods and simple controllers to effectively track the planned path. This section briefly outlines these implementations, which serve to validate the practical feasibility of the paths generated by FLISP. The overall control architecture is illustrated in Fig.~\ref{fig11}.

For both platforms, discrete waypoints from FLISP are converted into continuous, time-parameterized trajectories. For the UGV, we employ a standard minimum-jerk trajectory generation based on quintic polynomial interpolation between waypoints, of the form:
\begin{equation} \label{eq:36}
	\mathbf{p}(t) = (1 - h_1(t))\mathbf{p}_0 + h_1(t)\mathbf{p}_1,
\end{equation}
\noindent where $h_1(t)$ is the quintic polynomial basis function. To ensure feasibility, a dynamic time reallocation mechanism is applied to respect the vehicle's velocity and acceleration limits, following the approach in \cite{bib57}.

For the UAV, a trajectory is generated using a non-uniform B-spline representation, which ensures smoothness up to a high derivative order. To mitigate excessive velocities at sharp turns, we incorporate a curvature-aware time allocation mechanism by modulating a time scaling factor $\lambda_{\text{curv}}$:
\begin{equation} \label{eq:40}
	\lambda_{\text{curv}} = 1.0 + k_{\text{curv}}(1.0-\cos\theta)^{p}.
\end{equation}
\noindent
where ${\theta}$ is the heading change angle, $k_{\text{curv}}$ is a gain parameter, and $p$ is the curvature sensitivity exponent, empirically set to $k_{\text{curv}} = 8.0$ and $p = 1.5$. This ensures the UAV automatically slows down at corners. A final time-scaling pass guarantees that all dynamic constraints are met.

Finally, a feedforward Proportional-Derivative (PD) controller, a standard choice for trajectory tracking, serves as the outer-loop controller for both vehicles. It computes velocity commands based on the errors between the current state and the desired state from the generated trajectory, which are then sent to the inner-loop controllers of the UGV and UAV.

\section{SIMULATION EXPERIMENTS}

\begin{figure*}
	\centerline{\includegraphics[width=6in]{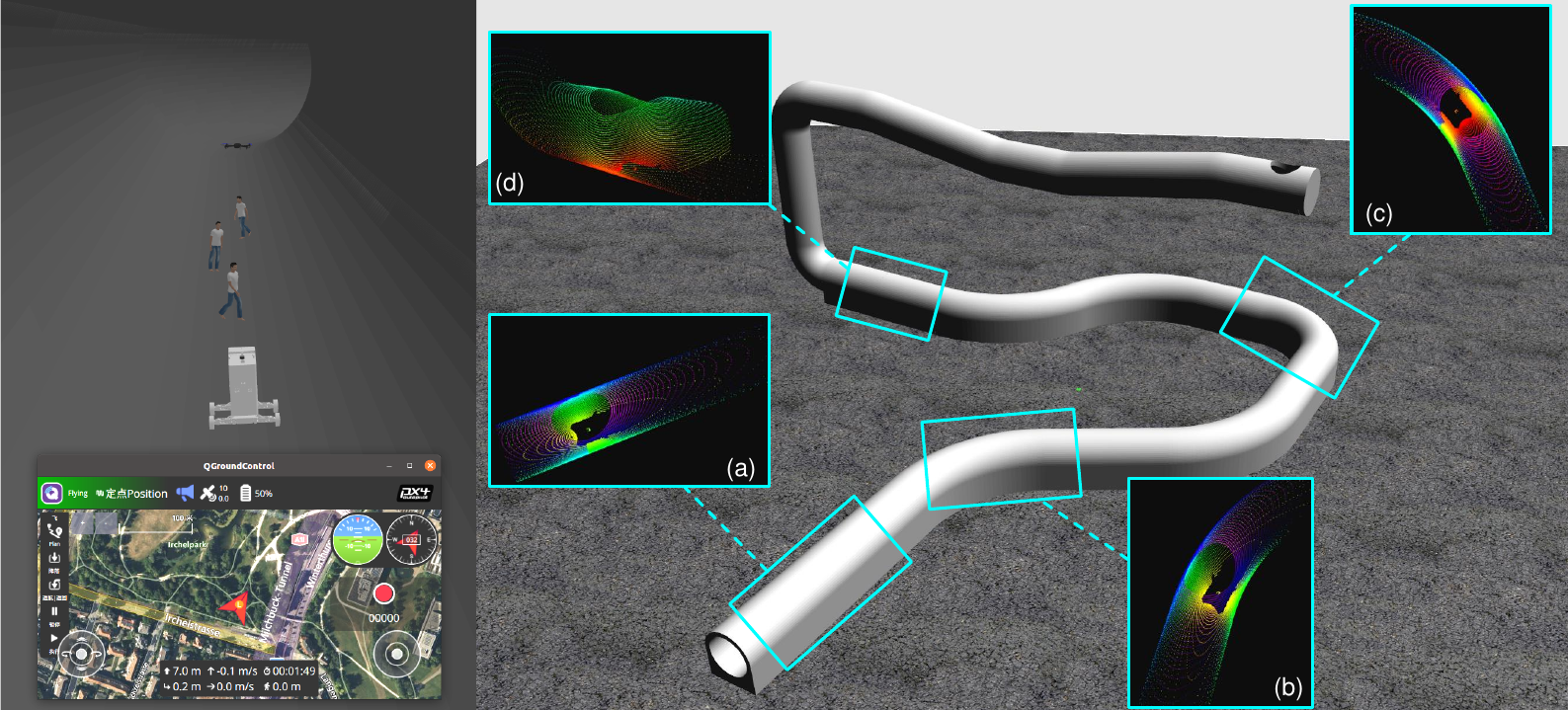}}  
	\caption{Overview of the simulated tunnel environment. Left: Operational scene with obstacles. Right: Representative point clouds of a straight segment (a), right turn (b), left turn (c), and a sluice gate (d).\label{fig12}}
\end{figure*}

To validate FLISP without risking physical assets, we developed a high-fidelity Gazebo environment (Fig.~\ref{fig12}) replicating real water-conveyance tunnel geometries, including 13 m diameter sections with varying curvatures and sluice gates. Simulations facilitate rigorous parameter tuning and stress testing across diverse topologies.

\subsection{Performance Evaluation}
We assessed FLISP across three core competencies: nominal tracking, dynamic obstacle avoidance, and runtime efficiency. The visualizations in Figs. \ref{fig13}--\ref{fig15} depict representative paths from single trial runs to illustrate topological behavior. Quantitative metrics (Table \ref{tab:1}) and runtime distributions (Fig.~\ref{fig16}) are derived from statistical analysis over 100 independent trials per scenario.

\subsubsection{Nominal Planning Robustness}
\begin{figure*}
	\centerline{\includegraphics[width=6.2in]{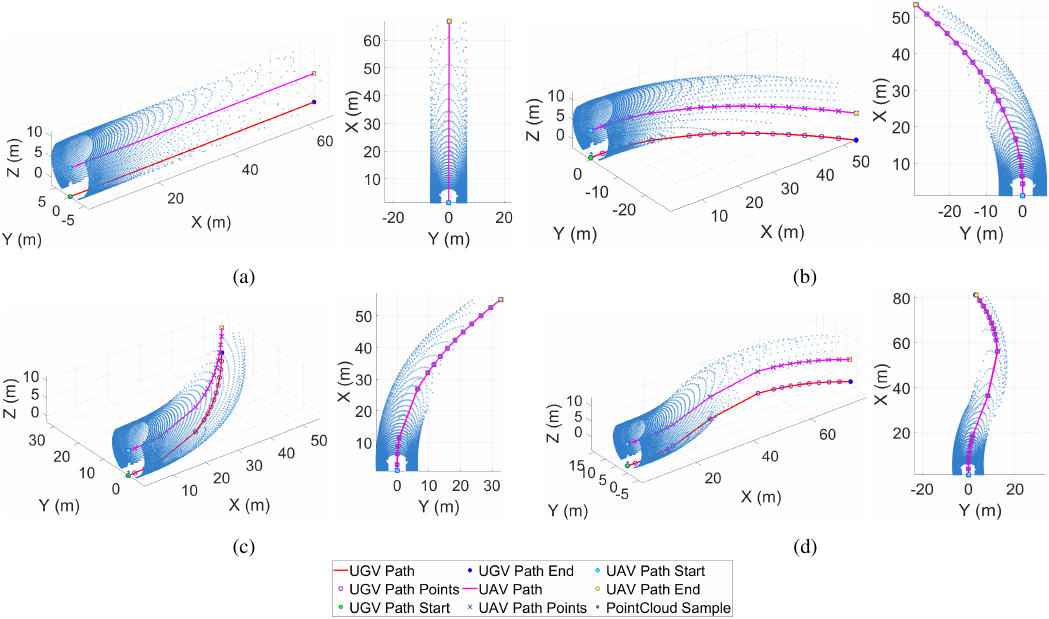}}  
	\caption{The path planning results of FLISP in simulated tunnel environment, encompassing scenarios with straight segments (a), left (b) and right turn segments (c), and consecutive curves (d).\label{fig13}}
\end{figure*}

We evaluated kinematic feasibility in four canonical geometries: straight, left/right turns, and consecutive curves (Fig.~\ref{fig13}). The UGV demonstrated exceptional path stability. In straight sections, deviations were negligible, appearing in just 6\% of trials due to simulated sensor noise. In curved sections, path stability positively correlated with point cloud density. FLISP's hierarchical fitting effectively compensated for sparse regions, maintaining continuous floor alignment. Even in challenging consecutive turns, where frequent curvature changes required rapid replanning, the system maintained high accuracy, with minor deviations occurring only under extreme roll/yaw conditions. UAV paths, derived from the UGV master path, maintained synchronized formation.

\subsubsection{Obstacle Avoidance Capabilities}
\begin{figure}
	\centering
	\includegraphics[width=1\linewidth]{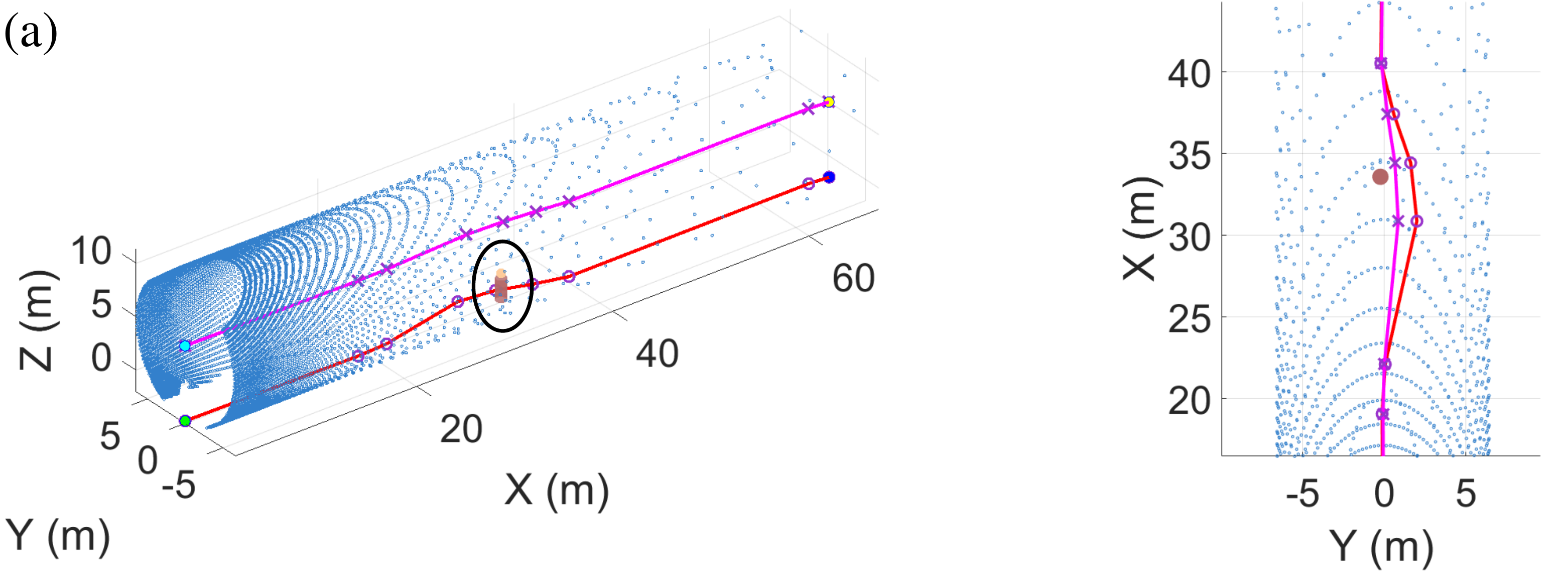} \\
	\vspace{1pt}
	\includegraphics[width=1\linewidth]{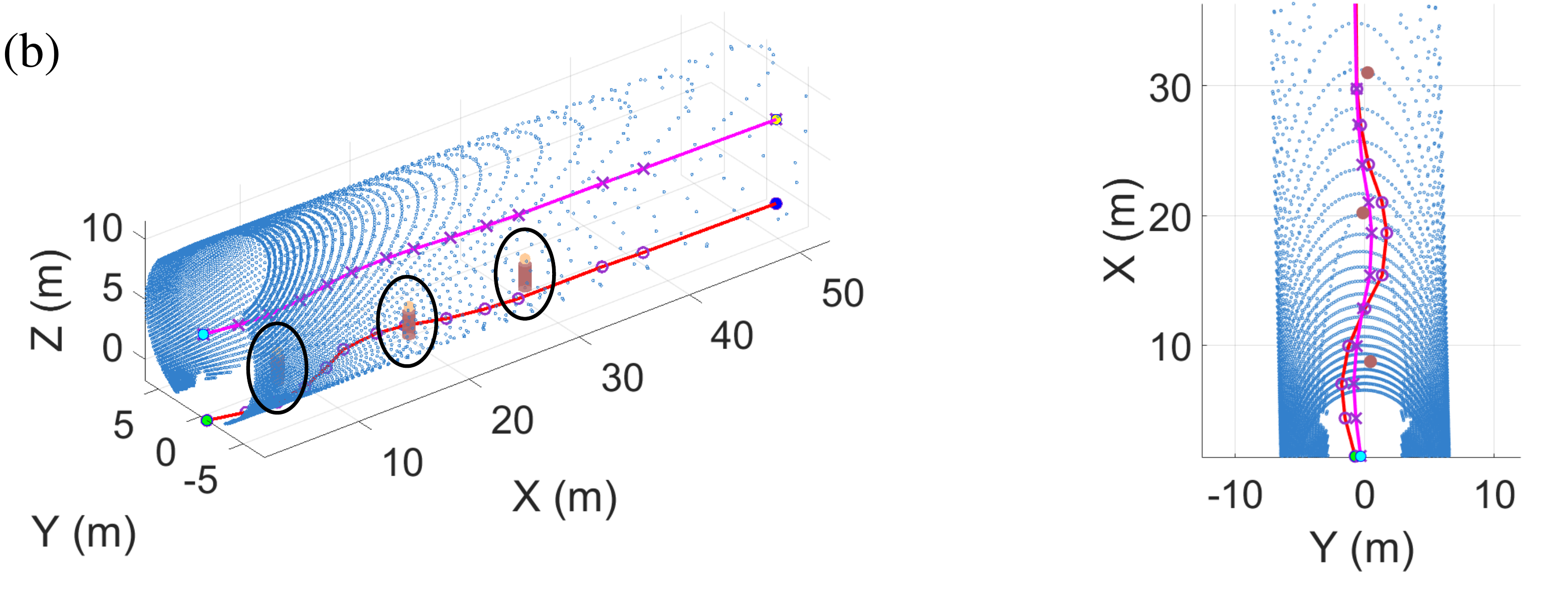} \\
	\vspace{1pt}
	\includegraphics[width=1\linewidth]{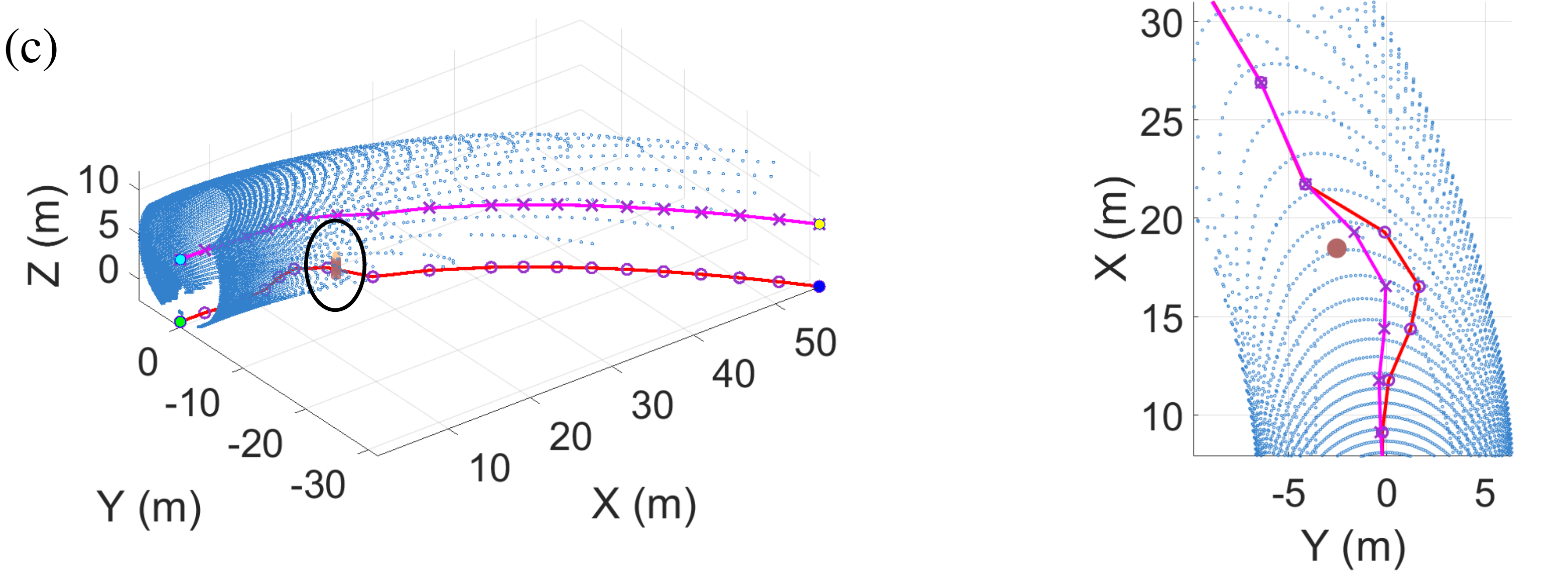} \\
	\vspace{1pt}
	\includegraphics[width=1\linewidth]{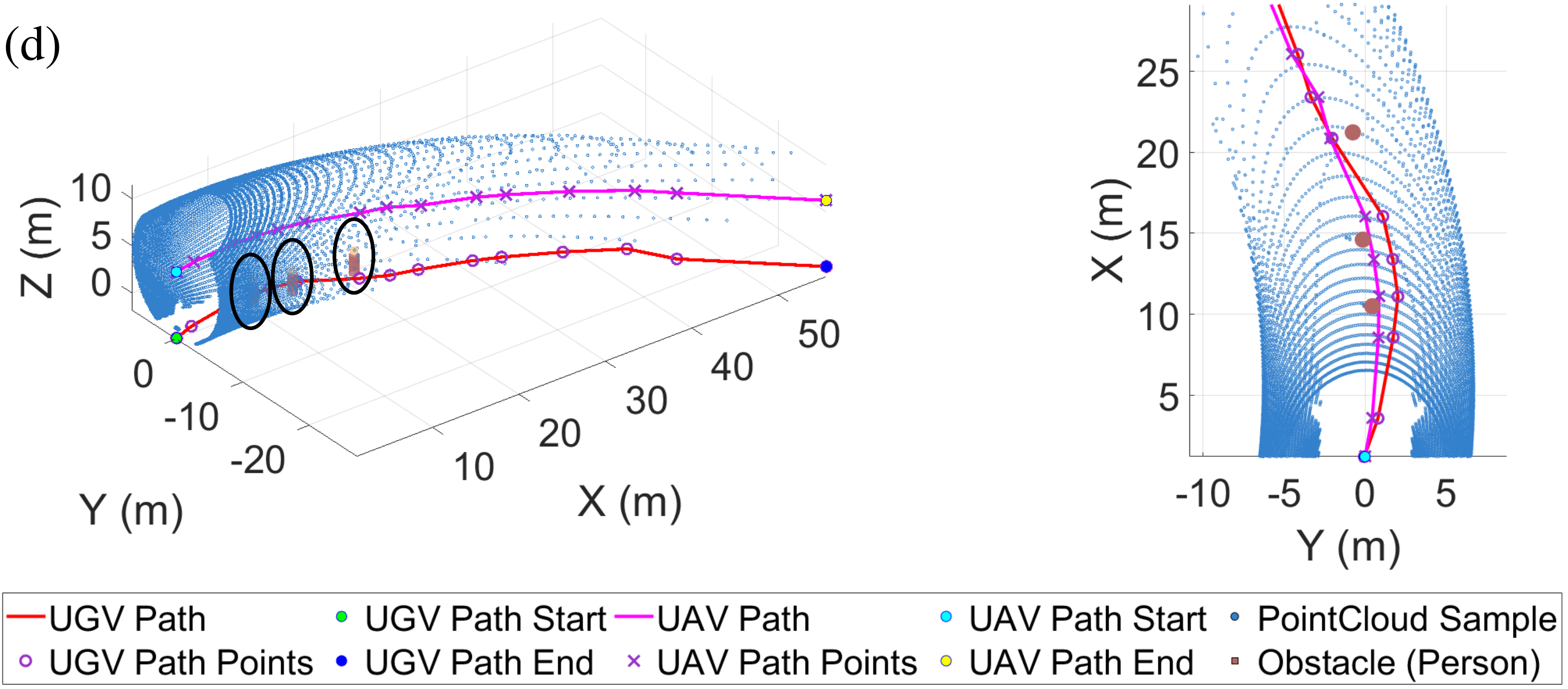} \\
	\vspace{1pt}
	\caption{FLISP obstacle avoidance (personnel) in the simulated tunnel. Scenarios include single and multiple obstacles in straight (a, b) and curved (c, d) segments.\label{fig14}}
\end{figure}

\begin{figure}
	\centering
	\includegraphics[width=1\linewidth]{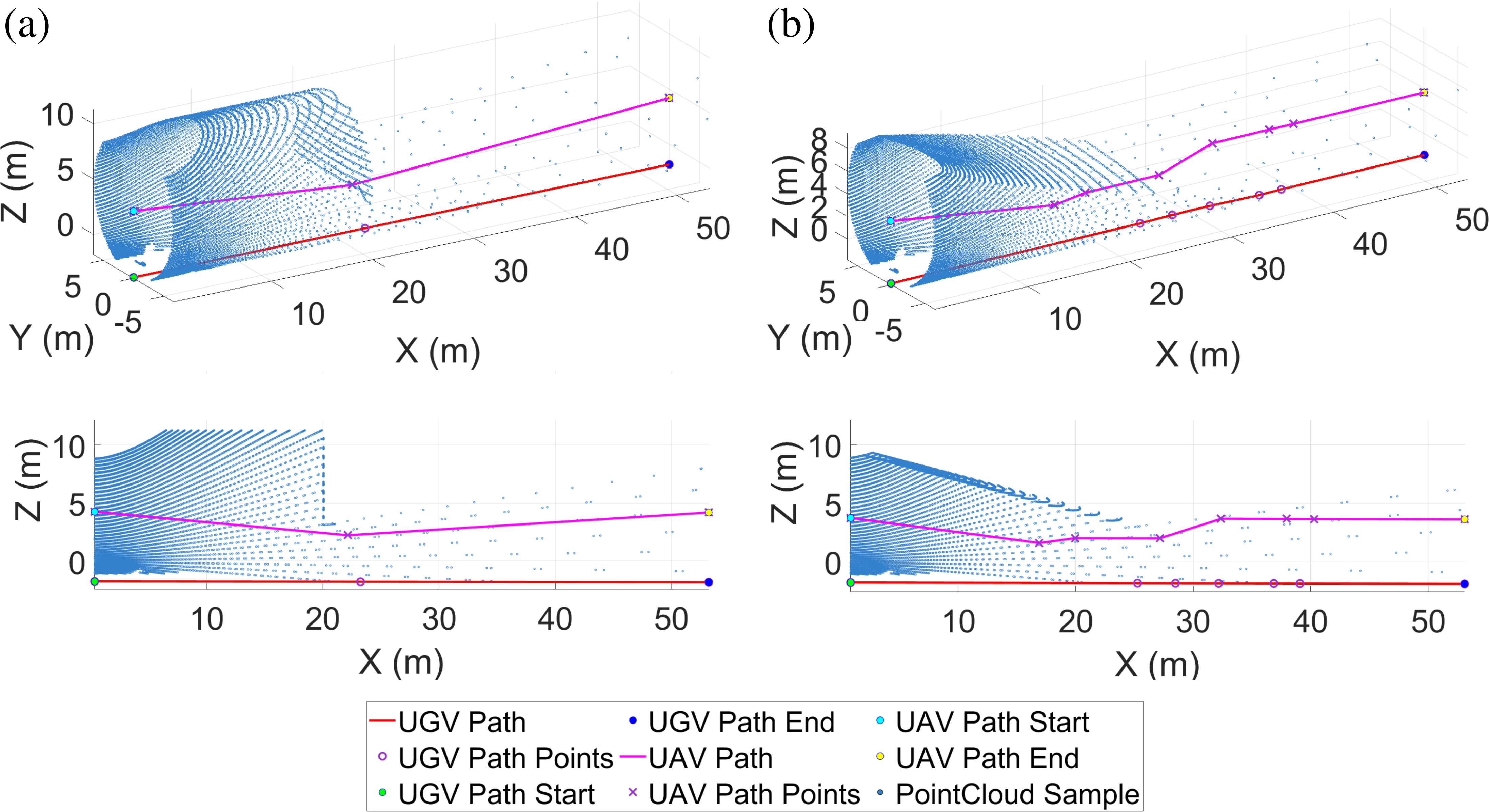}
	\caption{FLISP gate traversal in the simulated tunnel, demonstrating UAV altitude recovery paths during forward (a) and reverse (b) traversals.\label{fig15}}
\end{figure}

\begin{itemize}
	\item\textbf{Personnel Avoidance:} We introduced single and multi-agent obstacles (simulating maintenance personnel) in both straight and curved segments (Fig.~\ref{fig14}). FLISP consistently generated smooth evasion paths for the UGV, strictly adhering to slope and path length constraints to ensure floor alignment. For the UAV, altitude separation alone proved sufficient for collision avoidance, resulting in minimal path deviation. Notably, the algorithm demonstrated robust convergence even in curved, sparse-data regions without compromising downstream path fidelity.
	
	\item\textbf{Sluice Gate Traversal:} Gates impose vertical constraints, requiring the UAV to dip below the structure (Fig.~\ref{fig15}). Dimensionality reduction revealed gate sections geometrically resemble straight tunnels to the UGV, resulting in minimal path impact. For the UAV, the planner successfully generated ``dive-and-recovery'' paths in both forward and reverse directions. The altitude recovery phase occasionally exhibited discrete staging, an optimization solver characteristic ensuring safe clearance.

\end{itemize}

\subsubsection{Runtime Efficiency Analysis}
\begin{figure}
	\centerline{\includegraphics[width=3.5in]{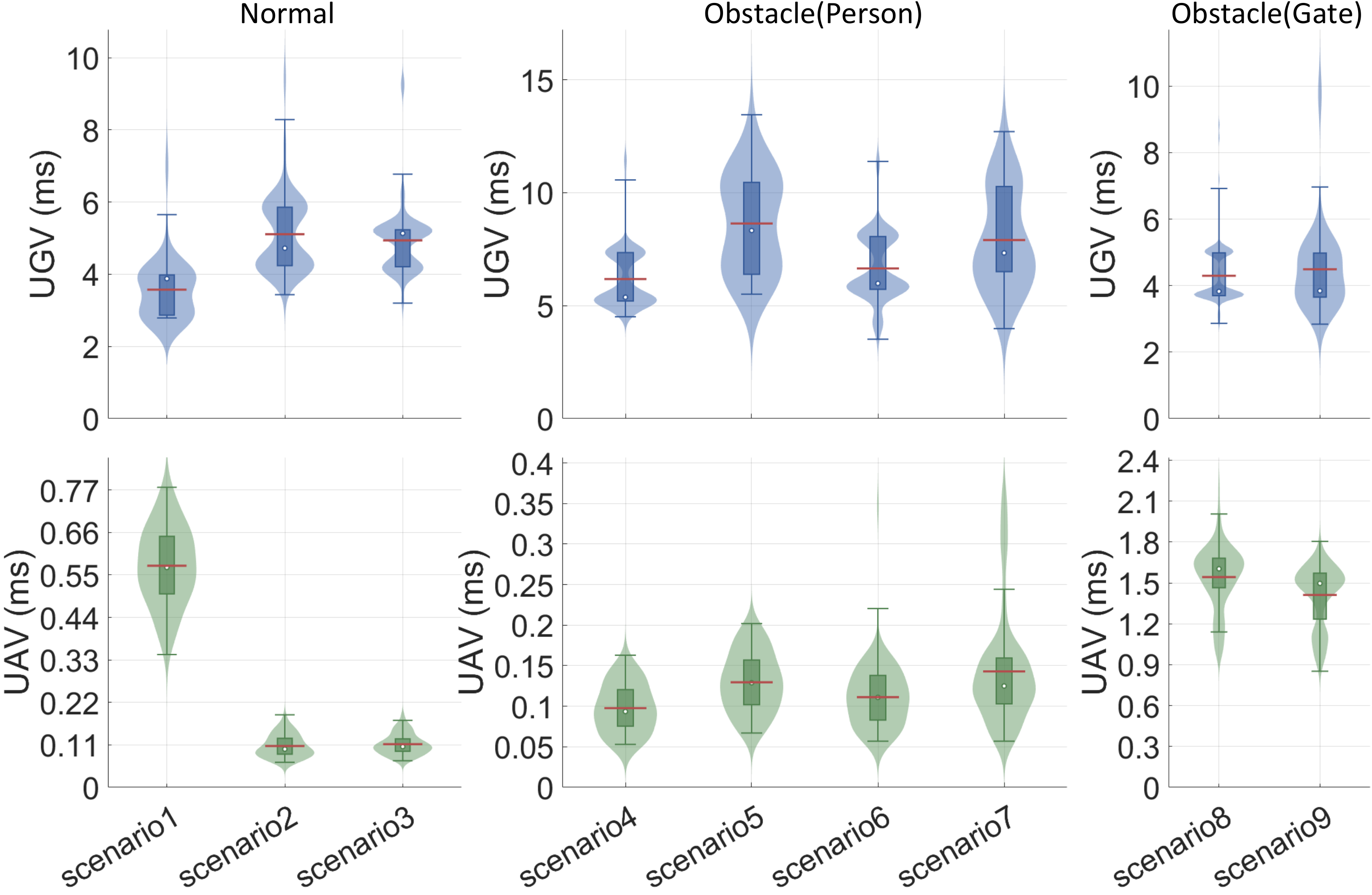}}  
	\vspace{10pt}
	\caption{Runtime distribution across nine scenarios ($N=100$ trials each). Red lines: Mean; White dots: Median.\label{fig16}}
\end{figure}

As illustrated in Fig.~\ref{fig16}, we evaluated FLISP's runtime performance across nine scenarios: (1) Normal straight segment, (2) Normal curved segment, (3) Normal consecutive curved segments, (4) Straight segment with a single obstacle, (5) Straight segment with multiple obstacles, (6) Curved segment with a single obstacle, (7) Curved segment with multiple obstacles, (8) Sluice gate segment (forward), and (9) Sluice gate segment (reverse).

Normal scenarios demonstrated the most efficient performance. Specifically, straight segments (Scenario 1) required only linear inference ($\approx$ 3.50 ms), making them faster than the polynomial fitting needed for curved segments (Scenarios 2 \& 3, $\approx$ 5.0 ms).

\begin{itemize}
	\item\textbf{UGV Performance:} The introduction of obstacles resulted in a marginal increase in computation time (Scenarios 4-7 averaged 6.2-8.8 ms). This slight variance is primarily attributed to the simulation host's overhead (e.g., physics engine and ray-casting load) rather than algorithmic complexity. Crucially, the planner maintains robust efficiency, keeping all execution times well within the 10 ms safety threshold.
	
	\item\textbf{UAV Performance:} While generally sub-1 ms in standard tunnels, UAV planning latency spiked in sluice gate scenarios (Scenarios 8 \& 9, $\approx$ 1.5 ms). This increase stems from the additional computational overhead required for 3D obstacle segmentation and feature extraction in the vertical domain to execute the ``dive-and-recovery'' maneuver.
\end{itemize}
These sub-10 ms latencies represent a significant margin over real-time requirements, validating framework efficiency.

\subsection{Quantitative Metrics}
Table \ref{tab:1} summarizes the aggregate metrics.

\begin{itemize}
	\item\textbf{Geometric Complexity:} Waypoint count and path length naturally correlate with environmental complexity. Curved segments induce longer paths and higher waypoint densities as vehicles execute finer maneuvers to track curvature.
	\item\textbf{Platform Comparison:} The UAV consistently generated shorter paths with fewer waypoints than the UGV. This efficiency derives from omnidirectional maneuverability and altitude freedom, allowing 3D corner-cutting unachievable by the ground-constrained UGV.
\end{itemize}

\begin{table}[htbp]
	\centering
	\caption{Average analysis metrics over 100 simulated trials.}
	\label{tab:1}
	\begin{tabular}{|>{\centering\arraybackslash}m{1.2cm}||cc|cc|cc|}
		\hline
		\multirow{2}{*}{\textbf{Scenario \#}}
		& \multicolumn{2}{c|}{\textbf{Time [ms]}}
		& \multicolumn{2}{c|}{\textbf{Waypoints}}
		& \multicolumn{2}{c|}{\textbf{Length [m]}} \\
		\cline{2-7}
		& \textbf{UGV} & \textbf{UAV}
		& \textbf{UGV} & \textbf{UAV}
		& \textbf{UGV} & \textbf{UAV} \\
		\hline
		1 & 3.50 & 0.57 & 2.38 & 2.03 & 51.57 & 51.51 \\
		2 & 5.06 & 0.11 & 17.14 & 15.62 & 62.05 & 61.93 \\
		3 & 4.88 & 0.11 & 13.43 & 12.42 & 57.39 & 56.66 \\
		4 & 6.21 & 0.10 & 11.93 & 10.38 & 52.20 & 51.90 \\
		5 & 6.61 & 0.11 & 20.69 & 20.38 & 66.47 & 64.45 \\
		6 & 8.75 & 0.13 & 11.76 & 10.04 & 51.55 & 51.23 \\
		7 & 7.82 & 0.14 & 20.07 & 19.18 & 74.89 & 72.17 \\
		8 & 4.32 & 1.54 & 3.58 & 4.68 & 49.98 & 50.25 \\
		9 & 4.50 & 1.41 & 6.38 & 7.40 & 50.18 & 50.69 \\
		\hline
	\end{tabular}
\end{table}

\section{FIELD EXPERIMENTS}
Field trials were conducted in a functional flood discharge tunnel at a major hydropower station. To rigorously evaluate the system's robustness against long-term geometric degeneracy, the experimental scope covered the full longitudinal extent of the facility. As illustrated in Fig.~\ref{fig17}, the environment transitions from an entrance gate into a proximal curve, followed by a dominant straight corridor. 

The UGV autonomously traversed 1.0 km; the final 200 m segment was excluded solely due to deep silt accumulation (Fig.~\ref{fig17}(h)) exceeding chassis mechanical clearance rather than algorithmic limits. Strict safety protocols were enforced given the confined terrain.

\subsection{Experimental Platform and Environment}

The experimental platform comprised a self-developed UGV and a DJI Mavic 3T industrial-grade UAV. The UGV was equipped with an Intel i7-12700 CPU, an NVIDIA GTX 4060-8G GPU, 32 GB of RAM, and an Ouster OS1-64 LiDAR for navigation. Additionally, a Livox Mid-70 LiDAR and an RER-USB4KHDR01 monocular RGB camera were utilized for UAV localization, while a DJI Action 5 camera was employed for tunnel inspection tasks unrelated to the FLISP framework. A ground control station with triple displays, running on a BWJ3P173-220 industrial computer, was used for visualization. The UAV, running on a PX4 autopilot system, was equipped with an infrared camera for inspection purposes independent of FLISP. Communication among all robotic agents was facilitated by two bridged 5 GHz WAVLINK AERIAL HD6 industrial routers, which were physically connected to the ground station via Ethernet.

\begin{figure*}
	\centering
	\includegraphics[width=6.2in]{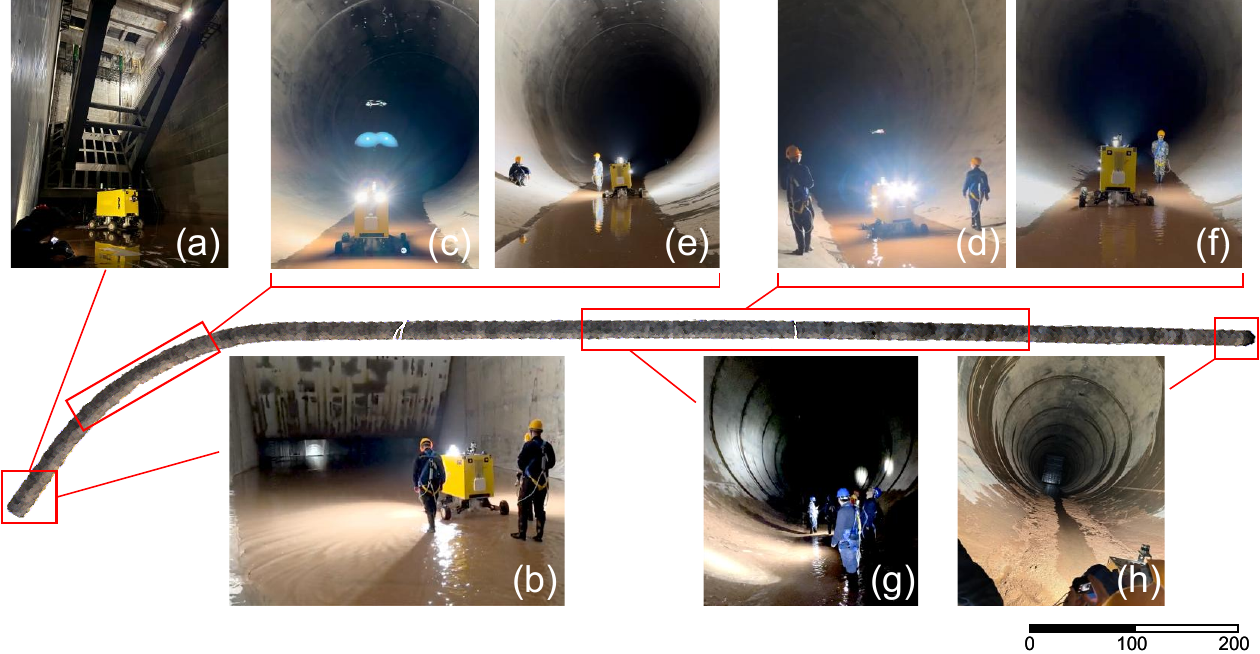}  
	\caption{Overview of the 1.2 km field experimental site. The central strip depicts the top-down point cloud projection. Insets (a)-(h) highlight representative scenarios including entrance gates, varying illumination, blind curves, multi-obstacle evasion, and deep silt accumulation at the terminal gate.\label{fig17}}
\end{figure*}

Given the tunnel's predominantly straight topology, we extracted eight representative scenarios (Fig.~\ref{fig17}) to stress-test specific geometric features, mirroring the simulation benchmarks (excluding the continuous curve). To execute the paths generated by FLISP, we implemented a feedforward path tracking Controller. This module converts the polynomial geometry into distinct chassis motor commands for the UGV and velocity setpoints for the UAV, transmitted via the MAVLink protocol. Detailed telemetry was recorded throughout the navigation process to validate performance under these realistic, constrained conditions.

\begin{figure}
	\centering
	\includegraphics[width=0.8\linewidth]{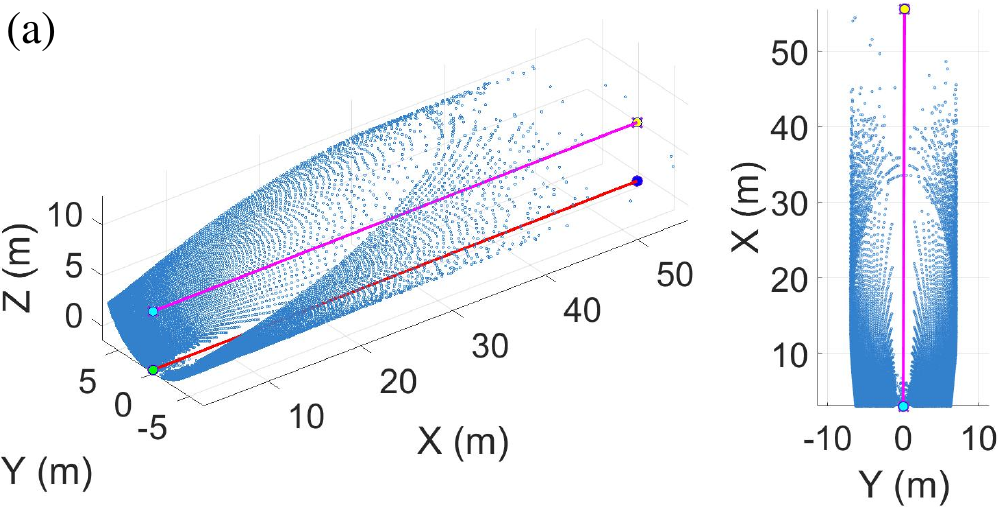} \\
	\centerline{}
	\includegraphics[width=0.8\linewidth]{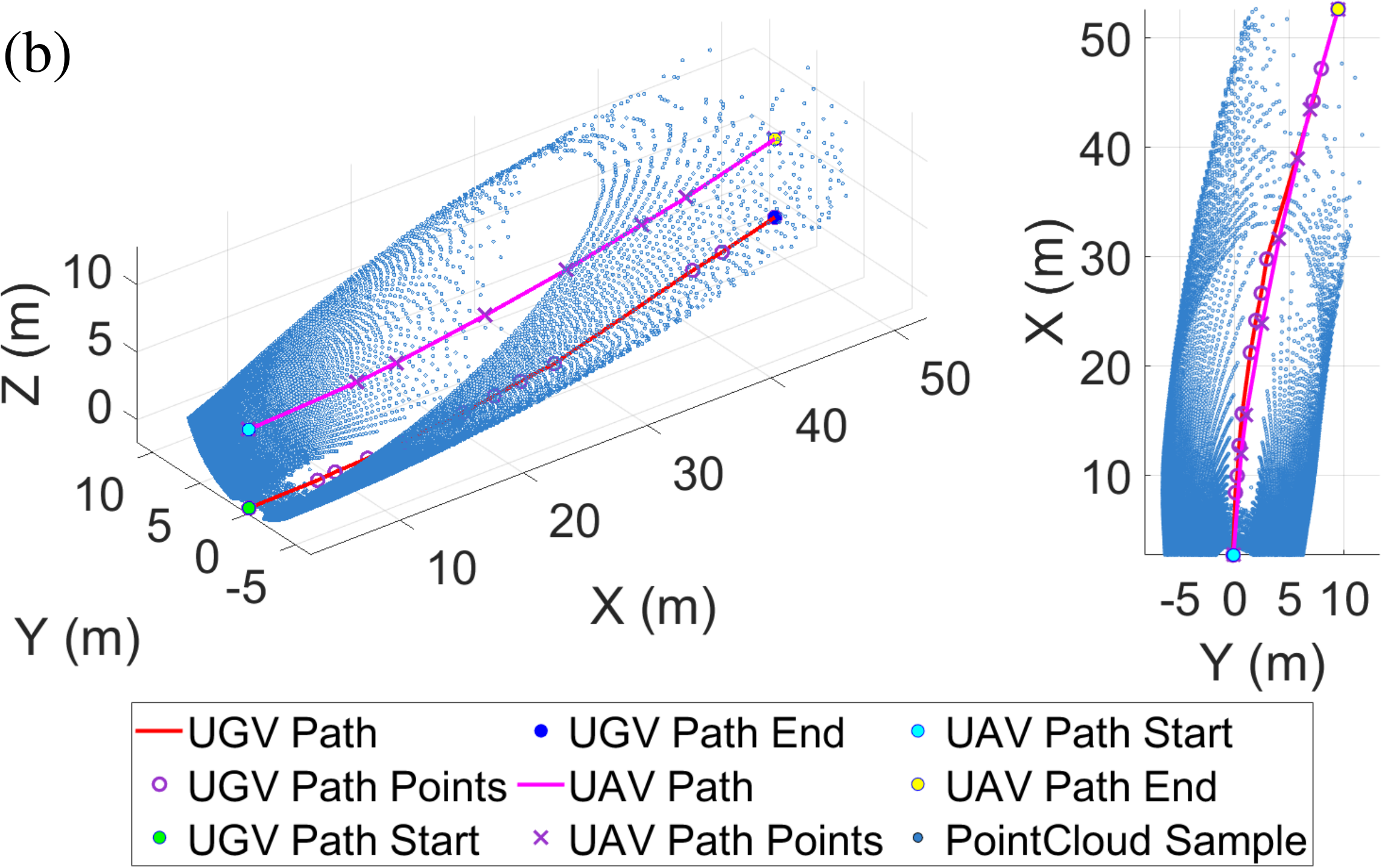} \\
	\caption{FLISP path planning during field experiments in straight (a) and slightly curved (b) segments of the flood discharge tunnel.\label{fig18}}
\end{figure}

\subsection{Performance Analysis and Results}
Building on the successful deployment, we analyze the planner's performance across three critical dimensions: environmental robustness, dynamic obstacle avoidance, and quantitative consistency with simulation.

\subsubsection{Environmental Robustness and Sensor Stability}
Field deployment revealed disturbances absent in simulation.
\begin{itemize}
	\item\textbf{Vibration Effects and Step Loss:} UGV locomotion on uneven terrain transmitted vibrations to the LiDAR. Beyond minor point misclassification, this introduced a ``step loss'' phenomenon where distal point instability truncated the final path segment (typically 2--3 m). Consequently, planned paths were slightly shorter (47--48 m) than the 50 m horizon. However, this marginal truncation is operationally benign: FLISP's high replanning frequency and receding horizon nature ensure that the remaining effective lookahead ($>45$ m) is continuously resolved, maintaining seamless navigation.
	\item\textbf{Atmospheric Resilience:} Despite airborne moisture, dripping water, and dust, the Ouster OS1-64 LiDAR demonstrated high resilience without significant data degradation, confirming its subterranean suitability.
	\item\textbf{Operational Constraints:} Strict safety protocols were enforced. For the narrow sluice gate traversal (Fig.~\ref{fig20}), data was collected via separate approaches from either side due to the insufficient physical opening for safe autonomous passage.
\end{itemize}

\subsubsection{Robustness Validation via Handheld Obstacle Scenarios}
\begin{figure}
	\centering
	\includegraphics[width=0.9\linewidth]{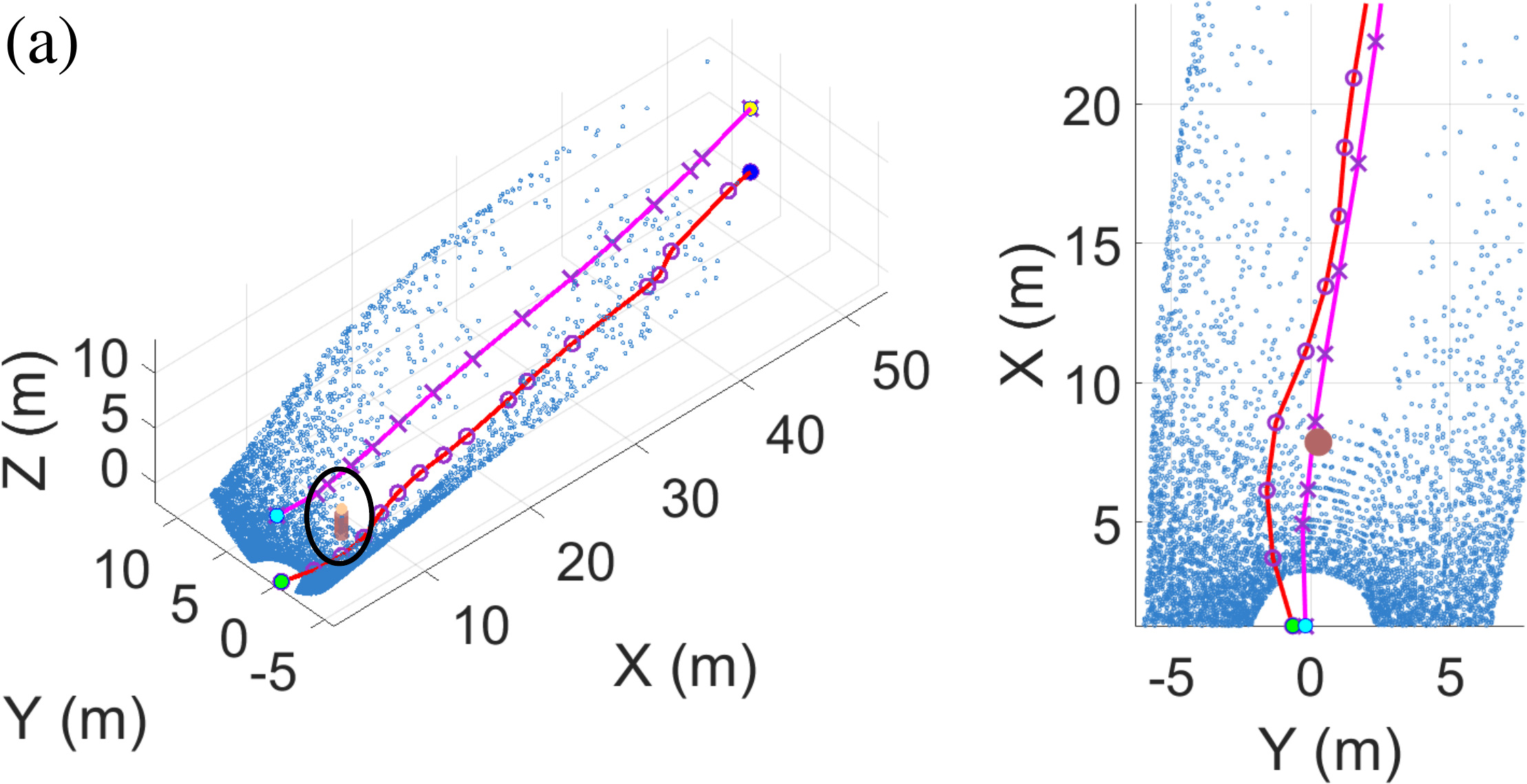} \\
	\vspace{2mm}
	\includegraphics[width=0.9\linewidth]{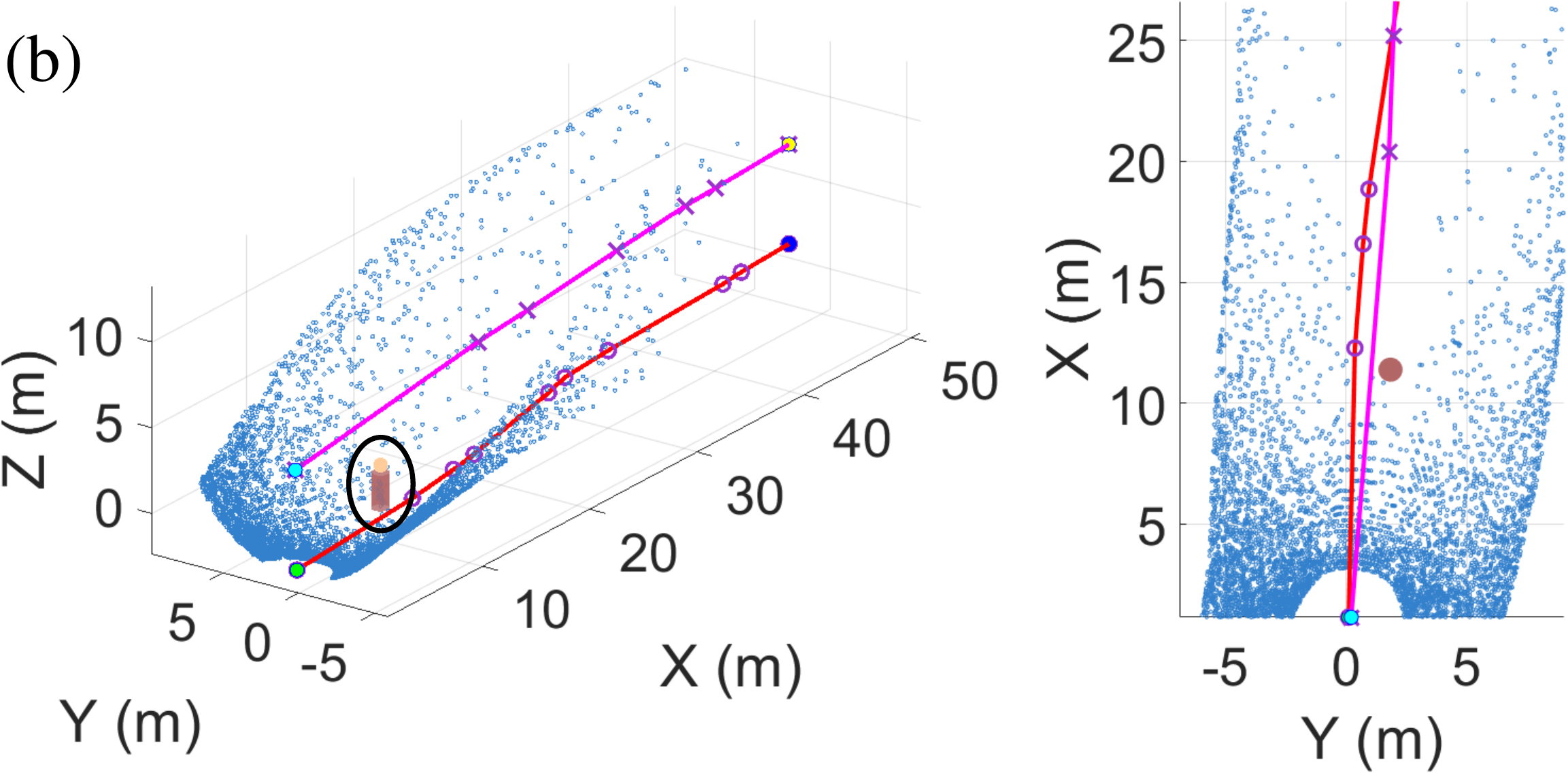} \\
	\vspace{2mm}
	\includegraphics[width=0.9\linewidth]{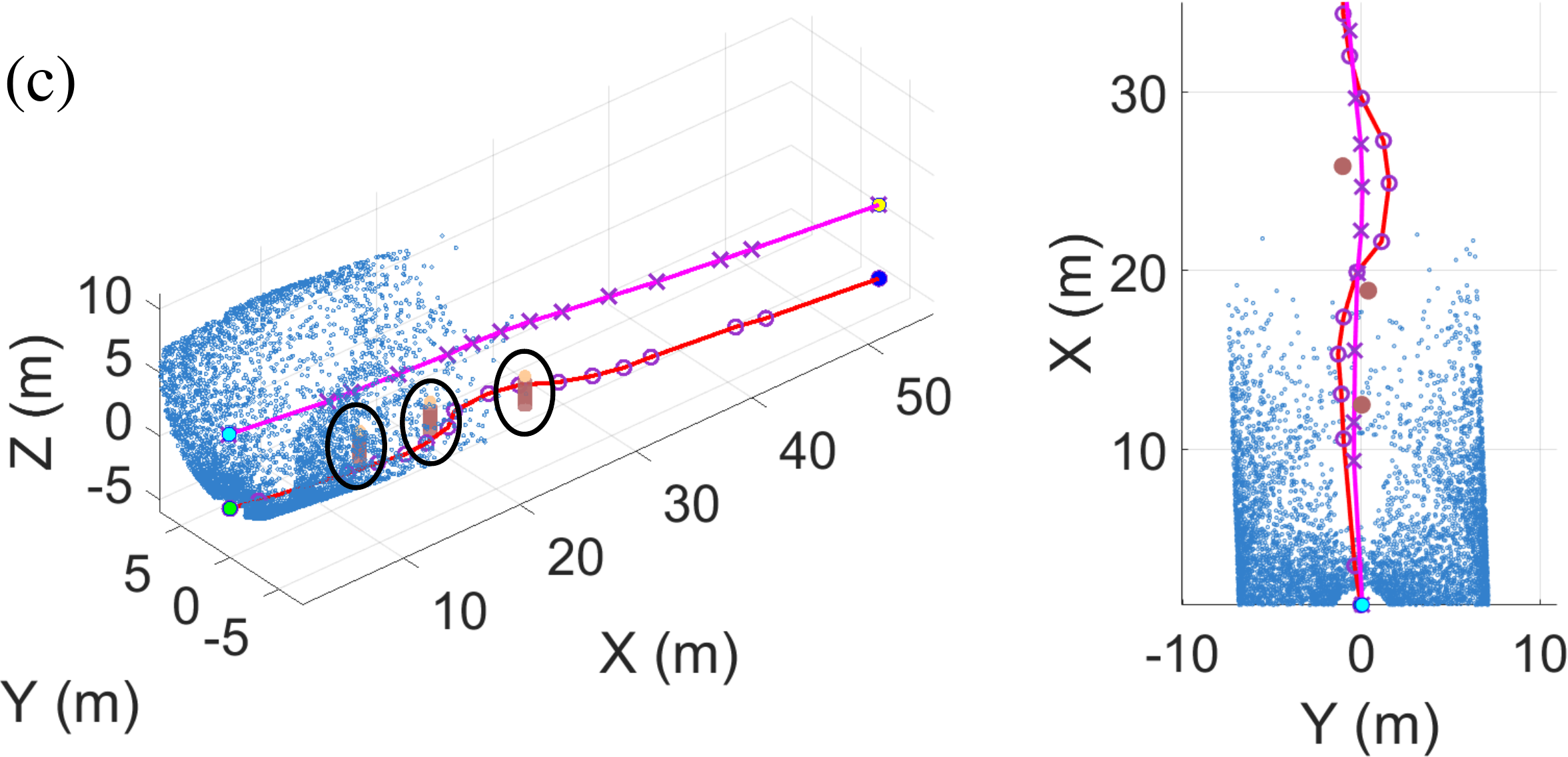} \\
	\vspace{2mm}
	\includegraphics[width=0.9\linewidth]{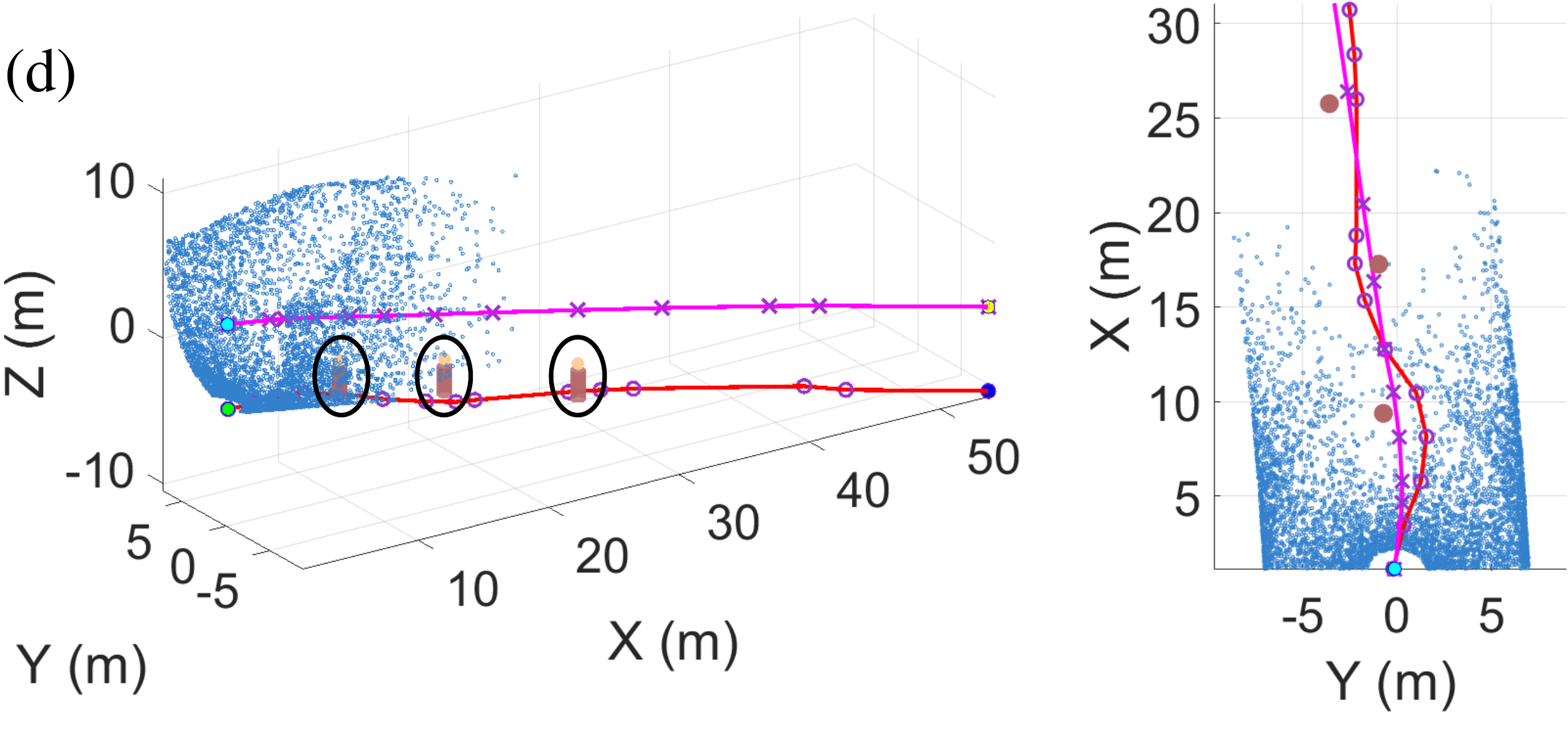} \\
	\includegraphics[width=1\linewidth]{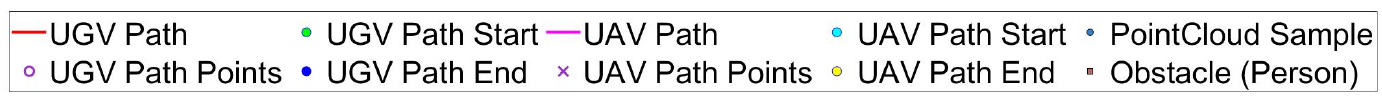} \\
	\caption{Obstacle avoidance field tests with personnel (circled). Scenarios include single obstacle in straight (a) and curved (b) segments, and multi-obstacle in straight (c) and curved (d) segments.\label{fig19}}

\end{figure}

\begin{figure}
	\centering
	\includegraphics[width=0.85\linewidth]{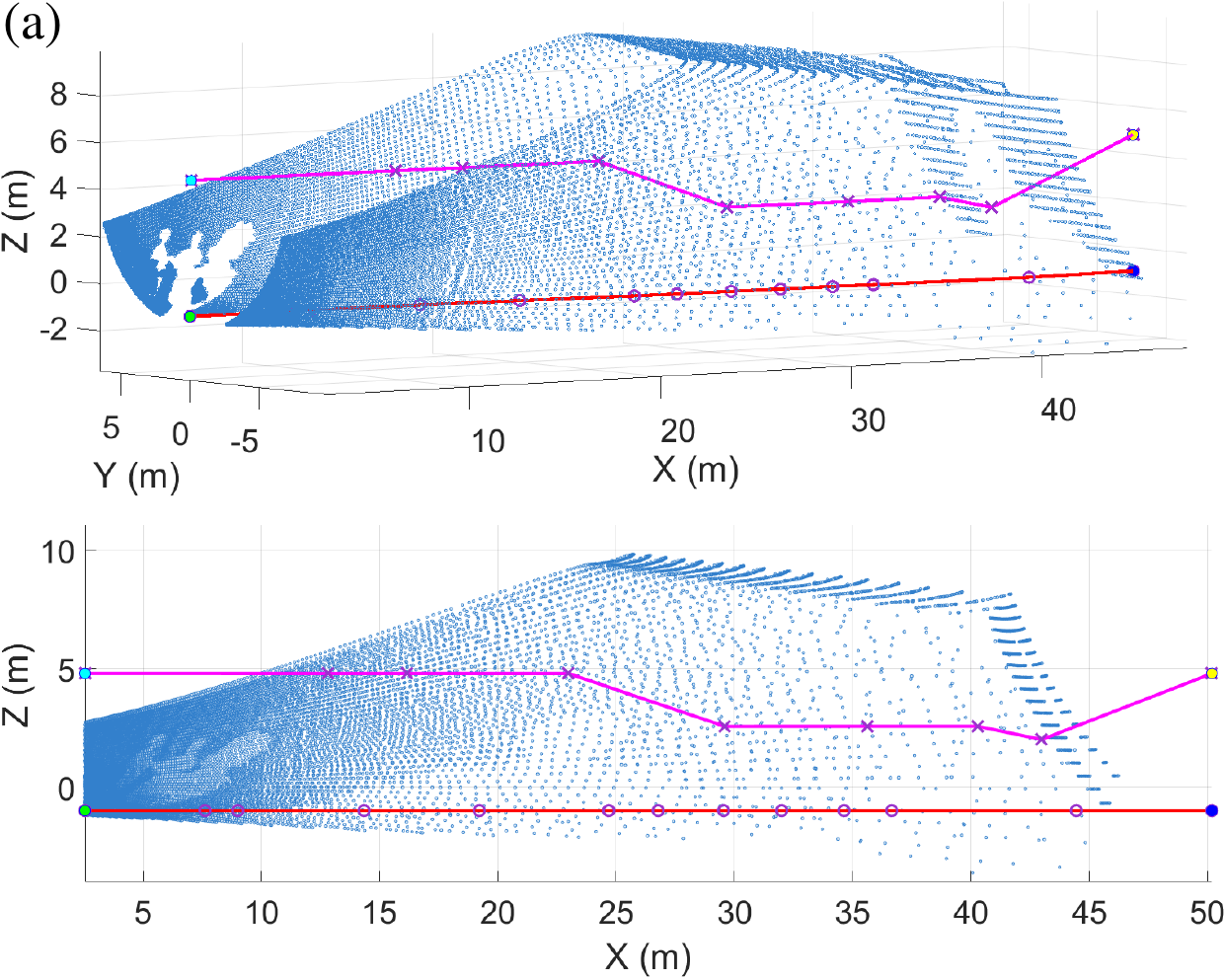} \\
	\includegraphics[width=0.85\linewidth]{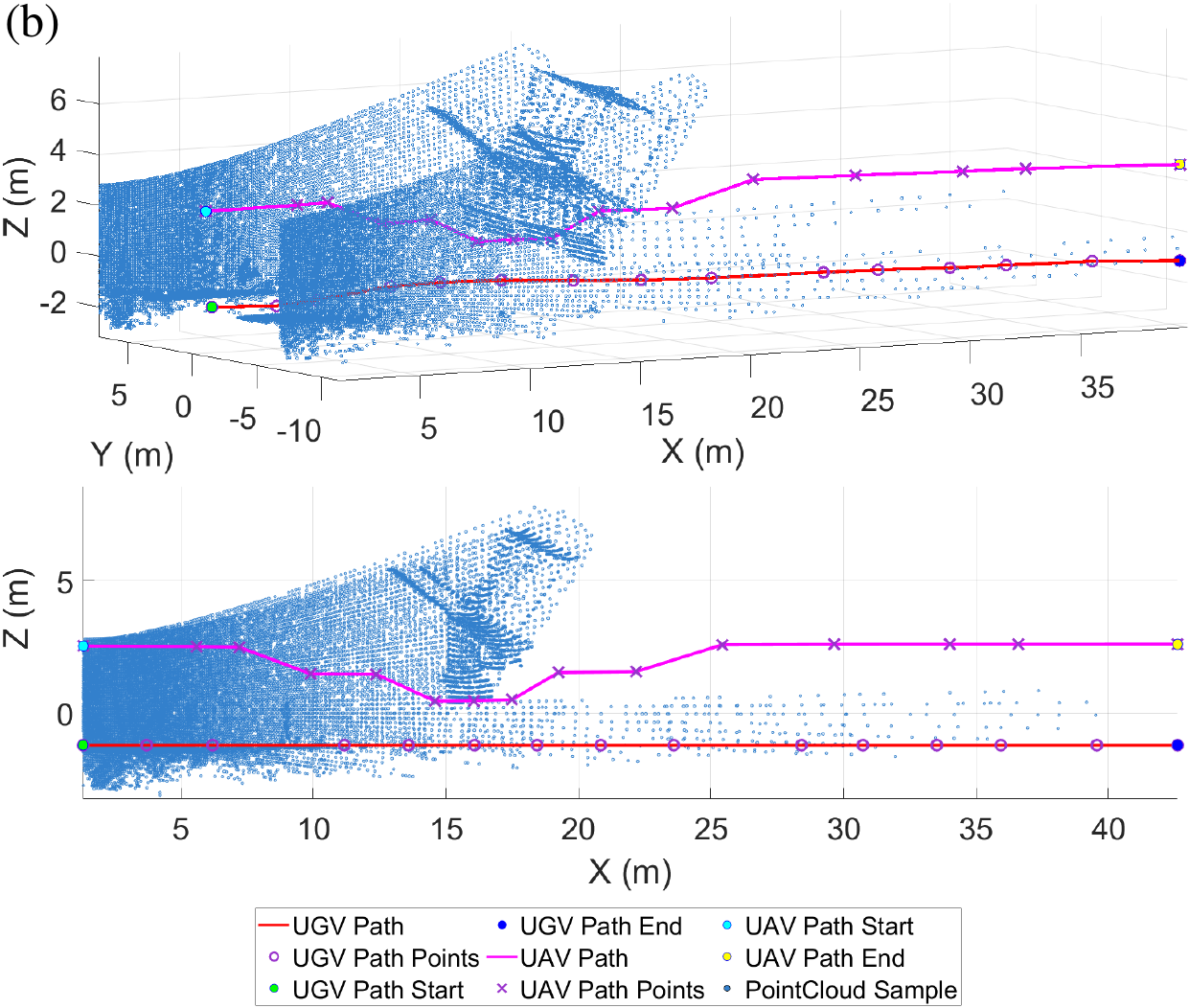} \\
	\caption{FLISP planning results in gate traversal scenarios: (a) Reverse traversal and (b) Forward traversal.\label{fig20}}
\end{figure}

For multi-obstacle scenarios (Fig.~\ref{fig19}), we adopted a safe-by-design protocol where a human operator carried the LiDAR to simulate the UGV path. This inadvertently subjected FLISP to a ``stress test'' with severe non-vehicular jitter.

\begin{itemize}
	\item \textbf{Path Length Truncation:} As detailed in Table \ref{tab:2}, while obstacle avoidance typically extends path length, these handheld trials (Scenarios 4 \& 6) paradoxically exhibited a reduction (45--48 m). We attribute this to severe jitter exacerbating the ``step loss'' combined with obstacle occlusion. Nevertheless, the total path length remained stable within 50 $\pm$ 5 m (Fig.~\ref{fig22}), maintaining sufficient lookahead.
	
	\item \textbf{Heading Estimation Resilience:} Crucially, despite these worst-case state inputs, the planner consistently recovered valid paths. The jitter challenged the system's orientation consistency, yet the \textbf{yaw estimation module} (Section IV-A-1) proved robust. By averaging normal vectors from multiple tunnel surfaces, it successfully filtered out high-frequency handheld disturbances, maintaining a stable heading reference.
\end{itemize}

\subsubsection{Runtime Efficiency and Simulation Consistency}
\begin{figure}
	\centerline{\includegraphics[width=3.4in]{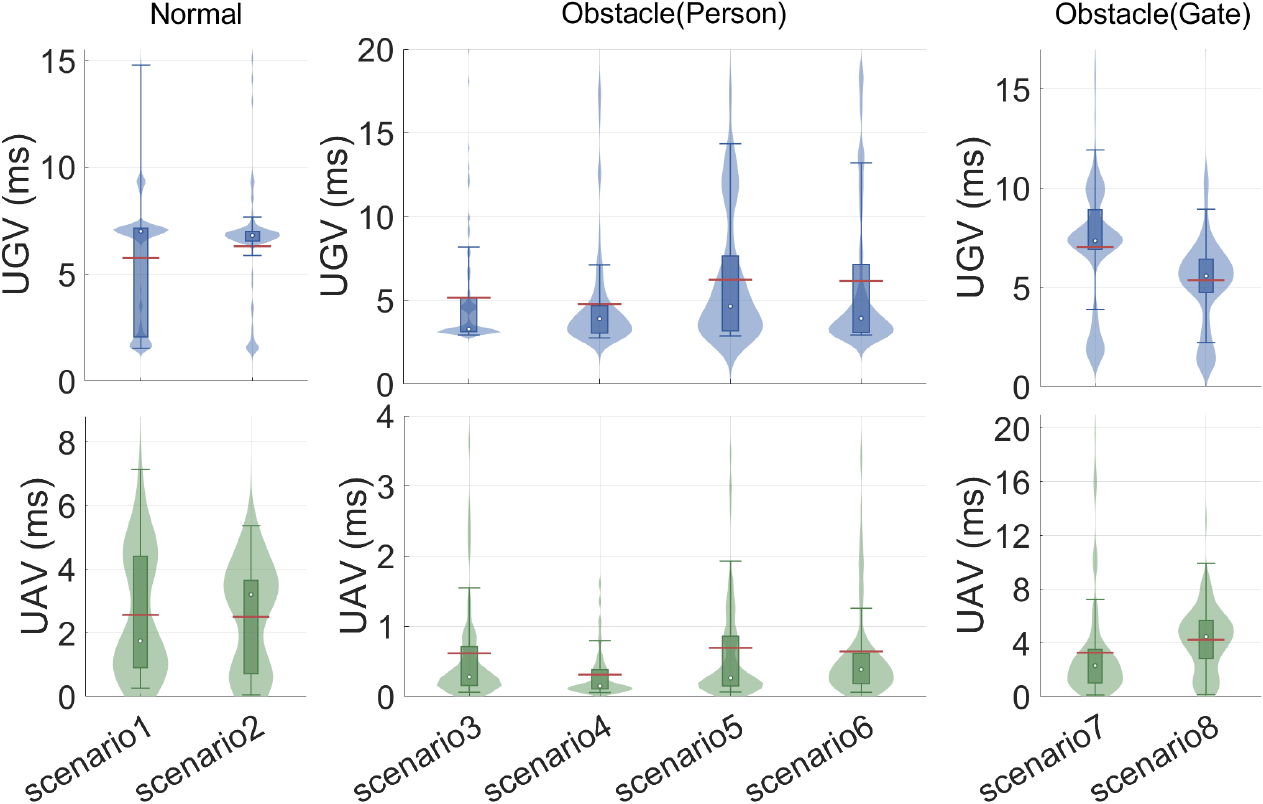}}  
	\caption{Runtime distribution across eight scenarios ($N=100$ trials each). Red lines: Mean; White dots: Median.\label{fig21}}
\end{figure}

\begin{figure*}
	\centerline{\includegraphics[width=6.5in]{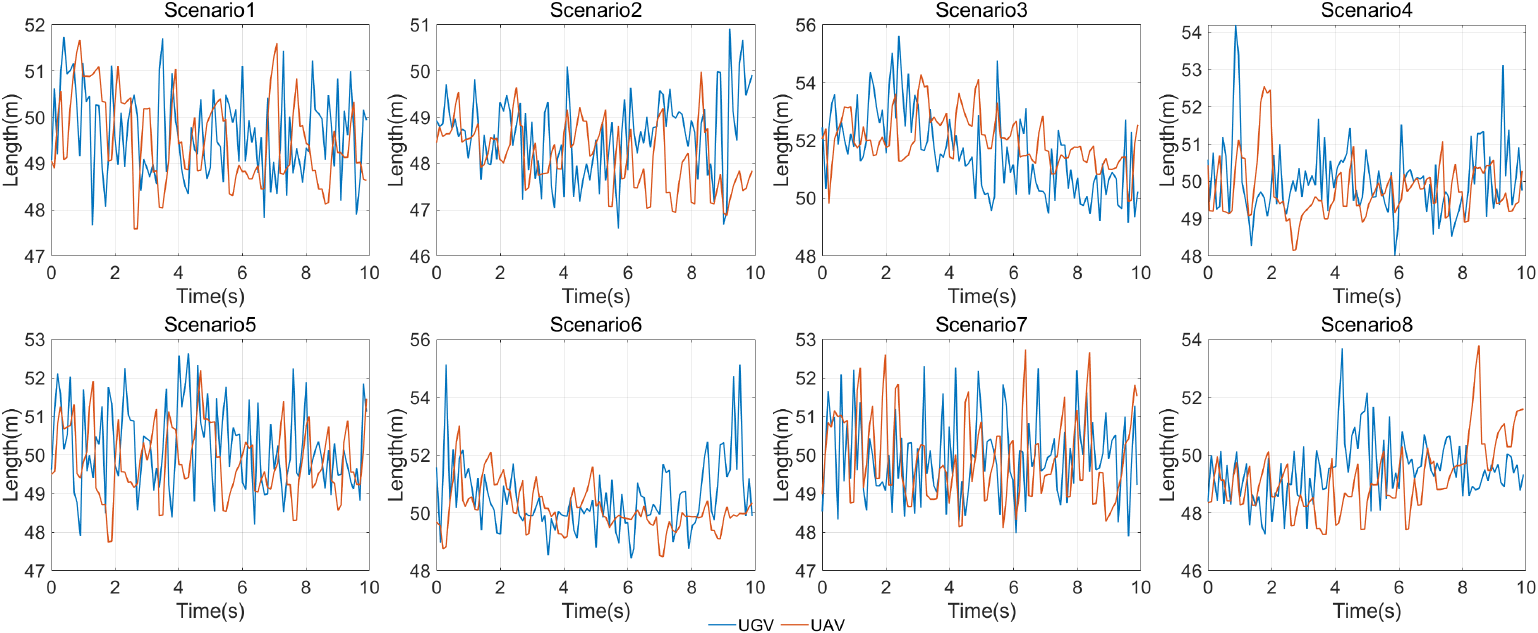}}  
	\caption{The length change of the paths planned by FLISP for the UGV and UAV across eight scenarios, evaluating the path validity rate. Notably, scenarios 4 and 6 were recorded using a handheld LiDAR, resulting in significant data jitter.\label{fig22}}
\end{figure*}

Quantitative metrics (Table \ref{tab:2}, Fig.~\ref{fig21}) confirm robust real-time performance and cross-domain consistency.

\begin{itemize}
	\item \textbf{Algorithmic Characteristics:} Standard planning times remained under 10 ms (UGV) and 3 ms (UAV). Notably, obstacles increased UAV latency (3--4 ms) due to optimization costs, whereas UGV latency remained invariant ($\approx$ 6 ms). This stability stems from the UGV's geometric architecture, which analytically resolves path structure and restricts FA optimization to the 1D -domain, minimizing sensitivity to clutter.
	
	\item \textbf{Simulation vs. Reality:} Real-world execution was slightly slower with wider distribution due to system overhead. However, path topology closely mirrors simulation. Despite ``step loss'' reducing actual path lengths, the system consistently maintained sufficient lookahead ($>45$ m) to ensure safe high-speed tracking.
\end{itemize}
	
\begin{table}[htbp]
	\centering
	\caption{Average analysis metrics over 100 field trials.}
	\label{tab:2}
	\begin{tabular}{|>{\centering\arraybackslash}m{1.2cm}||cc|cc|cc|}
		\hline
		\multirow{2}{*}{\textbf{Scenario \#}}
		& \multicolumn{2}{c|}{\textbf{Time [ms]}}
		& \multicolumn{2}{c|}{\textbf{Waypoints}}
		& \multicolumn{2}{c|}{\textbf{Length [m]}} \\
		\cline{2-7}
		& \textbf{UGV} & \textbf{UAV}
		& \textbf{UGV} & \textbf{UAV}
		& \textbf{UGV} & \textbf{UAV} \\
		\hline
		1 & 5.75 & 2.57 & 8.42 & 6.70 & 49.64 & 49.54 \\
		2 & 6.30 & 2.50 & 9.18 & 7.28 & 48.60 & 48.17 \\
		3 & 5.16 & 0.62 & 12.51 & 10.89 & 51.59 & 52.01 \\
		4 & 6.67 & 0.59 & 11.64 & 9.98 & 48.92 & 47.51 \\
		5 & 6.24 & 0.69 & 11.82 & 9.82 & 50.21 & 49.86 \\
		6 & 5.84 & 0.44 & 10.69 & 8.52 & 47.93 & 45.40 \\
		7 & 7.04 & 3.27 & 12.10 & 11.89 & 49.97 & 49.94 \\
		8 & 5.36 & 4.22 & 8.73 & 5.53 & 49.46 & 49.20 \\
		\hline
	\end{tabular}
\end{table}

\section{COMPARATIVE EXPERIMENTS AND DISCUSSION}
\label{sec:comparison}

This section evaluates the proposed FLISP framework against two state-of-the-art map-based paradigms using the 1.2 km tunnel dataset. We conduct a standardized offline benchmark to ensure identical sensor inputs and eliminate run-to-run physical variance.

\subsection{Experimental Setup and Baselines}
\label{sec:Setup}

We benchmark FLISP against two representative baselines:
1) \textbf{Method I (FLISP):} Our proposed mapless approach utilizing hierarchical polynomial fitting.
2) \textbf{Method II (Graph-based):} LIO-SAM \cite{bib58} tightly coupled with Informed RRT* \cite{bib59}. Informed RRT* is chosen for its ellipsoidal heuristic, theoretically accelerating convergence in confined tunnels.
3) \textbf{Method III (Filter-based):} Fast-LIO2 \cite{bib60} coupled with a Grid A* planner \cite{bib61}.

Operational parameters are detailed in Table \ref{tab:parameters}. These values derive from the sensitivity analysis in Section \ref{sec:sensitivity} to ensure each baseline performs at its Pareto-optimal limit.

\begin{table}[htbp]
	\centering
	\caption{Experimental Parameters and Constraints}
	\label{tab:parameters}
	\renewcommand{\arraystretch}{1.1} % 稍微压扁表格
	\setlength{\tabcolsep}{4pt}      % 稍微收紧列宽
	\begin{tabular}{|c||l|c|}
		\hline
		\textbf{Category} & \textbf{Parameter} & \textbf{Value} \\
		\hline
		\textbf{Physical} & UGV Dims ($L{\times}W{\times}H$) & $1.2 \times 0.8 \times 1.0$ m \\
		\hline
		\textbf{FLISP} & Planning Horizon & 50.0 m \\
		\hline
		\multirow{3}{*}{\textbf{Method II}} 
		& Planning Horizon & 25.0 m \\
		& Max Planning Time & 8.0 s \\
		& Success Threshold & 10.0 rad \\
		\hline
		\multirow{5}{*}{\textbf{Method III}} 
		& Grid Resolution & 0.1 m \\
		& Inflation Radius & 0.7 m \\
		& Planning Horizon & 30.0 m \\
		& Seal Radius & 1.0 m \\
		& Sliding Window & 50.0 m \\
		\hline
	\end{tabular}
\end{table}

To ensure fair benchmarking and real-time feasibility, both map-based baselines operate on a local sliding window extracted from the global SLAM map. This window is dynamically aligned with the vehicle's odometry (LIO-SAM for Method II, Fast-LIO2 for Method III) to ensure consistent sensor coverage in curved sections while bounding computational complexity.

To quantify performance (Section \ref{sec:quantitative}), we define five metrics:

\begin{itemize}
	\item \textbf{Path Smoothness ($\mathcal{S}$):} The accumulated angular deviation: $\mathcal{S} = \sum_{i=1}^{N-1} \arccos ( \hat{\mathbf{v}}_i \cdot \hat{\mathbf{v}}_{i+1} )$, where $\hat{\mathbf{v}}_i = \mathbf{v}_i / \|\mathbf{v}_i\|$ denotes the normalized segment vector. Lower $\mathcal{S}$ implies a smoother path.
	\item \textbf{System Latency ($T_{lat}$):} The operational bottleneck, calculated as the maximum wall-clock computation time per cycle: $T_{lat} = \max(T_{uav}, T_{ugv})$. This captures the total latency from sensor input to planning output.
	\item \textbf{Path Tortuosity ($\tau$):} Efficiency metric defined as the ratio of path length $\mathcal{L}_{path}$ to the Euclidean planning horizon $\mathcal{L}_{horizon}$: $\tau = \mathcal{L}_{path} / \mathcal{L}_{horizon}$. A value $\tau \approx 1$ indicates optimal straightness.
	\item \textbf{CPU Load:} Average single-core utilization and peak RAM usage.
	\item \textbf{Success Rate:} The percentage of frames where a valid collision-free path is generated. 
	
	\emph{Note: To decouple planning performance from SLAM drift, this is calculated only within segments where the state estimator provides valid localization.}
\end{itemize}

Finally, to address the stochasticity of SLAM front-ends, we report mean values across \textbf{5 independent trials} for each baseline. Additionally, to ensure consistent initialization, all experiments are aligned to start at $t=40$s, allowing sufficient time for IMU bias convergence while the UGV remains stationary at the tunnel entrance gate.

\subsection{Baseline Parameter Sensitivity Analysis}
\label{sec:sensitivity}

To ensure fair comparison, we calibrated the baselines via a two-stage protocol to identify their theoretical limits in the tunnel environment.

\subsubsection{Method II (Informed RRT*)}
We tuned the planner to decouple geometric constraints from computational latency.

\begin{figure}[htbp] 
	\centering
	\begin{tabular}{cc}
		\includegraphics[width=1.72in]{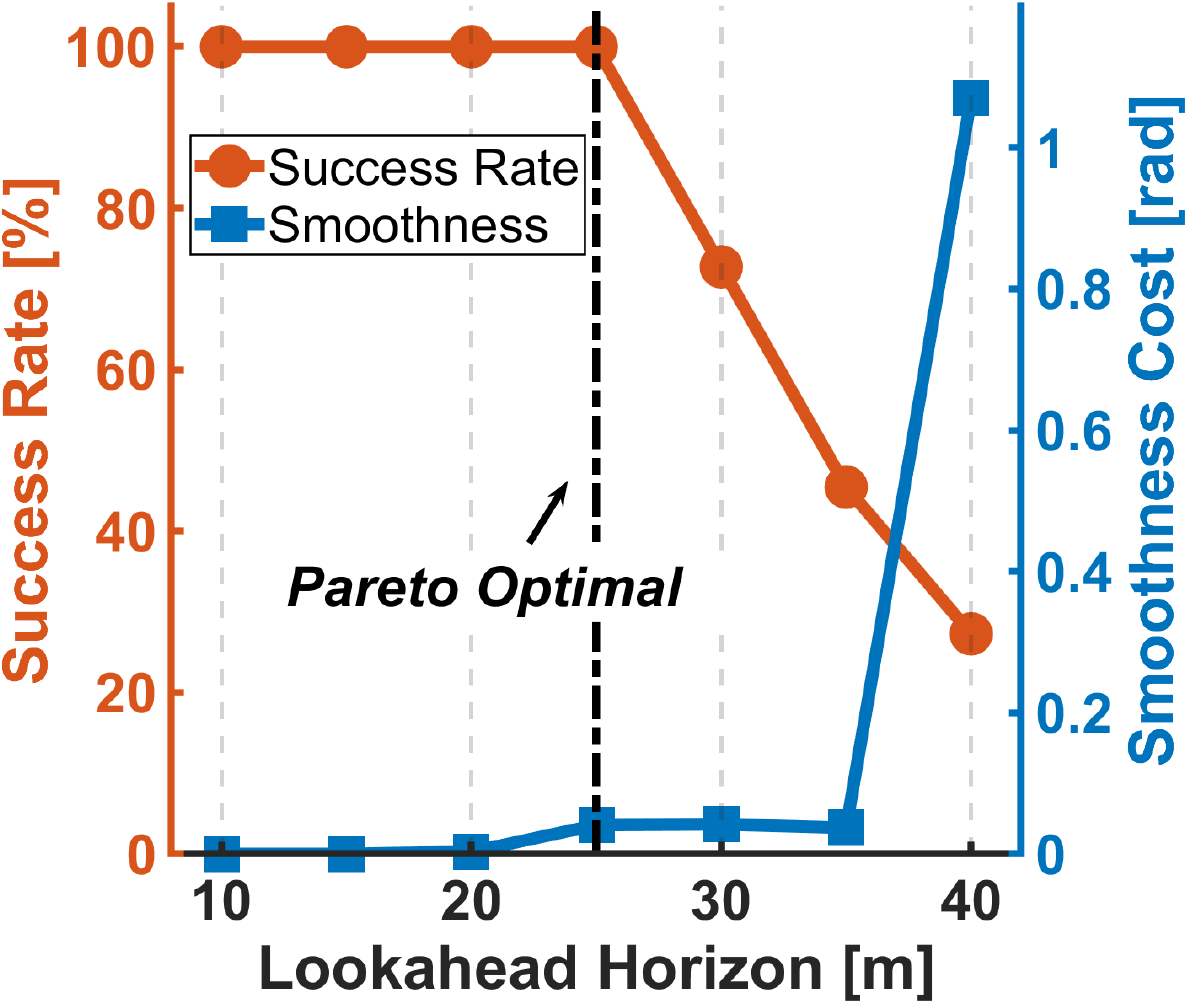} & 
		\includegraphics[width=1.44in]{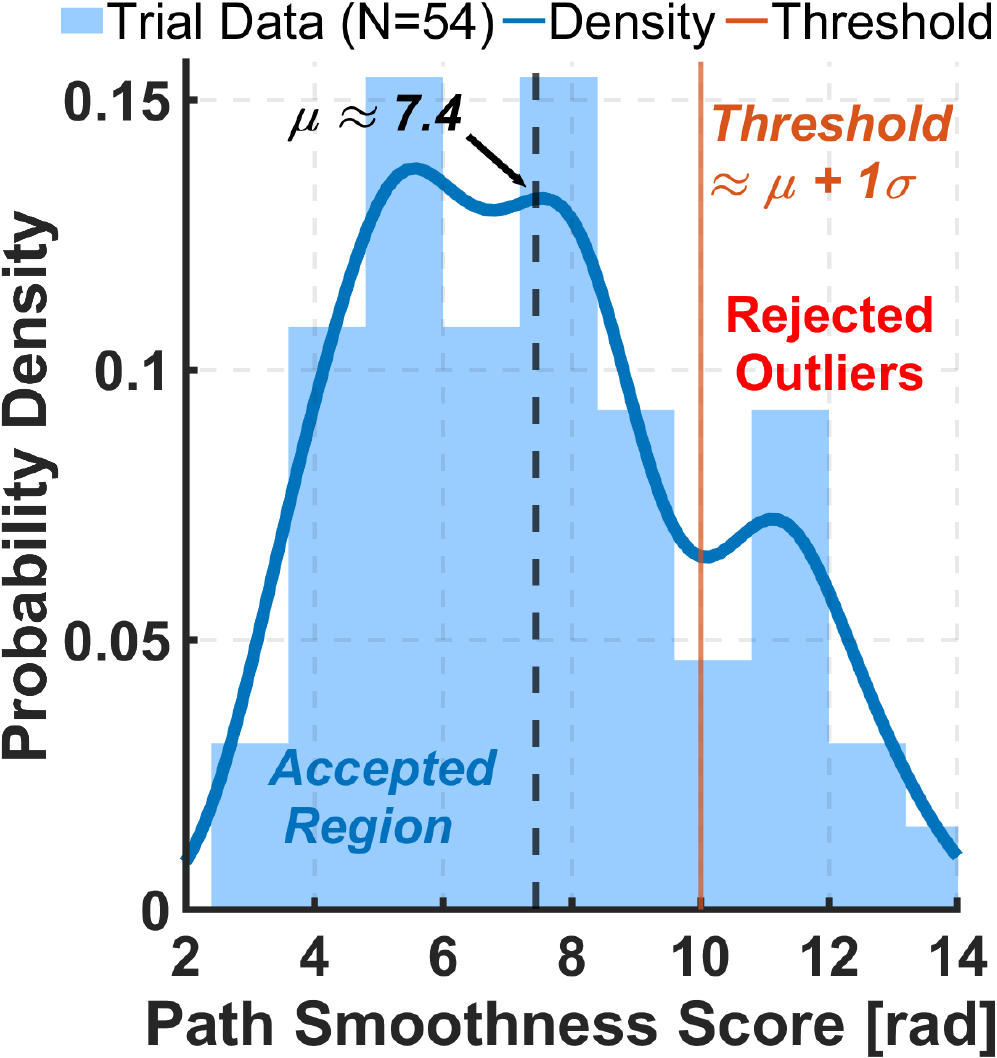} \\
		(a) & (b) 
	\end{tabular}
	\caption{Calibration for Method II. \textbf{(a)} Horizon: 25 m is selected as the Pareto optimal (100\% success). \textbf{(b)} Threshold: A 10.0 rad cutoff aligns with the statistical upper bound ($\mu + \sigma$), filtering stochastic outliers.
		\label{fig:rrt_sensitivity}}
\end{figure}

\textbf{Horizon Optimization:} To identify the physical planning limit, we tested lookahead distances (10--40 m) with a surplus budget (8.0 s). As shown in Fig.~\ref{fig:rrt_sensitivity}(a), success rates plummet beyond 30 m due to the exponential expansion of the sampling space. Thus, \textbf{25 m} was fixed as the Pareto-optimal horizon.

\textbf{Threshold Calibration:} Analyzing the smoothness distribution at this horizon (Fig.~\ref{fig:rrt_sensitivity}(b)) revealed a mean cost $\mu \approx 7.45$ rad. We established the Success Threshold at \textbf{10.0 rad} (statistical upper bound $\mu + \sigma$). Crucially, latency is recorded as the instant the \emph{first} valid path satisfying $\mathcal{S} < 10.0$ is found, representing the minimum time to viability.

\subsubsection{Method III (Grid A*)}

A similar protocol balanced map fidelity and range.

\begin{figure}
	\centerline{%
		\begin{tabular}{cc}
			\includegraphics[width=1.6in]{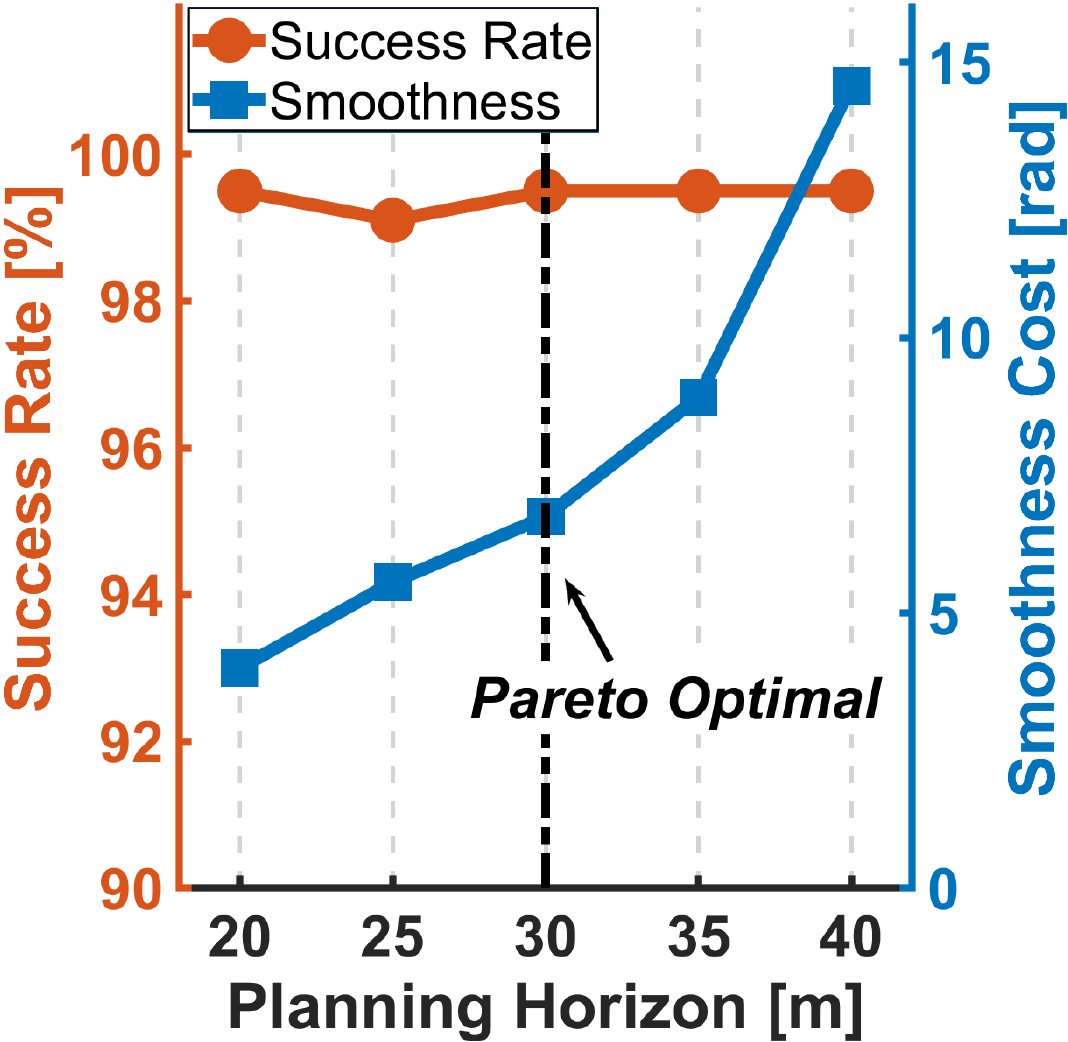} & \includegraphics[width=1.65in]{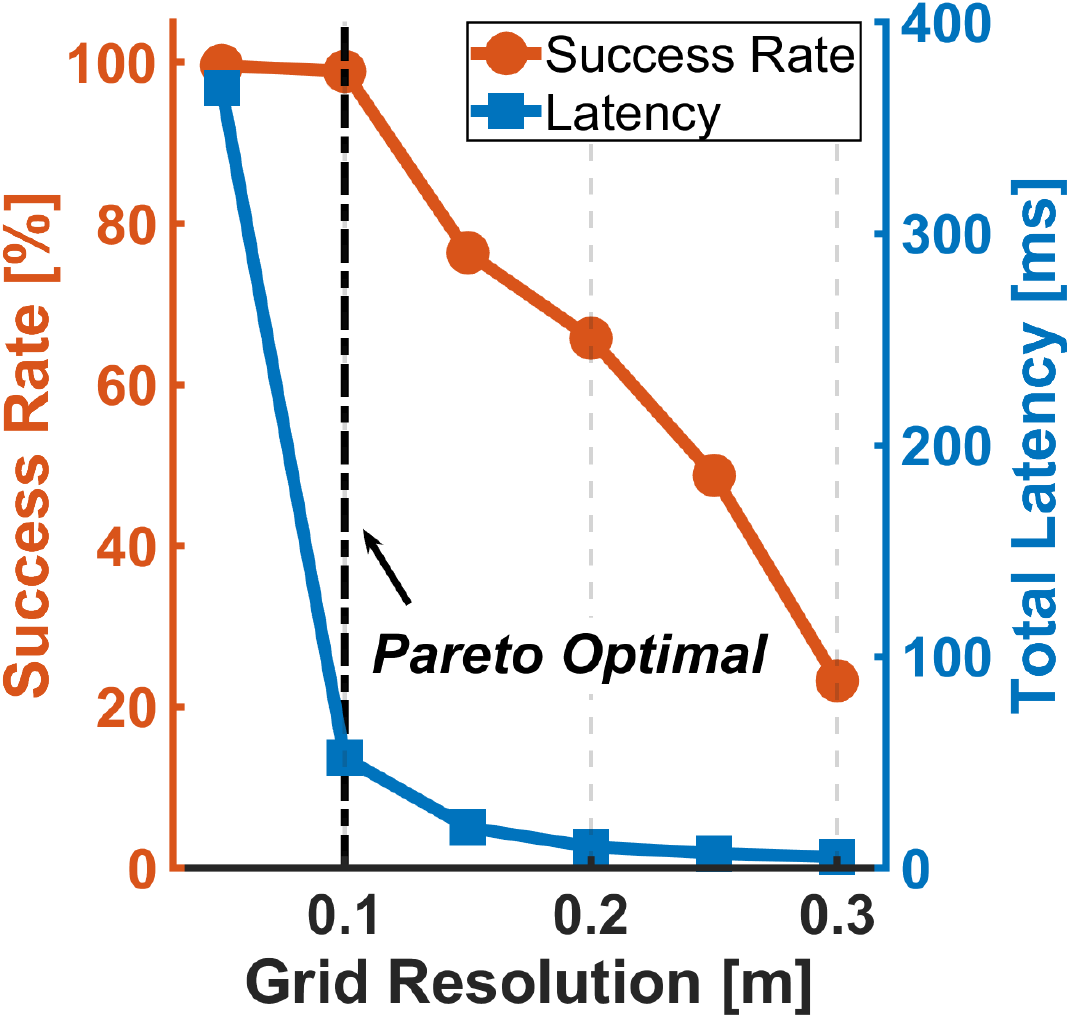} \\
			(a) & (b)
		\end{tabular}
	}
	\caption{Calibration for Method III. \textbf{(a)} Horizon: 30 m is identified as the geometric limit due to occlusion. \textbf{(b)} Resolution: 0.1 m balances system latency vs. obstacle dilation artifacts.\label{fig:Astar_sensitivity}}
\end{figure}

\textbf{Horizon Determination:} To isolate planning distance effects, we fixed the grid resolution (0.1 m) and swept horizons from 20 to 40 m. Fig.~\ref{fig:Astar_sensitivity}(a) reveals a geometric breakdown beyond 30 m, where sensor occlusion prevents the grid from capturing wall curvature ($\mathcal{S} > 14.5$). Thus, \textbf{30 m} was selected.

\textbf{Resolution Optimization:} With the fixed horizon (30 m), we swept grid resolutions (Fig.~\ref{fig:Astar_sensitivity}(b)). High fidelity (0.05 m) induced prohibitive latency ($\approx 369$ ms), while coarse resolutions ($>0.15$m) caused obstacle dilation, merging the floor with walls and blocking the passage. Thus, \textbf{0.1 m} was identified as the unique operating window.

\subsection{Qualitative Analysis of Navigation Behavior}
\label{sec:qualitative}

We analyze the evolution of mapping consistency and path quality on identical data.

\subsubsection{Phenomenology: Mapping and Path Evolution}

Distinct degradation patterns emerge in the deep tunnel (Supplementary Video, 02:10).
 
\textbf{Mapping Behavior:} While stable at the entrance, divergence occurs in the featureless curve.
 
\begin{figure}
	\centering
	\includegraphics[width=1.0\linewidth]{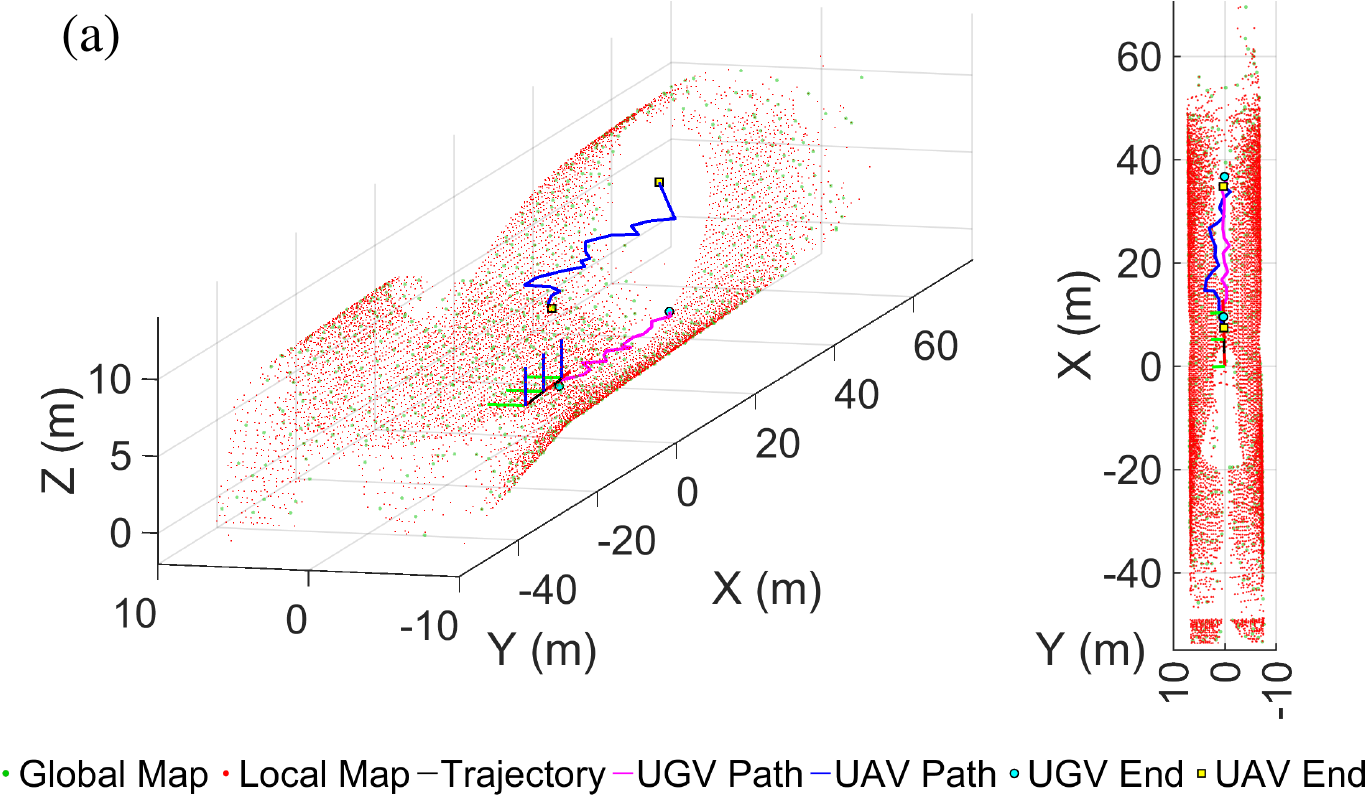} \\
	\vspace{2mm}
	\includegraphics[width=0.9\linewidth]{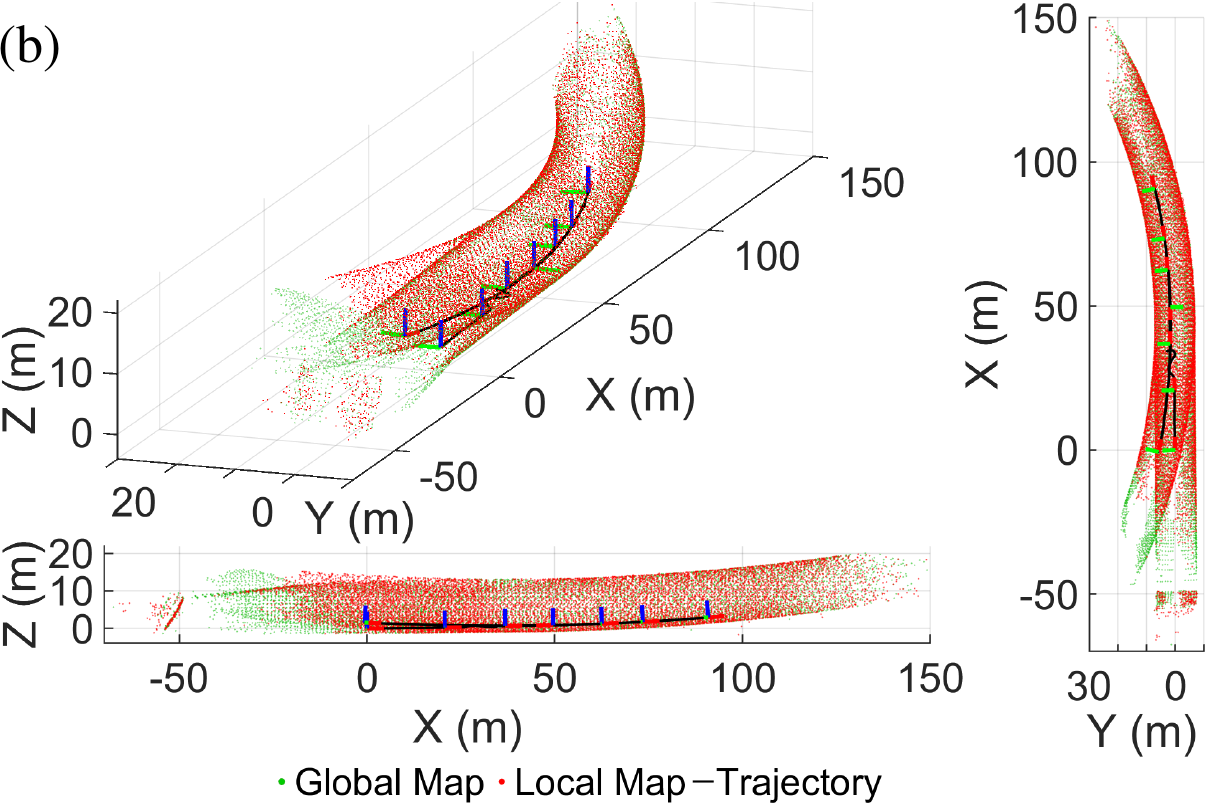} \\
	\caption{Performance evaluation of Method II (LIO-SAM + Informed RRT*). 
		(a) Effective Planning: Accurate mapping enables valid planning in the feature-rich entrance. 
		(b) Estimation Failure: In degenerate sections, drift causes map overlapping, oscillation, and erroneous elevation gain (upward curling), rendering navigation infeasible.
		\label{liosam1}}
\end{figure}

\begin{figure*}
	\centering
	\includegraphics[width=0.85\linewidth]{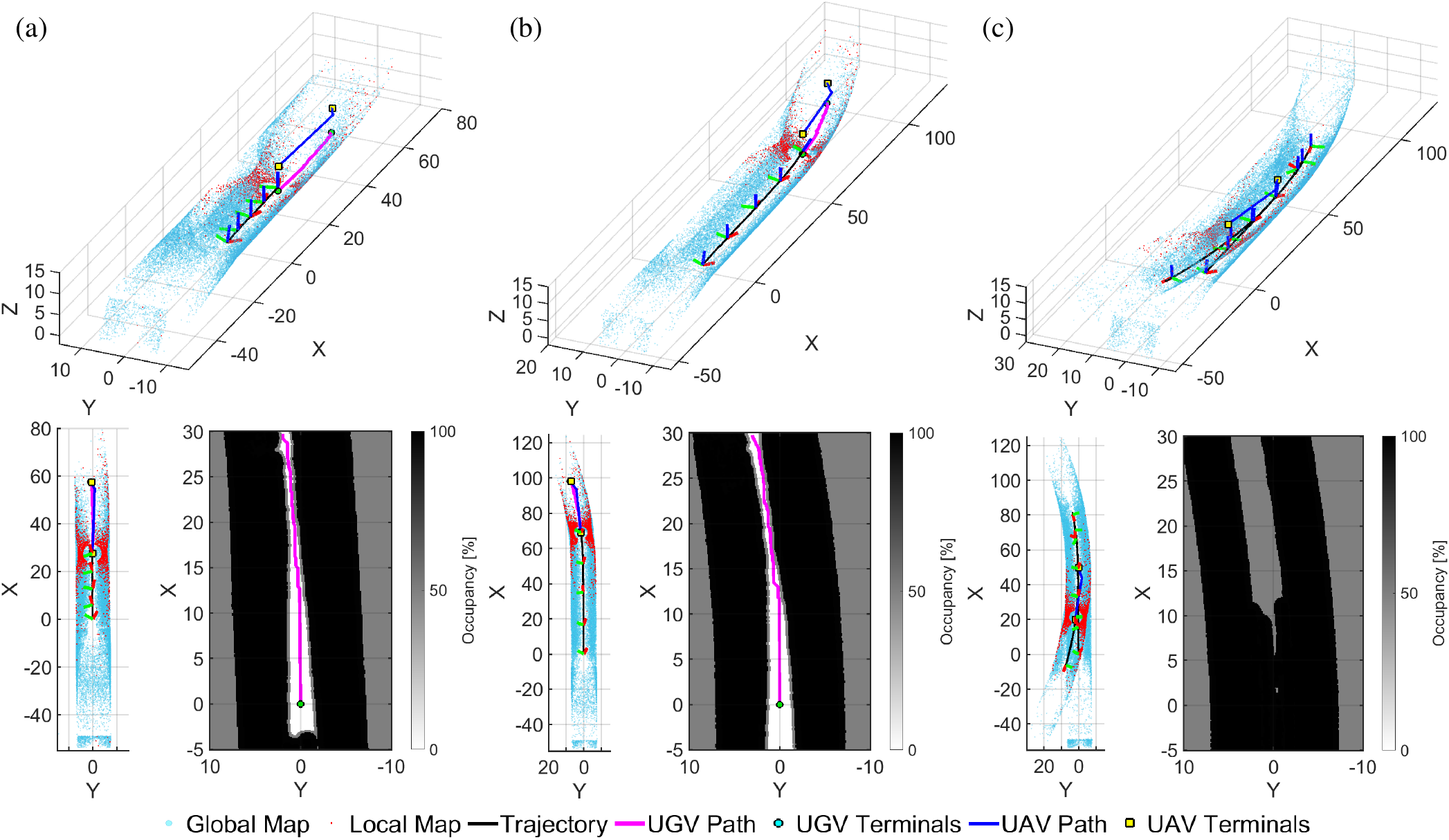}
	\caption{Performance evaluation of Method III (Fast-LIO2 + Grid A*). 
	\textbf{(a)} Start and \textbf{(b)} Turn phases show successful path generation. 
	\textbf{(c) Failure Phase:} Longitudinal stagnation causes structural overlap between straight and curved sections, creating a \textbf{``phantom blockage''} that renders navigation infeasible despite the physical passage being clear.
	\label{fastlio1}}
\end{figure*}

\begin{itemize}
	\item \textbf{Method II (LIO-SAM)} exhibits rapid instability. Its primary failure mode involves drift-induced \textbf{map overlapping} (Fig.~\ref{liosam1}(b)), where curved walls alias into the path. A distinct characteristic is severe vertical ($z$-axis) jitter. Furthermore, the failure is highly stochastic: in a subset of trials, pose estimation completely decouples (``fly-away''), rendering the map unusable.
	\item \textbf{Method III (Fast-LIO2)} maintains valid localization longer but converges to a similar overlapping artifact (Fig.~\ref{fastlio1}(c)). Crucially, the degradation dynamics differ: unlike the chaotic divergence of Method II, Method III suffers from \textbf{longitudinal stagnation}. The estimator ``slips'' along the tunnel axis, causing the virtual position to become static despite physical forward motion. This leads to continuous superposition of the local map, forming a ``phantom blockage'' without the unbounded spatial drift seen in Method II.
\end{itemize}

\textbf{Path Quality:} Even within the valid horizon, quality varies significantly.

\begin{itemize}
	\item \textbf{Method II} (Informed RRT*) circumvents the backward-planning issues inherent to standard RRT, but produces paths that remain tortuous and geometrically non-smooth (Fig.~\ref{liosam1}(a)). This stochastic irregularity is even more pronounced under the looser constraints applied to the UAV.
	\item \textbf{Method III} (A*) visually approximates the centerline (Fig.~\ref{fastlio1}(a, b)) but microscopic inspection reveals intrinsic ``sawtooth'' discretization artifacts. Critically, in scenarios with initial yaw deviation or curvature, it fails to correct cross-track error until traversing beyond the midpoint, frequently culminating in an \textbf{abrupt sharp turn} at the terminal state to satisfy the goal constraint.
\end{itemize}

\subsubsection{Root Cause and Dynamic Analysis}

We analyze these failure mechanisms, differentiating geometric causes (Mapping), dynamic consequences (Planning), and heterogeneous coordination limits.

\textbf{Mapping Failure Mechanism (The ``Anchor Loss''):} The contrast between the stable entrance phase and the failure in the deep tunnel indicates that performance relies on the visibility of the gate. This structure acts as a geometric anchor. When the gate becomes occluded by the curve, the system loses its primary longitudinal and yaw constraints. To verify this stems from feature loss rather than vibration, we conducted a control experiment using a handheld 128-beam LiDAR (Supplementary Video, 03:06). Despite significantly higher irregular motion (jitter), the resulting point cloud walls remained sharp and thin as the 64-beam UGV baseline, without thickening artifacts typical of vibration noise. However, the system failed \emph{more rapidly} due to the sensor's shorter effective range causing earlier anchor loss. This provides strong evidence that vibration is not the primary failure mechanism and confirms that geometric degeneracy is the fundamental limitation.

\textbf{Dynamic Implications of Path Artifacts:} Geometric artifacts introduce severe stability risks for UGV control.
\begin{itemize}
	\item \textbf{Instability and Inefficiency from Method II:} The stochastic irregularities necessitate frequent, erratic steering adjustments. Given the tunnel's concave profile, these lateral deviations cause the wide-bodied UGV to repeatedly ascend the sidewalls. This induces severe roll instability and high-frequency mechanical vibration, compromising data quality and causing excessive energy consumption due to redundant maneuvering and gravitational resistance.
	\item \textbf{Control Oscillation (Method III):} The ``sawtooth'' artifacts often fall within the actuator deadband, allowing cross-track error to accumulate until the UGV physically drifts onto the wall. The subsequent high-magnitude correction, compounded by inertia, triggers control oscillation. This poses a severe rollover risk, escalating at higher inspection speeds.
\end{itemize}

\textbf{Limitations of Local and Uncoordinated Paradigms:} 
To rigorously evaluate alternatives to SLAM-based mapping, we additionally benchmarked a pure local planner (TEB on streaming point clouds). While successfully bypassing SLAM drift, it exposed three fundamental vulnerabilities inherent to uncoordinated 2D methods in our specific environment (detailed empirical analyses and visual proofs are provided in Appendix A):

\begin{itemize}
	\item \textbf{Local Occlusion Trap (UGV Straight-Line Artifact):} Due to the inherent roof-shadow blind spot of UGV-mounted LiDAR, a cost-free void forms directly ahead. Pure local planners, optimizing for minimal immediate cost, invariably generate a reactive straight-line path through this void. In curved segments, high-frequency execution of these straight lines actively drives the UGV toward the outer wall. FLISP circumvents this by performing a macroscopic 50m polynomial fit, analytically guaranteeing an initial curvature-aligned arc that is impervious to local sensor blockages.
	
	\item \textbf{Topological Deadlock under Roll Perturbation:} Navigation relying on 2D costmap projections faces a catastrophic edge case. If the UGV mechanically slips onto the curved sidewall, the 2D projection casts the surrounding walls as an impenetrable forward obstacle, isolating the vehicle from the valid center channel. Even with IMU-corrected orientation, 2D local planners become topologically deadlocked. FLISP inherently avoids this by integrating the IMU gravity vector into its global 3D geometric evaluation, synthesizing a viable 3D recovery path back to the tunnel floor.
	
	\item \textbf{The Heterogeneous Lookahead Paradox:} Uncoordinated local planners face an irreconcilable dilemma regarding the planning horizon in UAV-UGV operations. A short horizon (e.g., 5m) forces the UAV, operating in the obstacle-free upper tunnel, to generate high-frequency, uncoordinated zigzags, inducing severe flight oscillation. Conversely, extending the horizon (e.g., 10m) in non-convex blind curves forces the optimization solver into conflicting soft constraints. Attempting to balance obstacle clearance and path continuity, the solver frequently generates abrupt mid-path kinks (similar to A* jaggedness), which are dynamically infeasible for heavy UGVs. Employing asymmetric horizons for each platform fundamentally destroys spatial synchronization. 
\end{itemize}

In contrast, FLISP establishes a robust foundation for heterogeneous collaborative planning. By hierarchically deriving the UAV path from the UGV's 50m geometric baseline, FLISP inherently resolves the lookahead paradox and enforces strict spatial synchronization. Furthermore, unlike optimization-based planners averaging $\sim100$ ms per cycle, FLISP operates at a highly efficient $7$ ms, effectively eliminating control phase lag. Ultimately, by anchoring trajectories directly to the global 3D tunnel geometry rather than reactive 2D costmaps, FLISP guarantees dynamic compliance and immunizes the system against local map aliasing, cumulative drift, and topological deadlocks.

\subsection{Quantitative Benchmark and Efficiency Analysis}
\label{sec:quantitative}

Complementing qualitative insights, we quantify system performance across the 1.2 km tunnel benchmark. As defined in Section \ref{sec:Setup}, data in Table \ref{tab:quantitative_results} represents mean performance over 5 independent trials. Crucially, to ensure fair comparison of planning capabilities, baseline statistics (Method II and III) are calculated \emph{exclusively within valid localization horizons}. This prevents including invalid timeout data from failure phases, presenting baselines at their operational best.

\begin{table}[htbp]
	\centering
	\caption{Quantitative Benchmark Results}
	\label{tab:quantitative_results}
	
	\renewcommand{\arraystretch}{1.1} 
	\setlength{\tabcolsep}{3pt}        
	
	\begin{tabular}{|l||c|c|c|}
		\hline
		\textbf{Metric} & \textbf{\begin{tabular}{@{}c@{}}Method I\end{tabular}} & \textbf{\begin{tabular}{@{}c@{}}Method II\end{tabular}} & \textbf{\begin{tabular}{@{}c@{}}Method III\end{tabular}} \\
		\hline
		
		\textbf{Success Rate} [\%] & \textbf{100.0} & 98.5 & 99.6 \\
		\textbf{Path Smoothness} [rad] & \textbf{0.016} & 9.638 & 6.068 \\
		\textbf{Path Tortuosity} [-] & 1.03 & 1.42 & \textbf{1.02} \\
		\textbf{Latency ($T_{lat}$)} [ms] & \textbf{7.05 $\pm$ 4.26} & 6492.61 $\pm$ 2973.8 & 49.93 $\pm$ 8.01 \\
		\textbf{CPU Load} [\%] & \textbf{1.37 $\pm$ 0.41} & 11.18 $\pm$ 3.94 & 8.72 $\pm$ 1.20\\
		\textbf{RAM Usage} [MB] & \textbf{39.7} & 760.5 & 1037.3 \\
		\hline
		
		% 下面是标准的 IEEE 官方表格脚注写法
		\multicolumn{4}{p{240pt}}{\footnotesize \textit{\textbf{Definitions:} \textbf{Method I:} FLISP (Ours); \textbf{Method II:} Graph-based (LIO-SAM + Informed RRT*); \textbf{Method III:} Filter-based (Fast-LIO2 + A*).}} \\
		\multicolumn{4}{p{240pt}}{\footnotesize \textit{\textbf{Note:} Data represents mean values over $N=5$ trials. Metrics for Baselines are calculated only within valid localization segments.}} \\
	\end{tabular}
\end{table}

\textbf{Performance and Reliability:}
Method I (FLISP) achieves a 100\% Success Rate. In contrast, map-based baselines exhibit occasional failures (98.5\% and 99.6\%) due to the mapping degradation discussed in Section \ref{sec:qualitative}.

\textbf{Path Quality Trade-off:}
A distinct trade-off is observed between macroscopic efficiency and microscopic continuity. Method III achieves the lowest Path Tortuosity ($\tau=1.02$), marginally outperforming FLISP ($\tau=1.03$). However, this marginal reduction is achieved at the expense of significantly poorer smoothness (6.068 rad vs 0.016 rad). This discrepancy arises because Method III generates rigid, piecewise-linear paths inherent to grid-based search. While geometrically shorter, these paths contain frequent angular discontinuities (``zig-zags'') detrimental to tracking. Conversely, FLISP prioritizes high-order continuity, accepting a negligible increase in path length to ensure the smooth control inputs essential for high-speed tracking. Method II exhibits the poorest quality in both metrics. The dynamic implications of these path artifacts are analyzed in Section \ref{sec:qualitative}.

\begin{figure}
	\centerline{\includegraphics[width=3.2in]{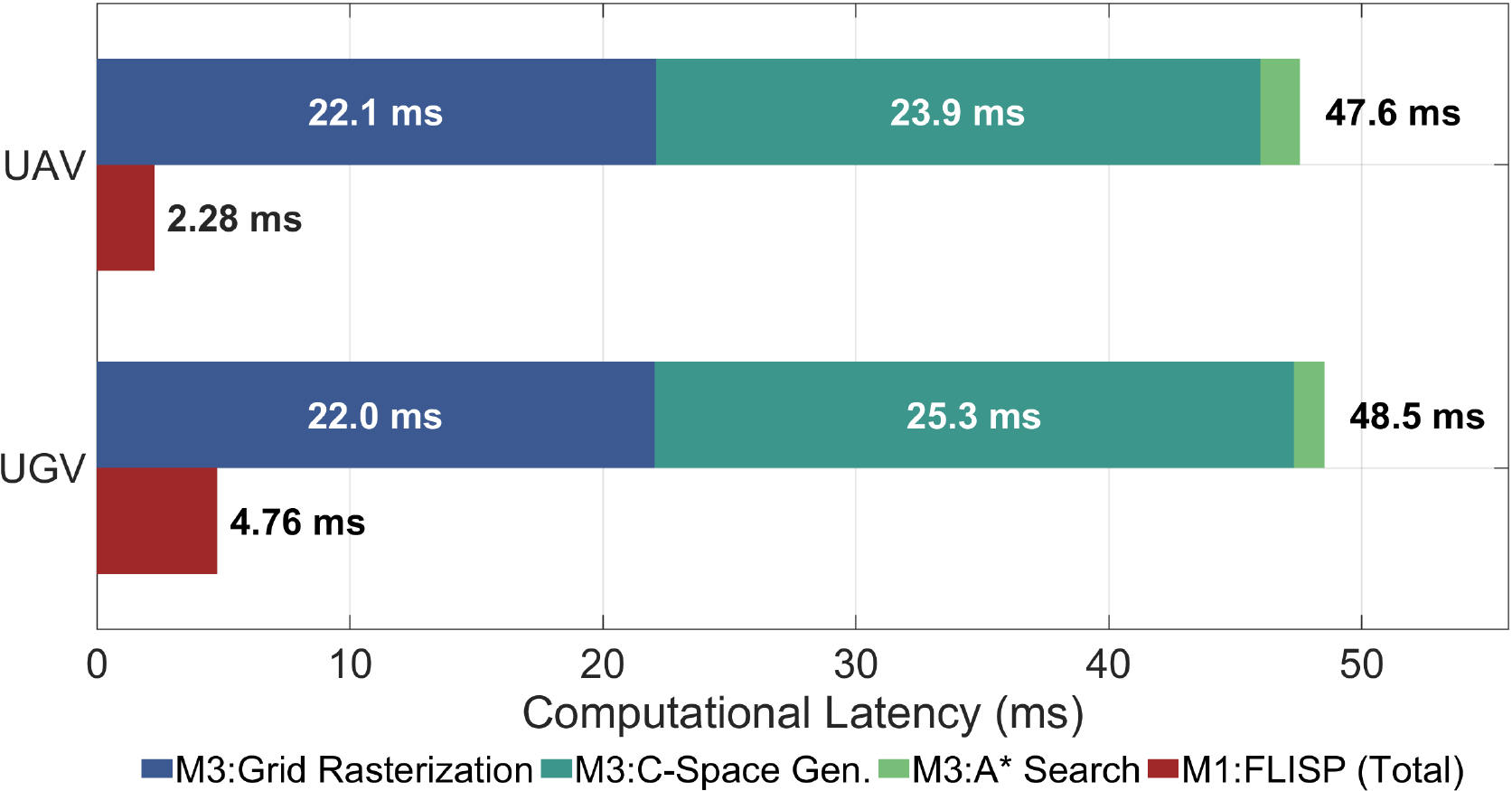}}  
	\caption{Computational Latency Breakdown (Method III vs. Method I).
		For the map-based Method III, total latency is dominated by Grid Rasterization (Blue) and C-Space Generation (Teal), acting as expensive pre-requisites for the actual A* Search (Green). 
		These map maintenance steps are mandated by the 0.1 m grid resolution for safety. 
		In contrast, FLISP (Burgundy) bypasses dense map updates, operating on sparse geometric features to achieve an order-of-magnitude latency reduction.
		\label{time_breakdown}}
\end{figure}

\textbf{Computational Efficiency Breakdown:}
Method I demonstrates a decisive advantage in computational efficiency, reducing System Latency to 7.05 ms, a $\sim$7$\times$ speedup over Method III (49.93 ms) and three orders of magnitude over Method II (Table \ref{tab:quantitative_results}). To elucidate the source of this efficiency, we decompose the pipeline in Fig.~\ref{time_breakdown}.

The breakdown reveals a critical insight: the computational bottleneck of map-based planning is not the graph search itself, but the map maintenance overhead. In Method III, the actual A* search (shown in green) accounts for a minor fraction of the total latency. The majority of the computational budget ($>80\%$) is consumed by Grid Rasterization (ray-casting from sparse LiDAR points to dense voxels) and C-Space Generation (obstacle inflation). This high overhead is the algorithmic penalty for maintaining a high-resolution (0.1 m) occupancy grid, which is geometrically necessary to ensure safety in narrow tunnels but computationally expensive to update in real-time.

In contrast, FLISP (shown in burgundy) entirely eliminates this intermediate representation. Formulating path generation as direct geometric fitting on the sparse point cloud bypasses expensive rasterization and inflation stages. This architectural advantage allows FLISP to maintain negligible latency, effectively decoupling planning speed from environmental resolution requirements.

\textbf{Resource Consumption and Real-time Feasibility:}
Beyond latency, the proposed framework exhibits a minimal resource footprint, which is decisive for autonomous inspection missions. On SWaP-constrained onboard platforms, high-fidelity data acquisition and high-frequency flight control already consume the majority of the computational budget. Consequently, the planning module must operate with a negligible footprint to avoid process starvation or resource contention that could jeopardize stability.

Method I maintains an exceptionally low CPU Load (1.37\%) and RAM Usage (39.7 MB), leaving ample resources for these concurrent mission-critical tasks. In sharp contrast, the map-based baselines incur heavy penalties:

\begin{itemize} 
	\item \textbf{Memory Overhead:} Both baselines require substantial memory. Method III consumes 1037.3 MB, a demand dominated by the maintenance of the \textbf{Global Occupancy Grid} and obstacle inflation layers required for A* search, rather than the lightweight Fast-LIO2 estimator itself. This cumulative memory consumption poses a severe risk of exhaustion during long-endurance exploration.
	
	\item \textbf{Update Frequency Disparity:} The computational bottleneck directly dictates the control bandwidth. 
	Both \textbf{Method I (FLISP)} and \textbf{Method III (A*)} are computationally efficient enough to saturate the set ROS publication frequency, ensuring continuous feedback control. 
	However, \textbf{Method II (Informed RRT*)} is severely constrained by its sampling complexity. Statistical analysis reveals it operates at an average frequency of only 0.16 Hz. This implies the system executes in an intermittent ``open-loop'' fashion for extended durations ($\approx$ 6 seconds per planning cycle), fundamentally explaining its inability to correct path deviations in the degenerate tunnel environment.

\end{itemize}

These results validate FLISP as uniquely suited for resource-constrained platforms, ensuring agility without compromising concurrent perception and control stability.

Furthermore, it is essential to contextualize the framework's performance against inherent physical limitations. In severe mud or water accumulation, or when driving diagonally on the curved floor causes a wheel to lose ground contact, the UGV's wheel encoders experience unavoidable slip, leading to momentary local odometry failure. However, FLISP's mapless, high-frequency replanning architecture effectively mitigates this by implicitly resetting the state relative to the latest valid LiDAR frame, preventing the catastrophic global map divergence seen in traditional SLAM pipelines. Furthermore, while high-frequency perception inevitably introduces minor waypoint jitter, the current physical deployment smooths these artifacts via the heavy chassis's mechanical low-pass filtering properties and path profiling, ensuring stable physical execution.

\section{CONCLUSION}
In this paper, we presented FLISP, a collaborative planning framework designed to enable the automated inspection of large-scale, geometrically degenerate infrastructure. Addressing the limitations of traditional robotics in hazardous hydropower tunnels, we proposed a lightweight mapless paradigm based on hierarchical polynomial fitting. By directly translating raw sensor streams from a single LiDAR-IMU suite into safe, synchronized trajectories, our system effectively decouples path generation from global state estimation. This architectural shift mitigates the drift and consistency issues inherent to SLAM-dependent planners, providing a robust solution for feature-poor conditions.

The system's practical feasibility and efficiency were validated through comprehensive benchmarks and a successful deployment in a 1.2 km operational tunnel. The comparative analysis demonstrated that FLISP significantly outperforms state-of-the-art map-based baselines. While the grid-based baseline suffered from excessive memory consumption due to map rasterization and the sampling-based baseline failed to maintain real-time control frequencies, FLISP achieved a 100\% success rate with negligible latency and a minimal resource footprint. Furthermore, empirical analysis confirmed that substituting our geometric approach with purely reactive local planners leads to unrecoverable topological deadlocks and synchronization failures. These results confirm our core hypothesis that for repetitive and resource-constrained environments, a mapless approach is a prerequisite for ensuring both system agility and operational stability.

Future work will expand the framework's versatility beyond linear structures by integrating a topological decision-making layer for complex junctions. Furthermore, while FLISP provides robust high-level mapless path guidance, executing these trajectories seamlessly under highly dynamic constraints requires rigorous low-level control. To this end, our recent parallel work \cite{bib_new_cbf} developed a control barrier function (CBF)-based consensus tracking controller in simulation. Since FLISP generates the reference targets and the CBF controller ensures safety-critical execution, the two frameworks are highly complementary. Integrating this theoretically guaranteed controller with FLISP on the physical platform to evaluate their combined performance represents our immediate next step.

\section*{APPENDIX: EMPIRICAL ANALYSIS OF LOCAL PLANNER}
To justify replacing reactive local planners with our macroscopic geometric approach, we benchmarked the Timed Elastic Band (TEB) algorithm against FLISP using identical LiDAR streams and safety parameters (e.g., $0.1\text{m}$ resolution, $0.7\text{m}$ inflation). This appendix evaluates their kinematic behavior, topological resilience, and computational efficiency in constrained 3D tunnels.

\subsection*{A. The Heterogeneous Lookahead Paradox}
\begin{figure}[htbp]
	\centering
	\includegraphics[width=0.9\linewidth]{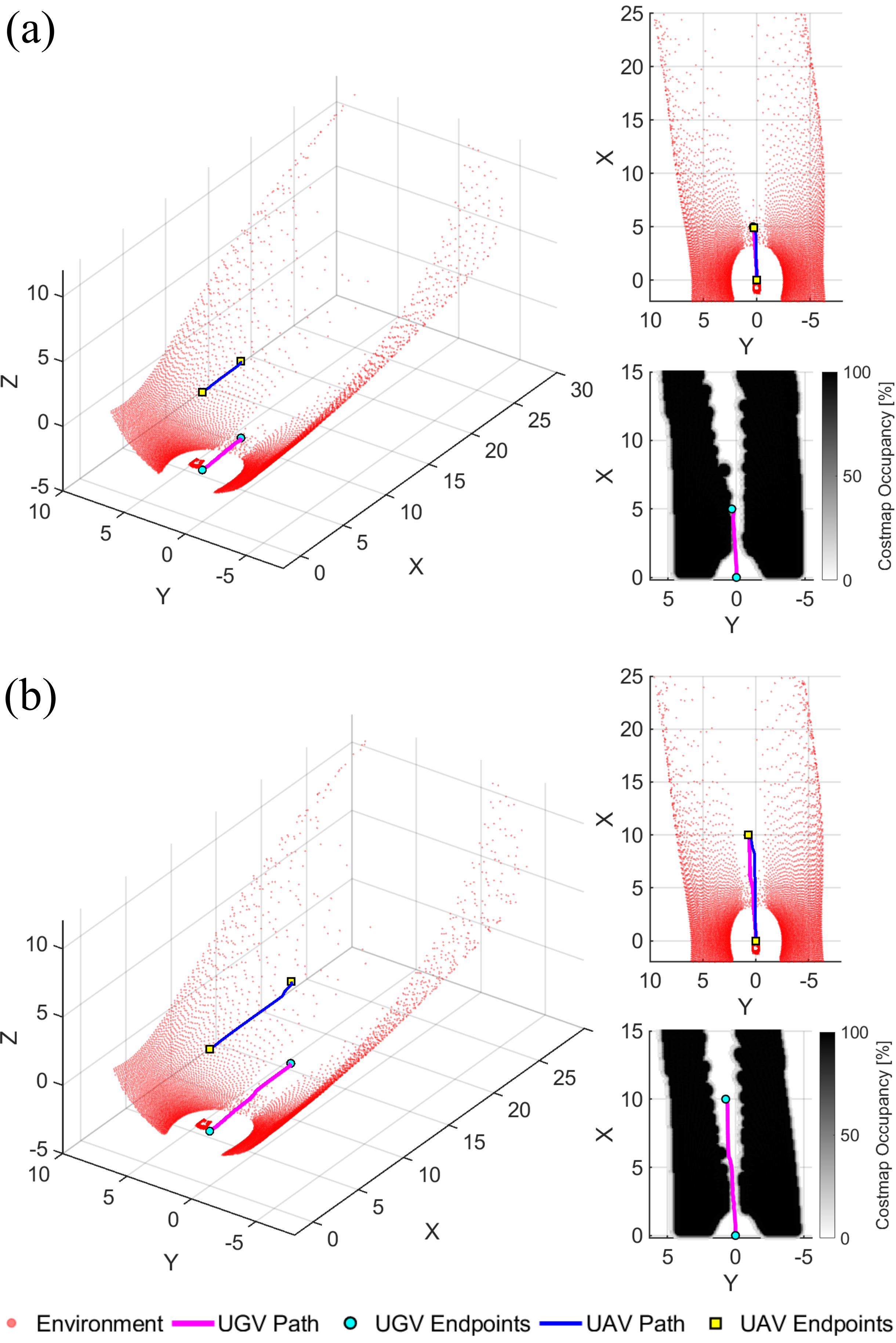}
	\caption{Evaluation of TEB local planner in curved tunnels. The roof-shadow blind spot induces rigid straight-line planning at the start of the trajectory, driving the UGV towards the outer wall regardless of the chosen horizon. \textbf{Top (5m):} Short horizons cause uncoordinated and oscillatory UAV flight and high-frequency UGV control jitter. \textbf{Bottom (10m):} Extended horizons force soft-constraint conflicts, resulting in dynamically infeasible mid-path kinks.}
	\label{fig:teb_paradox}
\end{figure}

Tuning the lookahead horizon for 2D local planners exposes a fundamental kinematic dilemma for heterogeneous teams.
\begin{itemize}
	\item \textbf{Short Horizon ($5\text{m}$):} Limited lookahead produces jittery and unstable paths. For the UGV, this local path jitter translates into high-frequency steering control inputs, which the heavy chassis cannot physically execute, rendering the path unusable. Conversely, for the UAV, the projected aerial costmap is highly open, causing the planner to simply generate reactive straight-line segments. Under continuous high-frequency replanning, these short straight segments piece together into a jagged polyline, resulting in non-smooth and severely oscillatory flight.
	\item \textbf{Long Horizon ($10\text{m}$):} Extending the horizon to improve anticipation forces the optimization solver into conflicting soft constraints within the non-convex curved tunnel. To balance obstacle clearance and goal progression, TEB frequently compromises by generating abrupt mid-path kinks (Fig. \ref{fig:teb_paradox}, bottom). In sharp contrast to the high-frequency jitter of short horizons, these mid-path kinks subject the UGV to sudden, large steering control inputs, severely destabilizing its dynamic control.
\end{itemize}

\subsection*{B. The Local Occlusion Trap (Roof-Shadow Blind Spot)}
Independent of the planning horizon, the UGV-mounted LiDAR inherently suffers from a roof-shadow blind spot, creating a perceived ``cost-free'' void directly ahead. To minimize optimization costs, TEB inherently tends to pull and straighten the trajectory through this void at the beginning of the planning cycle. 

While benign in straight sections, executing these rigid straight segments in curved pipes actively drives the UGV toward the outer wall, inevitably leading to a rollover hazard. Notably, while SLAM-based methods eventually suffer from global map collapse in deep curves, their cumulative mapping effectively fills this physical void, thereby successfully avoiding this specific straight-line artifact prior to localization failure.

\subsection*{C. Topological Deadlock under Roll Perturbation}
\begin{figure}[htbp]
	\centering
	\includegraphics[width=1\linewidth]{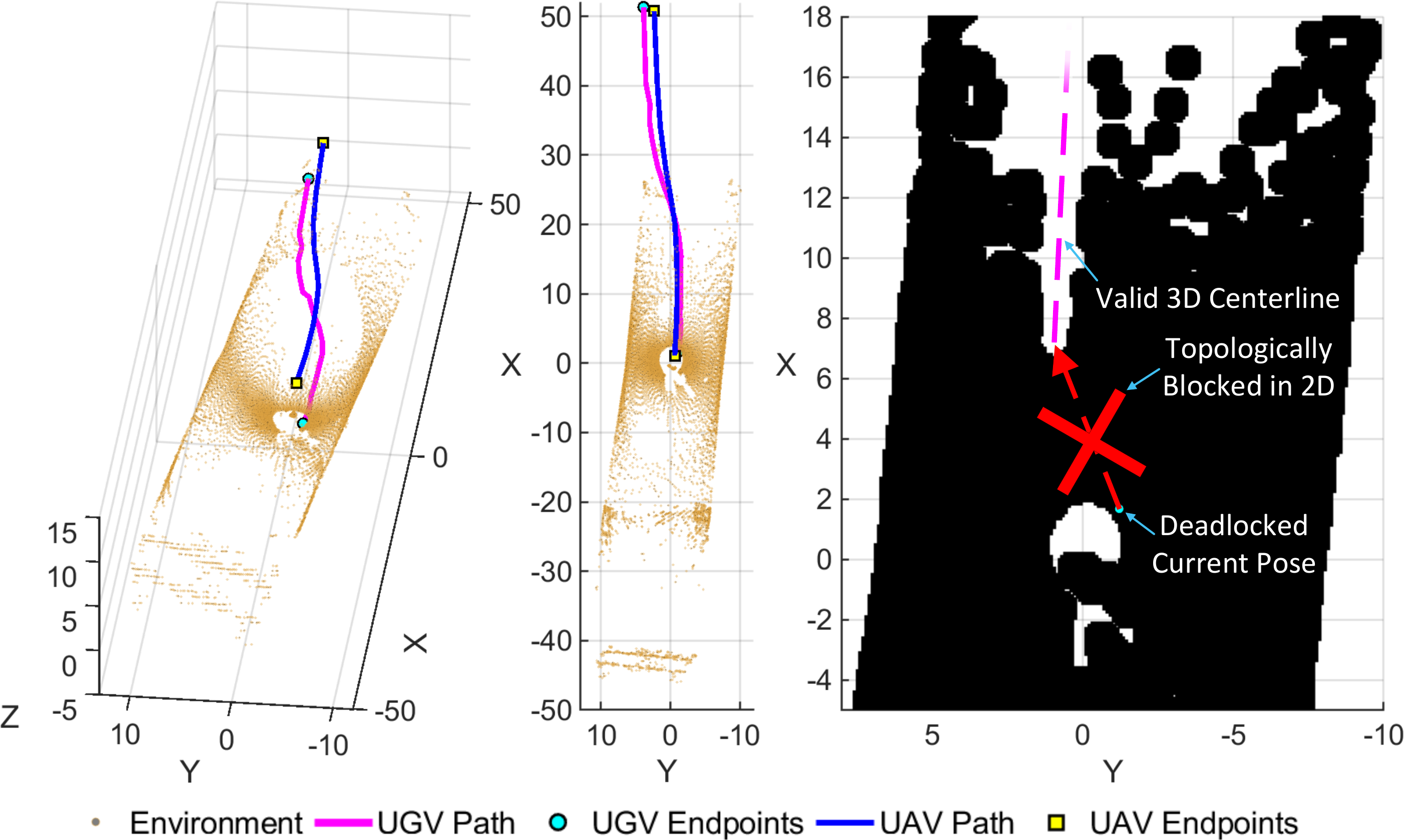}
	\caption{Topological deadlock under UGV roll perturbation. The 2D costmap  (right) artificially blocks the valid 3D corridor, trapping 2D planners. FLISP successfully extracts a 3D gravity-aligned recovery path.}
	\label{fig:teb_deadlock}
\end{figure}

Navigation relying on 2D costmap projections fails under 3D spatial disturbances (Fig. \ref{fig:teb_deadlock}). If the UGV slips onto the curved sidewall, the 2D projection artificially casts the surrounding walls as an impenetrable forward boundary. This pseudo-collision state forces standard 2D planners into an unrecoverable topological deadlock (marked by the red cross). Conversely, FLISP evaluates geometry natively in 3D. By integrating the IMU gravity vector, it synthesizes a valid 3D recovery path back to the tunnel floor (magenta path), remaining completely immune to 2D projection artifacts.

\subsection*{D. Computational Efficiency}
Table \ref{tab:computation_time} summarizes the processing time required per update. In constrained environments, TEB's dense graph optimization suffers from constraint explosion, pushing computation beyond $100\text{ms}$ with high variance. Conversely, FLISP abstracts 3D point clouds into a macroscopic geometric polynomial. This mathematical dimensionality reduction allows FLISP to evaluate a $50\text{m}$ horizon in just $\sim7\text{ms}$, providing the $>100\text{Hz}$ real-time performance essential for eliminating control phase lag in dynamic heterogeneous swarms.

\begin{table}[htbp]
	\centering
	\caption{Computation Time Comparison of Planning Algorithms}
	\label{tab:computation_time}
	
	\renewcommand{\arraystretch}{1.1} 
	\setlength{\tabcolsep}{3pt}        
	
	\begin{tabular}{|l||c|c|}
		\hline
		\textbf{Algorithm} & \textbf{\begin{tabular}{@{}c@{}}Mean Time [ms]\end{tabular}} & \textbf{\begin{tabular}{@{}c@{}}Variance [ms$^2$]\end{tabular}} \\
		\hline
		
		TEB Planner ($5\text{m}$) & 100.02 & 9.88 \\
		TEB Planner ($10\text{m}$) & 98.04 & 162.41 \\
		\textbf{FLISP (Ours)} & \textbf{7.05} & \textbf{4.26} \\
		\hline
		\multicolumn{3}{p{230pt}}{\footnotesize \textit{\textbf{Note:} Data collected over 300 planning cycles on an identical onboard computing platform. The lower mean time for the TEB $10\text{m}$ horizon is attributed to frequent early-exit optimization failures in non-convex blind curves, which simultaneously induces the significantly higher variance.}}\\
	\end{tabular}
\end{table}

\section*{ACKNOWLEDGMENT}
The authors are grateful for the assistance of State Key Laboratory of Autonomous Intelligent Unmanned Systems, Tongji University, China.

\begin{IEEEbiography}[{\includegraphics[width=1in,height=1.25in,clip,keepaspectratio]{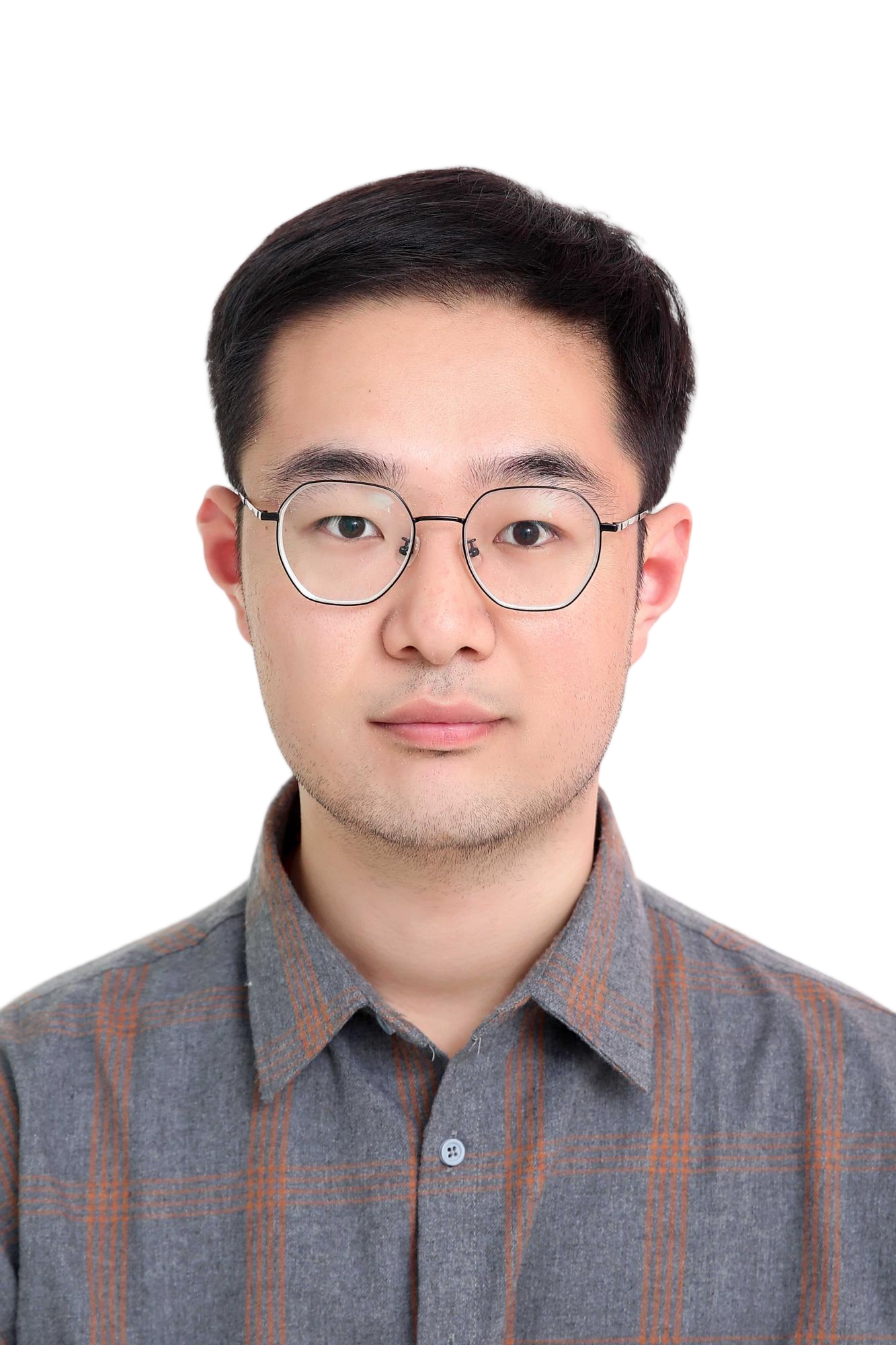}}]
{FENGHE GUO}~(Student Member, IEEE) received the B.S. degree in mechatronics from Xi'an Aeronautical University, Xi'an, China, in 2019, and the M.S. degree in advanced control and systems engineering from the University of Sheffield, Sheffield, U.K., in 2020. He is currently pursuing the Ph.D. degree in intelligent science and technology with Tongji University, Shanghai, China.

His current research interests include heterogeneous multi-agent systems, integrated planning and control, and interpretable learning for autonomous field robotics (e-mail: guofenghe@tongji.edu.cn).
\end{IEEEbiography}

\begin{IEEEbiography}[{\includegraphics[width=1in,height=1.25in,clip,keepaspectratio]{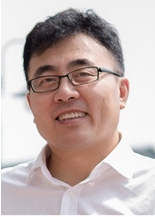}}]
	{RUNJIE SHEN}~received the B.S. degree from Xi’an Polytechnic University, China, in 1995, the M.S. degree from Xi’an Jiaotong University, China, in 1998, and the Ph.D. degree from Zhejiang University, China, in 2004. From 2004 to 2006, he held a post-doctoral position with Zhejiang University, where he became an Associate Researcher in 2006. Since 2011, he has been an Associate Researcher at Tongji University. 
	
	His current research interests are in positioning and navigation of UAVs, sensor measurement, image processing and multi-agent cooperative systems.
\end{IEEEbiography}

\begin{IEEEbiography}[{\includegraphics[width=1in,height=1.25in,clip,keepaspectratio]{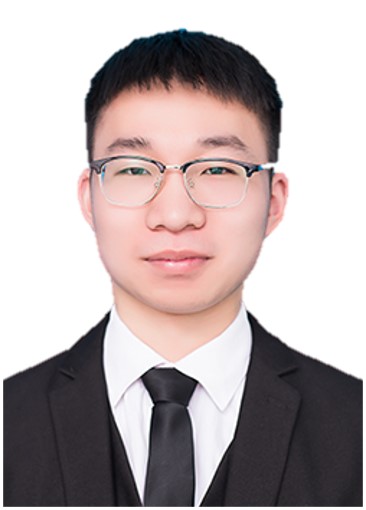}}]
	{CHENYANG SUN}~received the B.S. degree in Automation from Anhui University, Hefei, China, in 2021. He is currently pursuing the Ph.D. degree in Control Science and Engineering at Tongji University, Shanghai, China. 
	
	His current research interests include the hardware design, interactive control and motion planning of heterogeneous multirotors used for complex aerial physical interaction tasks.
\end{IEEEbiography}

\begin{IEEEbiography}[{\includegraphics[width=1in,height=1.25in,clip,keepaspectratio]{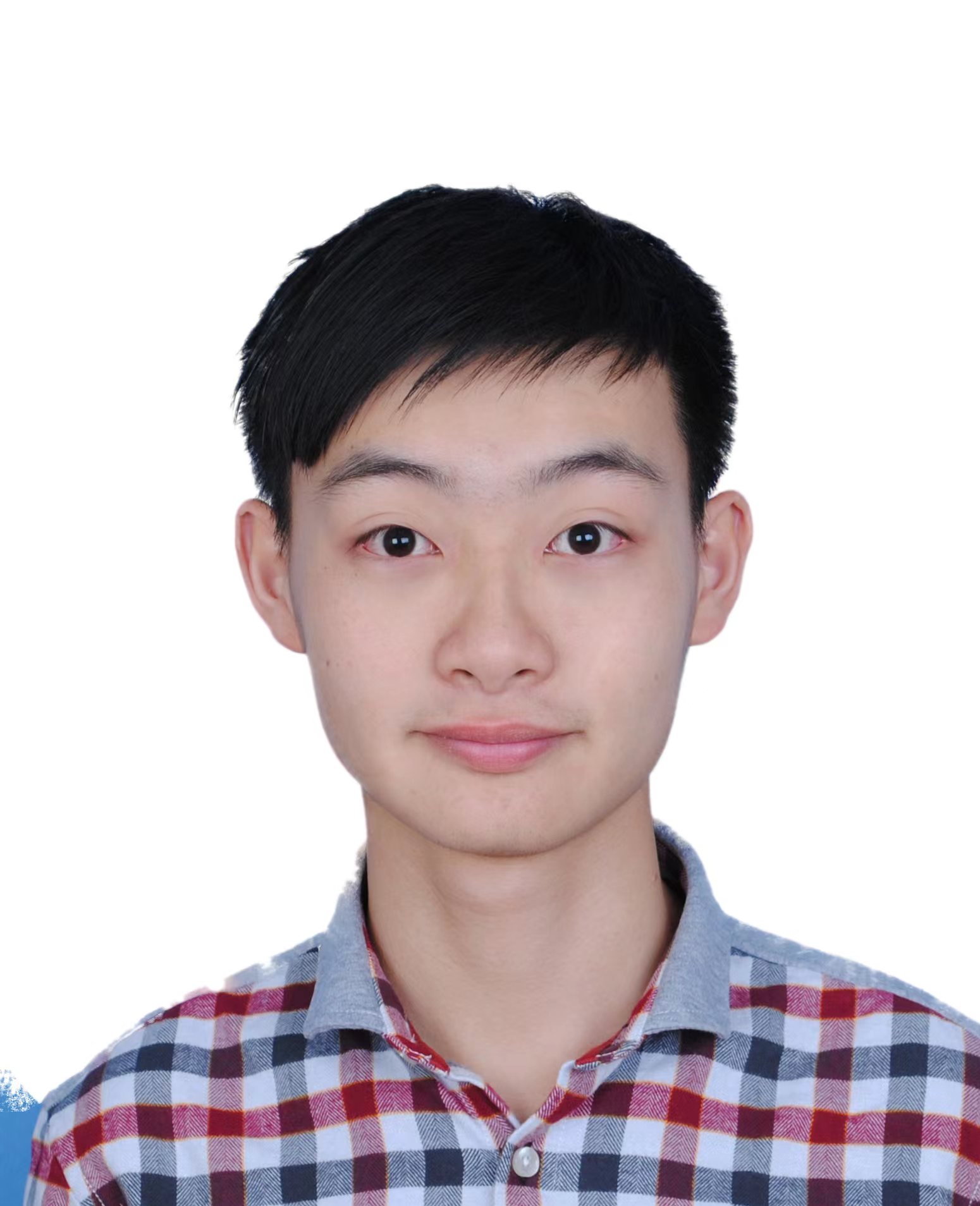}}]
	{JUNRUI ZHANG}~received the M.S. degree in Advanced control and systems engineering from the University of Sheffield, UK, in 2020, and the B.S. degree in Applied Physics from Northeastern University, Shenyang, China, in 2019. He is a Ph.D. student in Control Science and Engineering at Tongji University, Shanghai, China. 
	
	His current research interests include the multi-sensor data fusion for pose estimation of UAVs under the GNSS-denied environment.
\end{IEEEbiography}

\begin{IEEEbiography}[{\includegraphics[width=1in,height=1.25in,clip,keepaspectratio]{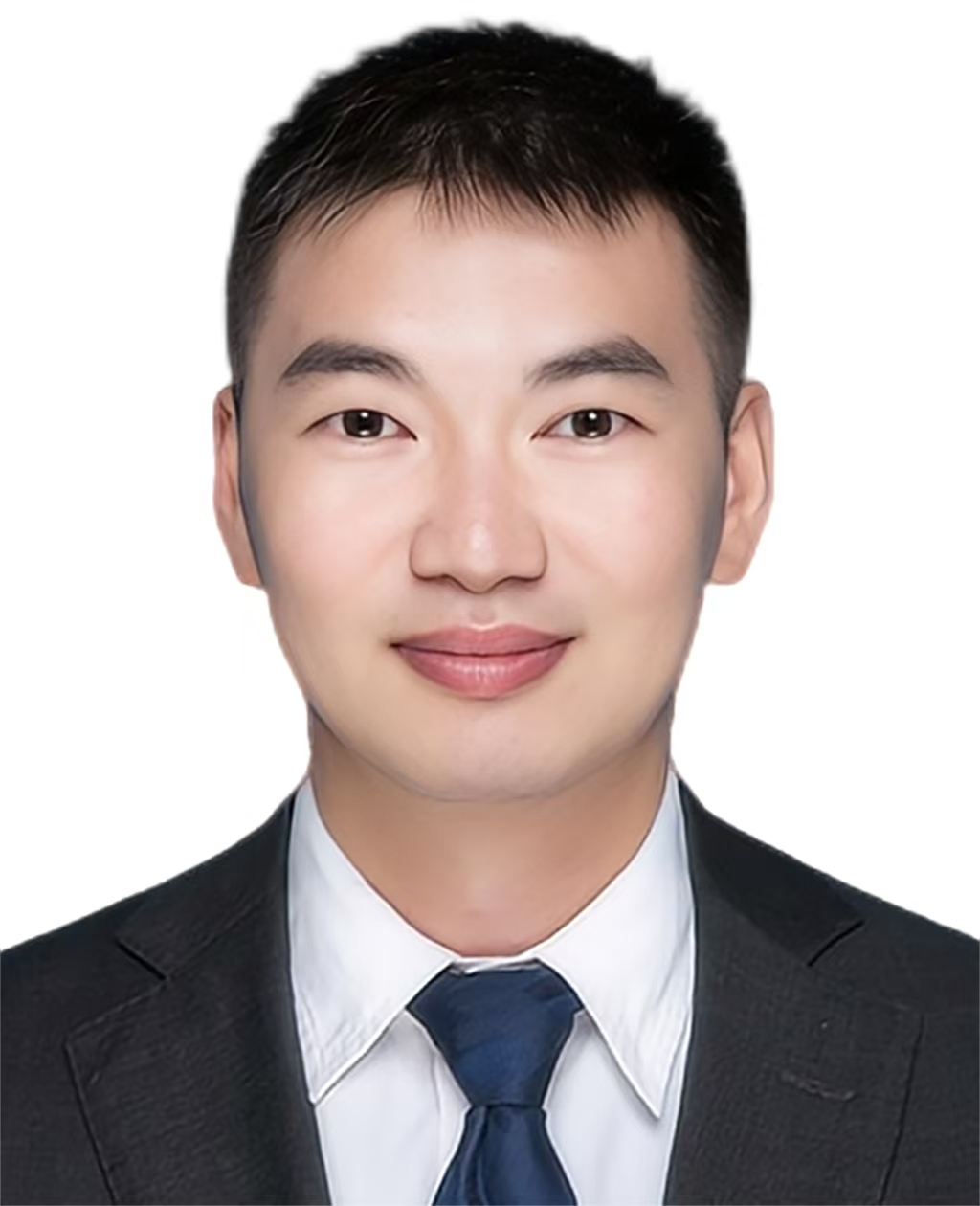}}]
	{QUANXI ZHAN}~received the B.S. degree in process equipment and control engineering from Dalian University of Technology, Dalian, China, in 2019. He is currently pursuing the Ph.D. degree in control science and engineering with Tongji University,  Shanghai, China.
	
	His current research interests include positioning and navigation of UAV, UAV control, drone inspections in underground spaces.
\end{IEEEbiography}

\begin{IEEEbiography}[{\includegraphics[width=1in,height=1.25in,clip,keepaspectratio]{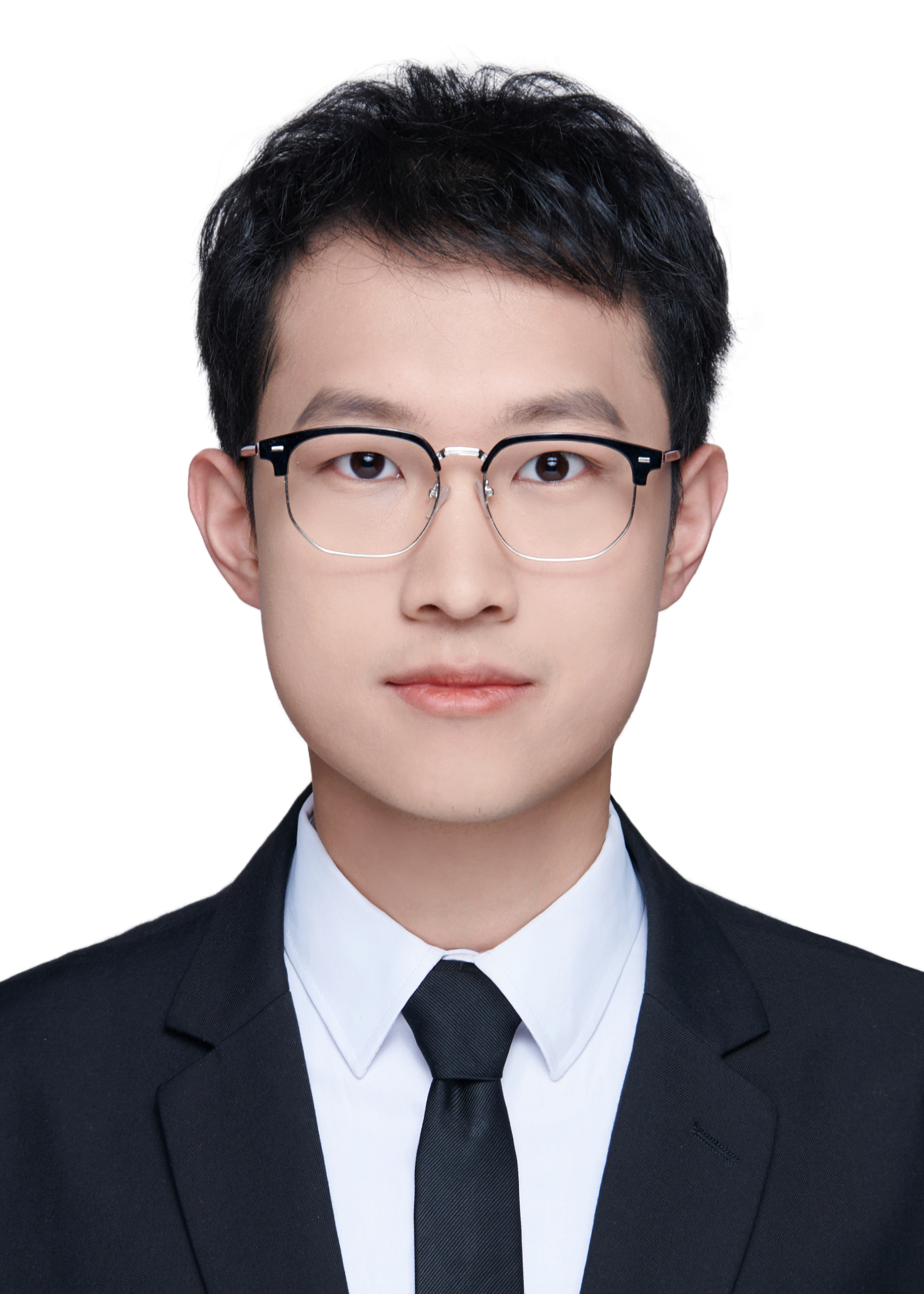}}]
	{YONGCHUN WANG}~received the B.S. degree in automation engineering from Tongji University, Shanghai, China, in 2023. He is currently pursuing the M.Sc. degree in control science and engineering with Tongji University, Shanghai, China.
	
	His current research interests include collaborative control of UGV and UAV in underground tunnels.
\end{IEEEbiography}

\begin{IEEEbiography}[{\includegraphics[width=1in,height=1.25in,clip,keepaspectratio]{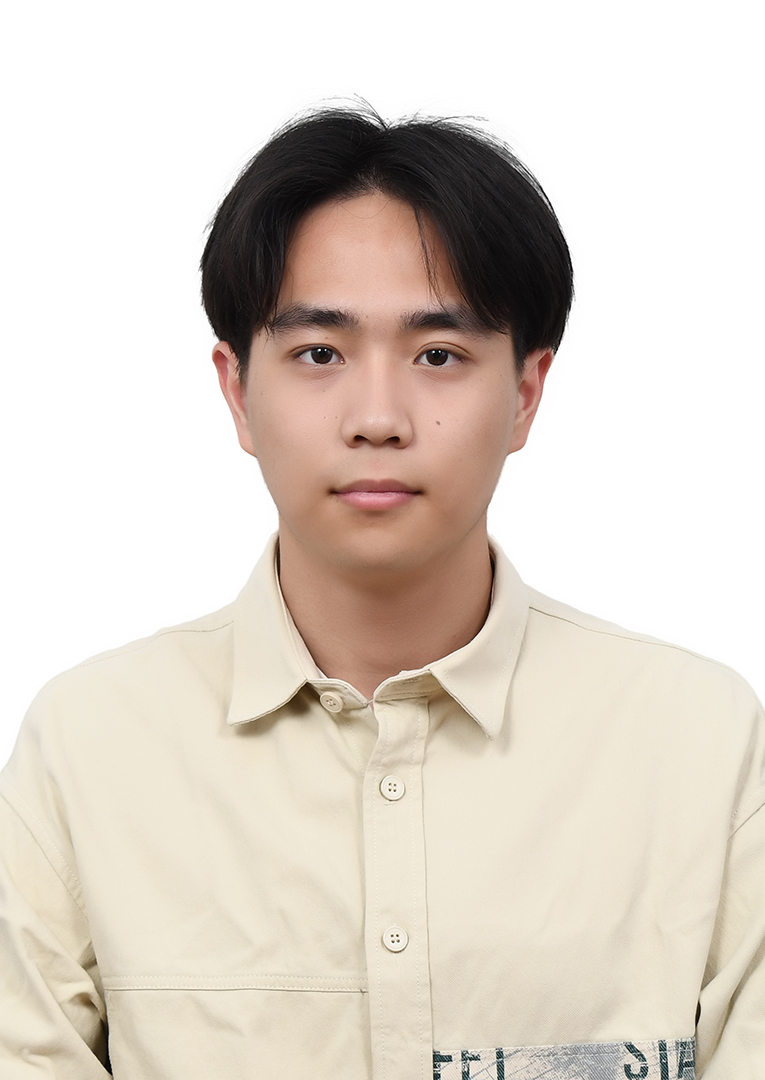}}]
	{JUNJIE ZHANG}~received the B.S. degree in Electrical Engineering and Automation from East China University of Science and Technology, Shanghai, China, in 2024. He is currently pursuing the M.S. degree in Control Science and Engineering with Tongji University, Shanghai, China.
	
	His current research interests include relative positioning and multi-agent collaborative system.
\end{IEEEbiography}

\vfill\pagebreak


\begin{thebibliography}{00}\leftskip1pc
	\bibitem{bib1} H. Huang, C. Wang, M. Zhou, {\it et al.}, ``Compressive strength detection of tunnel lining using hyperspectral images and machine learning,'' {\it Tunnelling and Underground Space Technology}, vol.  153, p. 105979, 2024.
	
	\bibitem{bib3} N. Özkan and H. Arabacı, ``Examination of the environmental and economic effects of hydroelectric power plants,'' {\it SSRJ Social Sciences Research Journal}, vol. 13, no. 1, pp. 285--297, 2024.
	
	\bibitem{bib4} X. Fu and N. Angkawisittpan, ``Detecting surface defects of heritage buildings based on deep learning,'' {\it Journal of Intelligent Systems}, vol. 33, no.  1, p. 20230048, 2024.
	
	\bibitem{bib5} W. Wang, X.  Xu, and H. Yang, ``Intelligent detection of tunnel leakage based on improved Mask R-CNN,'' {\it Symmetry}, vol. 16, no. 6, p.  709, 2024.
	
	\bibitem{bib6} R. Zhang, G.  Hao, K. Zhang, {\it et al.}, ``Reactive UAV-based automatic tunnel surface defect inspection with a field test,'' {\it Automation in Construction}, vol. 163, p.  105424, 2024.
	
	\bibitem{bib7} Y. He, Z. Liu, Y. Guo, {\it et al.}, ``UAV based sensing and imaging technologies for power system detection, monitoring and inspection: A review,'' {\it Nondestructive Testing and Evaluation}, pp. 1--68, 2024.
	
	\bibitem{bib8} H. Liang, J. K. W. Yeoh, and D. K. H. Chua, ``Robust alignment of UGV perspectives with BIM for inspection in indoor environments,'' {\it Journal of Computing in Civil Engineering}, vol. 38, no. 4, p. 04024018, 2024.
	
	\bibitem{bib9} A. J. Lee, W. Song, B. Yu, {\it et al.}, ``Survey of robotics technologies for civil infrastructure inspection,'' {\it Journal of Infrastructure Intelligence and Resilience}, vol. 2, no. 1, p. 100018, 2023.
	
	\bibitem{bib24} H. Qin, S. Shao, T. Wang, {\it et al.}, ``Review of autonomous path planning algorithms for mobile robots,'' {\it Drones}, vol. 7, no. 3, p.  211, 2023.
	
	\bibitem{bib25} X. Xu, J. Zeng, Y.  Zhao, {\it et al.}, ``Research on global path planning algorithm for mobile robots based on improved A*,'' {\it Expert Systems with Applications}, vol. 243, p. 122922, 2024.
	
	\bibitem{bib26} C. S. Tan, R. Mohd-Mokhtar, and M. R. Arshad, ``A comprehensive review of coverage path planning in robotics using classical and heuristic algorithms,'' {\it IEEE Access}, vol. 9, pp. 119310--119342, 2021.
	
	\bibitem{bib27} R. Chai, Y. Guo, Z. Zuo, {\it et al.}, ``Cooperative motion planning and control for aerial-ground autonomous systems: Methods and applications,'' {\it Progress in Aerospace Sciences}, vol. 146, p. 101005, 2024.
	
	\bibitem{bib28} Z. Wang and X. Fan, ``A systematic review and analysis of A-star pathfinding algorithms and its variations,'' in {\it Proc. 4th Int. Conf. Signal Processing and Machine Learning (CONF-SPML)}, 2024, vol. 13077, pp. 157--182.
	
	\bibitem{bib29} S. Xiong, Y. Zhang, C. Wu, {\it et al.}, ``Energy management strategy of intelligent plug-in split hybrid electric vehicle based on deep reinforcement learning with optimized path planning algorithm,'' {\it Proc.  Inst. Mech. Eng., Part D:  J. Automobile Engineering}, vol. 235, no. 14, pp. 3287--3298, 2021.
	
	\bibitem{bib30} D. K. Muhsen, F. A. Raheem, and A. T. Sadiq, ``A systematic review of rapidly exploring random tree RRT algorithm for single and multiple robots,'' {\it Cybernetics and Information Technologies}, vol. 24, no. 3, 2024.
	
	\bibitem{bib31} A. Tanveer, M.  T. Ashraf, and U. Khan, ``Motion planning for autonomous ground vehicles using artificial potential fields:  A review,'' {\it arXiv preprint arXiv: 2310.14339}, 2023.
	
	\bibitem{bib32} Y. Hu, H. Long, and M. Chen, ``The analysis of pedestrian flow in the smart city by improved DWA with robot assistance,'' {\it Scientific Reports}, vol. 14, no. 1, p.  11456, 2024.
	
	\bibitem{bib33} M. Ganesan, S.  Kandhasamy, B. Chokkalingam, {\it et al.}, ``A comprehensive review on deep learning-based motion planning and end-to-end learning for self-driving vehicle,'' {\it IEEE Access}, vol. 12, pp. 41983--42014, 2024.
	
	\bibitem{bib_ozaslan} T. Özaslan, G. Loianno, J. Keller, {\it et al.}, ``Autonomous navigation and mapping for inspection of penstocks and tunnels with MAVs,'' {\it IEEE Robotics and Automation Letters}, vol. 2, no. 3, pp. 1740--1747, 2017.
	
	\bibitem{bib34} W. Hu, S. Chen, Z. Liu, {\it et al.}, ``HA-RRT: A heuristic and adaptive RRT algorithm for ship path planning,'' {\it Ocean Engineering}, vol. 316, p. 119906, 2025.
	
	\bibitem{bib35} C. Huang, T. Wang, S. Wang, {\it et al.}, ``Dynamic path planning for spacecraft rendezvous and approach based on hybrid honey badger algorithm,'' {\it Journal of the Franklin Institute}, vol. 362, no. 1, p. 107398, 2025.
	
	\bibitem{bib36} Y. Zhong and Y. Wang, ``Cross-regional path planning based on improved Q-learning with dynamic exploration factor and heuristic reward value,'' {\it Expert Systems with Applications}, vol. 260, p. 125388, 2025.
	
	\bibitem{bib37} X. Zhao, R. Yang, L. Zhong, {\it et al.}, ``Multi-UAV path planning and following based on multi-agent reinforcement learning,'' {\it Drones}, vol.  8, no. 1, p. 18, 2024.
	
	\bibitem{bib38} B. Feng, Y. Bi, M. Li, {\it et al.}, ``A decentralized multi-agent path planning approach based on imitation learning and selective communication,'' {\it Journal of Computing and Information Science in Engineering}, vol. 24, no. 8, 2024.
	
	\bibitem{bib39} L. Zhang, Z. Cai, Y. Yan, {\it et al.}, ``Multi-agent policy learning-based path planning for autonomous mobile robots,'' {\it Engineering Applications of Artificial Intelligence}, vol.  129, p. 107631, 2024.
	
	\bibitem{bib40} C. Fang, J. Mao, D. Li, {\it et al.}, ``A coordinated scheduling approach for task assignment and multi-agent path planning,'' {\it Journal of King Saud University-Computer and Information Sciences}, vol. 36, no. 1, p.  101930, 2024.
	
	\bibitem{bib41} J. Li, T. Sun, X. Huang, {\it et al.}, ``A memetic path planning algorithm for unmanned air/ground vehicle cooperative detection systems,'' {\it IEEE Trans. Automation Science and Engineering}, vol. 19, no. 4, pp.  2724--2737, 2021.
	
	\bibitem{bib42} G. Niu, L. Wu, Y. Gao, {\it et al.}, ``Unmanned aerial vehicle (UAV)-assisted path planning for unmanned ground vehicles (UGVs) via disciplined convex-concave programming,'' {\it IEEE Trans. Vehicular Technology}, vol. 71, no. 7, pp. 6996--7007, 2022.
	
	\bibitem{bib43} D. Hu, V. J. L. Gan, T. Wang, {\it et al.}, ``Multi-agent robotic system (MARS) for UAV-UGV path planning and automatic sensory data collection in cluttered environments,'' {\it Building and Environment}, vol. 221, p. 109349, 2022.
	
	\bibitem{bib44} Z. Li, W. Zhao, and C. Liu, ``Completion time minimization for UAV-UGV-enabled data collection,'' {\it Sensors}, vol. 22, no.  15, p. 5839, 2022.
	
	\bibitem{bib45} S. Martinez-Rozas, D.  Alejo, F. Caballero, {\it et al.}, ``Path and trajectory planning of a tethered UAV-UGV marsupial robotic system,'' {\it IEEE Robotics and Automation Letters}, vol. 8, no. 10, pp. 6475--6482, 2023.
	
	\bibitem{bib46} J. Zhang, Z. Huang, X. Zhu, {\it et al. }, ``LOFF: LiDAR and optical flow fusion odometry,'' {\it Drones}, vol. 8, no. 8, p. 411, 2024.
	
	\bibitem{bib47} Q. Zhan, J. Zhang, C. Sun, {\it et al.}, ``Automatic UAV inspection of hydropower station diversion pipelines using robust cylinder-like fitting for incomplete point clouds,'' in {\it Proc. 2024 China Automation Congress (CAC)}, 2024, pp. 5931--5937.
	
	\bibitem{bib48} M. De Petrillo, J. Beard, Y. Gu, {\it et al.}, ``Search planning of a UAV/UGV team with localization uncertainty in a subterranean environment,'' {\it IEEE Aerospace and Electronic Systems Magazine}, vol. 36, no. 6, pp. 6--16, 2021.
	
	\bibitem{bib49} N. Kottege, J. Williams, B. Tidd, {\it et al. }, ``Heterogeneous robot teams with unified perception and autonomy: How Team CSIRO Data61 tied for the top score at the DARPA Subterranean Challenge,'' {\it Field Robotics}, vol. 4, pp. 313--359, 2024.
	
	\bibitem{bib_costar} B. Morrell, K. Otsu, A. Agha, {\it et al.}, ``An addendum to NeBula: Toward extending team CoSTAR's solution to larger scale environments,'' {\it IEEE Trans. Field Robotics}, vol. 1, pp. 476--526, 2024.
	
	\bibitem{bib_step} A. Dixit, D. D. Fan, K. Otsu, {\it et al.}, ``STEP: Stochastic traversability evaluation and planning for risk-aware navigation; Results from the DARPA Subterranean Challenge,'' {\it Field Robotics}, vol. 4, pp. 182--210, 2024.
	
	\bibitem{bib_csiro_frontier} J. Williams, S. Jiang, M. O'Brien, {\it et al.}, ``Online 3D frontier-based UGV and UAV exploration using direct point cloud visibility,'' in {\it Proc. IEEE Int. Conf. Multisensor Fusion and Integration for Intelligent Systems (MFI)}, 2020, pp. 263--270.

	\bibitem{bib50} C. Zhang, X. Yang, R. Zhou, {\it et al.}, ``A path planning method based on improved A* and fuzzy control DWA of underground mine vehicles,'' {\it Applied Sciences}, vol. 14, no. 7, p. 3103, 2024.
	
	\bibitem{bib51} X. Chen, C. Yang, H. Hu, {\it et al.}, ``A hybrid DWA-MPC framework for coordinated path planning and collision avoidance in articulated steering vehicles,'' {\it Machines}, vol. 12, no. 12, p. 856, 2024.
	
	\bibitem{bib52} L. Wang, Y. Ning, H. Chen, {\it et al.}, ``Autonomous flights inside narrow tunnels,'' {\it IEEE Trans. Robotics}, vol. 41, pp. 596--612, 2025.
	
	\bibitem{bib57} X. Zhou, Z. Wang, H. Ye, {\it et al.}, ``EGO-Planner: An ESDF-free gradient-based local planner for quadrotors,'' {\it IEEE Robotics and Automation Letters}, vol. 6, no. 2, pp.  478--485, 2021.
	
	\bibitem{bib58} T. Shan, B. Englot, D. Meyers, {\it et al. }, ``LIO-SAM: Tightly-coupled lidar inertial odometry via smoothing and mapping,'' in {\it Proc. IEEE/RSJ Int. Conf.  Intelligent Robots and Systems (IROS)}, 2020, pp. 5135--5142.
	
	\bibitem{bib59} J. D. Gammell, S. S. Srinivasa, and T. D. Barfoot, ``Informed RRT*: Optimal sampling-based path planning focused via direct sampling of an admissible ellipsoidal heuristic,'' in {\it Proc. IEEE/RSJ Int. Conf.  Intelligent Robots and Systems (IROS)}, 2014, pp. 2997--3004.
	
	\bibitem{bib60} W. Xu, Y. Cai, D. He, {\it et al.}, ``FAST-LIO2: Fast direct LiDAR-inertial odometry,'' {\it IEEE Trans. Robotics}, vol. 38, no. 4, pp. 2053--2073, 2022.
	
	\bibitem{bib61} P. E. Hart, N. J. Nilsson, and B. Raphael, ``A formal basis for the heuristic determination of minimum cost paths,'' {\it IEEE Trans. Systems Science and Cybernetics}, vol. 4, no. 2, pp. 100--107, 1968.
	
	\bibitem{bib_new_cbf} F. Guo, R. Shen, J. Zhang, et al., ``Adaptive and resilient consensus for UAV-UGV teams via virtual agent mediation and topology switching,'' in \textit{2025 IEEE International Conference on Robotics and Biomimetics (ROBIO)}. IEEE, 2025, pp. 745--750.
	
\end{thebibliography}
\end{document}